\documentclass[lettersize,journal]{IEEEtran}
\usepackage{amsmath,amsfonts}
\usepackage{algorithmic}
\usepackage{algorithm}
\usepackage{array}
\usepackage[caption=false,font=normalsize,labelfont=sf,textfont=sf]{subfig}
\usepackage{textcomp}
\usepackage{stfloats}
\usepackage{url}
\usepackage{verbatim}
\usepackage{graphicx}
\usepackage{cite}
\usepackage{multirow}
\usepackage{romannum}
\usepackage{diagbox}
\usepackage{optidef}
\usepackage{bm}
\usepackage{hyperref}
\usepackage{xcolor}

\usepackage{amssymb,stmaryrd}

\newtheorem{remark}{Remark}

\begin{document}

\title{
Multi-Robot Relative Pose Estimation in SE(2) with Observability Analysis: A Comparison of Extended Kalman Filtering and Robust Pose Graph Optimization
%Multi-Robot Relative Pose Estimation in SE(2) with Observability Analysis: An Overview with Some New Results
%An Overview of Cooperative Multi-Robot Localization: Relative Pose Estimation in SE(2) with Observability Analysis
%An Overview of Observability Analysis for Cooperative Multi-Robot Localization: Relative Pose Estimation in SE(2)
%Observability Analysis for Cooperative Multi-Robot Localization: Absolute and Relative Pose Estimation in 2D Space
}

\author{Kihoon Shin, Hyunjae Sim, Seungwon Nam, Yonghee Kim, Jae Hu and Kwang-Ki K. Kim${}^{*}$
\thanks{The authors are with the department of electrical and computer engineering at Inha University,
Incheon 22212, Korea}.
\thanks{${}^{*}$Corresponding author (Email: {\tt kwangki.kim@inha.ac.kr})}
\thanks{The first two authors, K. Shin and H. Sim, contributed equally.}
\thanks{This research was supported by the Basic Science Research Program through the National Research Foundation of Korea (NRF) funded by the Ministry of Education (NRF-2022R1F1A1076260).}
}

\maketitle

\begin{abstract}
In this study, we address multi-robot localization issues, with a specific focus on cooperative localization and observability analysis of relative pose estimation. Cooperative localization involves enhancing each robot's information through a communication network and message passing. If odometry data from a target robot can be transmitted to the ego robot, observability of their relative pose estimation can be achieved through range-only or bearing-only measurements, provided both robots have non-zero linear velocities. In cases where odometry data from a target robot are not directly transmitted but estimated by the ego robot, both range and bearing measurements are necessary to ensure observability of relative pose estimation.
For ROS/Gazebo simulations, we explore four sensing and communication structures. We compare extended Kalman filtering (EKF) and pose graph optimization (PGO) estimation using different robust loss functions (filtering and smoothing with varying batch sizes of sliding windows) in terms of estimation accuracy. In hardware experiments, two Turtlebot3 equipped with UWB modules are used for real-world inter-robot relative pose estimation, applying both EKF and PGO and comparing their performance. \\[1mm] 
\indent{\tt Code:} \href{https://github.com/22222140/MR_RPEstm}{\tt https://github.com/iASL/MRS/MR\_RPEstm} \\[1mm]
\indent{\tt video:} \href{https://youtu.be/KG3AJOKxPkA}{\tt https://youtu.be/KG3AJOKxPkA}
\end{abstract}


%\begin{figure*}[!th]
%\tableofcontents
%\end{figure*}
	
%%%%%%%%%%%%%%%%%%%%%%%%%%%%%%%%%%%%%%%%%%%
% MAIN PART %%%%%%%%%%%%%%%%%%%%%%%%%%%%%%%%%%%
%%%%%%%%%%%%%%%%%%%%%%%%%%%%%%%%%%%%%%%%%%%

%	\phantom{
%	% mobile robot observability analysis 
%		\cite{martinelli2005observability}
%		\cite{Huang2010consistentEKF}
%		\cite{Hesch2014cameraimu}
%		\cite{Martinelli2011calibration}
%		\cite{huang2011observability}
%	% cooperative localization
%		\cite{miller2020cooperative}
%		\cite{martinelli2005multi}
%		% relative pose estimation
%			\cite{wang2020anomaly}
%			\cite{zhou2008robot}
%			% manifold optimization
%				\cite{garcia2021certifiable}
%			% Kalman filtering
%				\cite{Fang2022adaptive}
%		% range-only
%			\cite{araki2019range}
%		% bearing-only
%			\cite{Sharma2012graph}
%		% sensor fusion
%			% camera
%			% uwb
%			% imu
%	% multi-robot coordination control
%		\cite{das2002vision} % vision-based 
%		\cite{sakurama2021generalized} % without treating any extrinsic calibration -> purely control-theoretic point of view
%	% app: ground vehicles
%%		\cite{}
%	% app: marine vehicles
%		\cite{Bahr2009underwatervehicles}
%	}
	
%==============================================================%
\section{Introduction}
\label{sec:intro}
%==============================================================%

%this is a review paper ... 
%nothing new (yet) ...
%a tutorial overview ...
%future directions ...
%perspectives ... 
%\begin{itemize}
%	\item
%	Cooperative localization
%	\item
%	Cooperative SLAM
%	\item
%	Autonomous driving
%	\item
%	Robot fleet
%	\item
%	Formation control (not just position, but also attitude or orientation)
%	\item
%	Multi UAV control
%	\item
%	Distributed autonomous mobile robots
%\end{itemize}

\begin{figure}[!t]
	\centerline{\includegraphics[width=.95\linewidth]{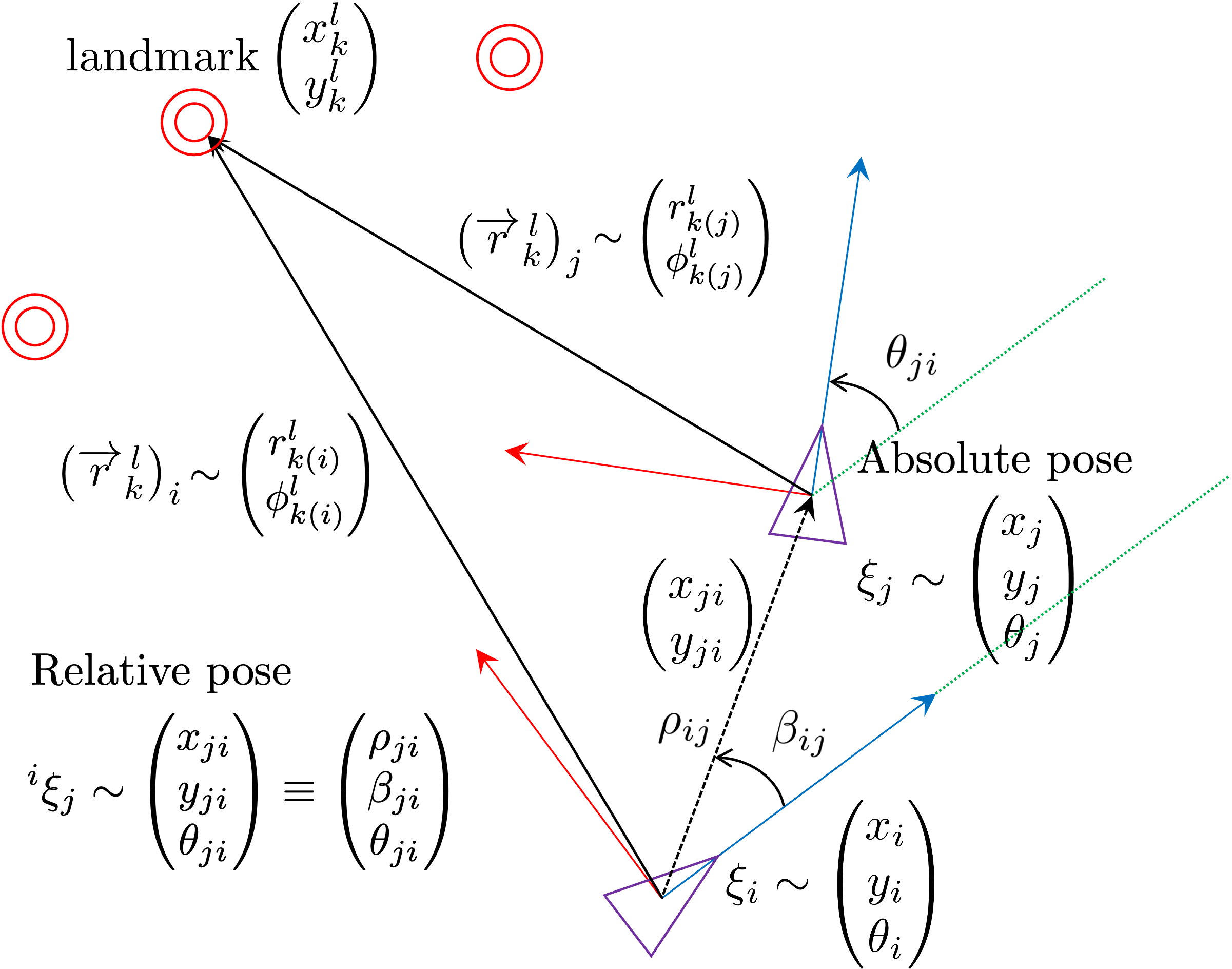}}\vspace{-1mm}
	\caption{Schematic for multi-robot localization: (a) (Centralized) multi-robot localization is to estimate the set of {\em absolute} poses $\{\xi_{i}\}_{i \in {\mathcal N}}$ for multiple robots in the set ${\mathcal N}$; (b) Distributed multi-robot localization is to estimate the state $\{\xi_{i}, \{{}^{i}\xi_{j}\}_{j \in {\mathcal N}_i}\}$ concatenating the absolute pose of the ego-robot and the {\em relative} poses of the neighborhood-robots for each Robot ${\tt R}_{i}$, $i \in {\mathcal N}$ where the sets of robots ${\mathcal N}$ and ${\mathcal N}_{i}$ (for $i \in {\mathcal N}$) could be time-varying; and (c) Cooperative localization is to solve (a) or (b) by communicating extra information such as odometry data and state estimate.}
	\label{fig:multirobotlocalization}
\end{figure}

% general statement of multi-robot localization: necessity and background of research
For fully autonomous mobile robot systems (AMRS), each robot in the team must estimate its absolute pose (position and orientation) and the relative poses of its neighboring robots in the body-fixed reference frame of the corresponding ego robot \cite{siegwart2011introduction,Yang2008decentralized,sakurama2021generalized}. This problem is known as multi-robot localization\cite{roumeliotis2002distributed}, a crucial capability for AMRS in applications such as search and rescue missions, warehouse logistics, formation control, and sensor coverage \cite{howard2006multi,thrun2005multi,Queralta2020collaborative}. As depicted in Fig.\ref{fig:multirobotlocalization}, the problem involves determining the absolute and relative poses of the robots. A particular interest lies in distributed multi-robot localization, where each robot determines the ego-centric state, absolute pose, and relative pose of neighboring robots in terms of ego-robot reference coordinates.

% cooperative localization of multi-robot systems
In multi-robot state estimation, cooperative localization (of absolute and relative pose estimation) via communicating messages (state, odometry, and raw data exchange) has been extensively researched ~\cite{martinelli2005multi,miller2020cooperative,wang2023distributed}. Centralized cooperative localization in AMRS estimates the absolute pose of all robots, utilizing both relative measurements between robots and absolute measurements of landmarks. Distributed cooperative localization determines the absolute and relative poses of neighboring robots for each ego robot in AMRS.

% relative pose estimation of multi-robot systems
Real-time distributed relative pose estimation is crucial for the cooperative control of multirobot systems~\cite{sakurama2021generalized,Desai2001formationcontrol,das2002vision}. Various sensors (Camera, LiDAR, UWB, etc.) and estimation methods, such as maximum likelihood estimation (MLE), Kalman filtering (KF), and factor graph optimization (FGO), have been employed for relative pose estimation in multirobot systems. Examples include vision-based relative pose estimation using manifold optimization~\cite{garcia2021certifiable,kim2023} and LiDAR-based relative pose estimation using point-cloud registration with an adaptive cubature split-covariance intersection filter~\cite{Fang2022adaptive}.

% observability analysis of wheeled-mobile multi-robot systems 
Observability is a fundamental property of robotic perception for state estimation in dynamic systems. For state estimation of mobile multirobot systems, the theory of nonlinear observability~\cite{haynes1970nonlinear,hermann1977nonlinear,van1982controllability} is applied to observability analysis for mobile robot localization~\cite{bicchi1998observability,conticelli2000observability,lorussi2001optimal} and multi-robot localization concerning various types of sensor measurements. 
(i) range and bearing measurements~\cite{martinelli2005observability,Huang2010consistentEKF,huang2011observability,Martinelli2011calibration}, (ii) range measurements~\cite{araki2019range} and (iii) bearing measurements~\cite{Sharma2012graph}.
Observability analysis for a visual-inertial navigation system (VINS) of mobile robots has also been investigated~\cite{hesch2014cameraimu,panahandeh2013observability,yang2019aided,panahandeh2016planar,paul2018alternating,yang2019observability,yang2019aided,huai2022observability}, and observability-constrained navigation~\cite{hesch2012observability,gomaa2020observability,liu2022variable} is a key challenge owing to the limited field of view in visual perception.
% nonlinear unknown input observability:
In addition to addressing the observability of multirobot systems with known kinematic or dynamic inputs, the extension of the observability rank condition to nonlinear systems driven by unknown inputs is nontrivial~\cite{martinelli2018nonlinear, martinelli2022nonlinear}.

% the contributions and organization of this manuscript
The contributions of this study are outlined as follows.
This paper offers an overview of nonlinear observability analysis for wheeled mobile robots in $\text{SE}(2)$. For simplicity, we consider two-robot systems. Referring to~\cite{martinelli2005observability}, when the linear velocities of both robots are non-zero, either range-only or bearing-only measurements ensure observability in relative pose estimation, provided that the odometry data of the two robots are mutually available. This result is extended to cases where the odometry data of the other (neighboring) robot is not available to the ego robot. Additionally, the observability analysis presented in this paper is demonstrated in ROS/Gazebo simulation environments. We apply and compare two different \emph{de facto} state estimation methods, EKF and PGO, using four distinct cases of information structures. Furthermore, we test the observability analysis and estimation methods in real-world mobile robot hardware, utilizing two Turtlebot3 robots to demonstrate EKF and PGO-based robot-to-robot relative pose estimation. 
Furthermore, we discuss future research directions for multi-robot relative pose estimation, specifically focusing on (i) uncertainty quantification and propagation in distributed data fusion, transforming and exchanging spatial information, (ii) outlier-robust iterative EKF and PGO using robust kernel functions, and (iii) distributed multi-robot pose SLAM explicitly exploiting range measurements, where a non-convex PGO can be relaxed to a second-order cone program (SOCP) or a semidefinite program (SDP) providing a lower-bound certificate.

% paper organization
The remainder of this paper is organized as follows.
Section~\ref{sec:cooploc} provides an overview of multi-robot state estimation, encompassing localization and mapping, using a distributed EKF and PGO.
Section~\ref{sec:nonobsanal} presents observability analysis of inter-robot relative pose estimation with different measurement and information structures.
In Section~\ref{sec:gazebo}, we demonstrate and compare EKF and PGO-based estimation methods for inter-robot relative pose estimation in four different scenarios of information structures and in the presence of outliers.
Section~\ref{sec:conclusion} concludes the paper and suggests directions for future research.

%%-----------------------------------------------------------------------------%
%\subsection{Related Work}
%\label{sec:intro:literature}
%%-----------------------------------------------------------------------------%
%\noindent
%\textbf{Relative pose estimation}
%
%\noindent
%\textbf{Cooperative localization}
%
%\noindent
%\textbf{Cooperative SLAM}

%\newpage
%==============================================================%
\section{Multi-robot Localization}%\section{Cooperative Localization}
\label{sec:cooploc}
%==============================================================%
We examine two mobile ground robots, denoted as ${\tt R}{i}$ and ${\tt R}{j}$, situated in a 2D workspace. The configuration state of this multiple-robot system can be represented as the vector $X=[x_{i}, y_{i}, \theta_{i}, x_{j}, y_{j}, \theta_{j}]^\top \! \in \text{SE}(2) \times \text{SE}(2)$ that includes the Cartesian absolute coordinates and absolute orientation of the two robots. For Robot ${\tt R}_{i}$, the state of interest is a combination of its absolute pose $X_{i}^{\rm a} = [x_{i}, y_{i}, \theta_{i}] \in \text{SE}(2)$ and the relative pose the Robot ${\tt R}_{j}$ with respect to the ego-Robot ${\tt R}_{i}$ that can be represented by either $X_{ji}^{\rm c} = [x_{ji}, y_{ji}, \theta_{ji}]^\top$ in the cartesian coordinates with the {\tt atan2} orientation or $X_{ji}^{\rm p} = [\rho_{ji}, \beta_{ji}, \theta_{ji}]^\top$ in the polar coordinates with the {\tt atan2} orientation. The augmented state for the robot ${\tt R}_{i}$ is defined as the concatenation of the absolute and relative poses, that is, $X_{i} = [(X_{i}^{\rm a})^{\!\top}, X_{ji}^\top]^\top \! \in \text{SE}(2) \times \text{SE}(2)$.

%-----------------------------------------------------------------------------%
\subsection{Motion model}
\label{sec:cooploc:motion}
%-----------------------------------------------------------------------------%
The dynamics of the augmented state $X_{i}$ can be described by the following dynamic system equation:
\begin{equation}
\dot{X}_{i} = f_{i}(X_{i}, U_{i})
\end{equation}
where $U_{i}=[v_{i}, \omega_{i}, v_{j}, \omega_{j}]^\top$ denotes the control input vector. For a unicycle vehicle model, the analytical expression for the vector field $f_{i}: \text{SE}(2)^2 \times {\mathbb R}^{4}$ is as follows: 
\begin{equation}
f_{i}(X_{i}^{\rm a}, X_{ji}, U_{i}) = 
\begin{bmatrix}
f_{i}^{\rm a}(X_{i}^{\rm a}, U_{i}) \\
f_{ji}^{\rm r}(X_{ji}, U_{i}) 
\end{bmatrix}
\end{equation}
where the absolute pose kinematics are given by
\begin{equation}
%\frac{d}{dt} X_{i}^{\rm a} 
\dot{X}_{i}^{\rm a} 
= f_{i}^{\rm a}(X_{i}^{\rm a}, U_{i}) = 
\begin{bmatrix}
v_{i} \cos\theta_{i} \\
v_{i} \sin\theta_{i} \\
\omega_{i}
\end{bmatrix}
\end{equation}
and the relative pose kinematics are expressed as 
\begin{equation}\label{eq:relpose:1}
%\frac{d}{dt} X_{ji}^{\rm c} 
\dot{X}_{ji}^{\rm c} 
= f_{ji}^{\rm r}(X_{ji}^{\rm c}, U_{i}) = \!
\begin{bmatrix} v_{j}\cos\theta_{ji}\\ v_{j}\sin\theta_{ji} \\ \omega_{j} \end{bmatrix} \!+\!
\begin{bmatrix} y_{ji} \omega_{i} - v_{i} \\ -x_{ji}\omega_{i} \\ -\omega_{i} \end{bmatrix}
\end{equation}
for $X_{ji}=X_{ji}^{\rm c}=[x_{ji}, y_{ji}, \theta_{ji}]^\top$
{or}
\begin{equation}\label{eq:relpose:2}
%\frac{d}{dt} X_{ji}^{\rm p} 
\dot{X}_{ji}^{\rm p} 
= f_{ji}^{\rm r}(X_{ji}^{\rm p}, U_{i}) =\! 
\begin{bmatrix} v_{j}\cos(\theta_{ji} \!-\! \beta_{ji}) - v_{i}\cos\beta_{ji} \\ v_{j}\! \left( \sin(\theta_{ji} \!-\! \beta_{ji}) - \sin\beta_{ji}\right)\!/\rho_{ji} \\ \omega_{j} - \omega_{i} \end{bmatrix}
\end{equation}
for $X_{ji}=X_{ji}^{\rm p}=[\rho_{ji}, \beta_{ji}, \theta_{ji}]^\top$.

In this study, we consider both Cartesian \eqref{eq:relpose:1} and polar \eqref{eq:relpose:2} coordinate systems to define the configuration state vector for (cooperative) localization. For notational convenience, we use $c(\cdot)$ for $\cos(\cdot)$ and $s(\cdot)$ for $\sin(\cdot)$.

%\item
\noindent
\textbf{[Case M-I] With communicating odometry data}
Define the state vector $X_{i}=[x_{i}, y_{i}, \theta_{i}, x_{ji}, y_{ji}, \theta_{ji}]^\top$ or $X_{i}=[x_{i}, y_{i}, \theta_{i}, \rho_{ji}, \beta_{ji}, \theta_{ji}]^\top$ with the odometry vector $U_{i}=[u_i^\top, u_j^\top]^\top = [v_{i}, \omega_{i}, v_{j}, \omega_{j}]^\top$.
\begin{equation}\label{eq:case:1}
%\frac{d}{dt} X_{i}
\dot{X}_{i}
= \sum_{k=1}^{4} g_{k}(X_i) U_{ik} + W_i
\end{equation}
where $g_k(\cdot)$ is either $g_k^{\rm r}(\cdot)$ defined as
\begin{equation}\label{eq:case:1:cartesian}
%g_{1} \!=\!\! \begin{bmatrix} \cos\theta_{i} \\ \sin\theta_{i} \\ 0 \\ -1 \\ 0 \\ 0 \end{bmatrix} \!\!, \, 
%g_{2} \!=\!\! \begin{bmatrix} 0 \\ 0 \\ 1 \\ y_{ji} \\ -x_{ji} \\ -1 \end{bmatrix} \!\!, \, 
%g_{3} \!=\!\! \begin{bmatrix} 0 \\ 0 \\ 0 \\ \cos\theta_{ji} \\ \sin\theta_{ji} \\ 0 \end{bmatrix} \!\!, \, 
%g_{4} \!=\!\! \begin{bmatrix} 0 \\ 0 \\ 0 \\ 0 \\ 0 \\ 1 \end{bmatrix} 
g_{1}^{\rm r} \!=\!\! \begin{bmatrix} c(\theta_{i}) \\ s(\theta_{i}) \\ 0 \\ -1 \\ 0 \\ 0 \end{bmatrix} \!\!, \, 
g_{2}^{\rm r} \!=\!\! \begin{bmatrix} 0 \\ 0 \\ 1 \\ y_{ji} \\ -x_{ji} \\ -1 \end{bmatrix} \!\!, \, 
g_{3}^{\rm r} \!=\!\! \begin{bmatrix} 0 \\ 0 \\ 0 \\ c(\theta_{ji}) \\ s(\theta_{ji}) \\ 0 \end{bmatrix} \!\!, \, 
g_{4}^{\rm r} \!=\!\! \begin{bmatrix} \ 0 \ \, \\ 0 \\ 0 \\ 0 \\ 0 \\ 1 \end{bmatrix} 
\end{equation}
for $X_{i}=[x_{i}, y_{i}, \theta_{i}, x_{ji}, y_{ji}, \theta_{ji}]^\top$, that is, the relative pose of the rectangular coordinates.
or $g_k^{\rm p}(\cdot)$ defined as
\begin{equation}\label{eq:case:1:polar}
%g_{1} \!=\!\! \begin{bmatrix} \cos\theta_{i} \\ \sin\theta_{i} \\ 0 \\ -\cos\beta_{ji} \\ -\frac{\sin\beta_{ji}}{\rho_{ji}} \\ 0 \end{bmatrix} \!\!, \,
%g_{2} \!=\!\! \begin{bmatrix} 0 \\ 0 \\ 1 \\ 0 \\ 0 \\ -1 \end{bmatrix} \!\!, \,
%g_{3} \!=\!\! \begin{bmatrix} 0 \\ 0 \\ 0 \\ \cos(\theta_{ji}\!-\!\beta_{ji}) \\ \frac{\sin(\theta_{ji}-\beta_{ji})}{\rho_{ji}} \\ 0 \end{bmatrix} \!\!, \,
%g_{4} \!=\!\! \begin{bmatrix} 0 \\ 0 \\ 0 \\ 0 \\ 0 \\ 1 \end{bmatrix} 
g_{1}^{\rm p} \!=\!\! \begin{bmatrix} c(\theta_{i}) \\ s(\theta_{i}) \\ 0 \\ - c(\beta_{ji}) \\ -\frac{s(\beta_{ji})}{\rho_{ji}} \\ 0 \end{bmatrix} \!\!, \,
g_{2}^{\rm p} \!=\!\! \begin{bmatrix} 0 \\ 0 \\ 1 \\ 0 \\ 0 \\ -1 \end{bmatrix} \!\!, \,
g_{3}^{\rm p} \!=\!\! \begin{bmatrix} 0 \\ 0 \\ 0 \\ c(\psi_{ji}) \\ \frac{s(\psi_{ji})}{\rho_{ji}} \\ 0 \end{bmatrix} \!\!, \,
g_{4}^{\rm p} \!=\!\! \begin{bmatrix}  \ 0 \ \, \\ 0 \\ 0 \\ 0 \\ 0 \\ 1 \end{bmatrix} 
\end{equation}
where $X_{i}=[x_{i}, y_{i}, \theta_{i}, \rho_{ji}, \beta_{ji}, \theta_{ji}]^\top$. Here, $\psi_{ji}=\theta_{ji}-\beta_{ji}$, that is, the relative pose of the polar coordinates.
The process noise $W_{i}$ is assumed to be a Gaussian random process for Kalman filtering-based state estimation.

%\item
\noindent
\textbf{[Case M-II] Without communicating odometry data}
Define the state vector $X_{i}=[x_{i}, y_{i}, \theta_{i}, x_{ji}, y_{ji}, \theta_{ji}, v_{j}, \omega_{j}]^\top$ or $X_{i}=[x_{i}, y_{i}, \theta_{i}, \rho_{ji}, \beta_{ji}, \theta_{ji}, v_{j}, \omega_{j}]^\top$ with the odometry vector $u_{i}=[v_{i}, \omega_{i}]^\top$.
\begin{equation}\label{eq:case:2}
%\frac{d}{dt} X_{i} 
\dot{X}_{i}
= \sum_{k=1}^{2} a_{k}(X_i) + \sum_{k=1}^{2} b_{k}(X_i) u_{ik} + W_i
\end{equation}
where $a_k(\cdot)$ and $b_{k}(\cdot)$ are either $a_k^{\rm r}(\cdot)$ or $b_k^{\rm r}(\cdot)$ defined asfollows:
\begin{equation}\label{eq:case:2:cartesian}
%a_{1} \!=\!\! \begin{bmatrix} 0 \\ 0 \\ 0 \\ v_{j}\cos\theta_{ji} \\ v_{j}\sin\theta_{ji} \\ 0 \\ 0 \\ 0 \end{bmatrix} \!\!, \,
%a_{2} \!=\!\! \begin{bmatrix} 0 \\ 0 \\ 0 \\ 0 \\ 0 \\ \omega_{j} \\ 0 \\ 0 \end{bmatrix} \!\!, \,
%b_{1} \!=\!\! \begin{bmatrix} \cos\theta_{i} \\ \sin\theta_{i} \\ 0 \\ -1 \\ 0 \\ 0 \\ 0 \\ 0 \end{bmatrix} \!\!, \,
%b_{2} \!=\!\! \begin{bmatrix} 0 \\ 0 \\ 1 \\ y_{ji} \\ -x_{ji} \\ -1 \\ 0 \\ 0 \end{bmatrix} 
a_{1}^{\rm r} \!=\!\! \begin{bmatrix} 0 \\ 0 \\ 0 \\ v_{j}c(\theta_{ji}) \\ v_{j}s(\theta_{ji}) \\ 0 \\ 0 \\ 0 \end{bmatrix} \!\!, \,
a_{2}^{\rm r} \!=\!\! \begin{bmatrix} 0 \\ 0 \\ 0 \\ 0 \\ 0 \\ \omega_{j} \\ 0 \\ 0 \end{bmatrix} \!\!, \,
b_{1}^{\rm r} \!=\!\! \begin{bmatrix} c(\theta_{i}) \\ s(\theta_{i}) \\ 0 \\ -1 \\ 0 \\ 0 \\ 0 \\ 0 \end{bmatrix} \!\!, \,
b_{2}^{\rm r} \!=\!\! \begin{bmatrix} 0 \\ 0 \\ 1 \\ y_{ji} \\ -x_{ji} \\ -1 \\ 0 \\ 0 \end{bmatrix} 
\end{equation}
for $X_{i}=[x_{i}, y_{i}, \theta_{i}, x_{ji}, y_{ji}, \theta_{ji}, v_{j}, \omega_{j}]^\top$, that is, the relative pose of the rectangular coordinates.
or $a_k^{\rm p}(\cdot)$ and $b_k^{\rm p}(\cdot)$ defined as
\begin{equation}\label{eq:case:2:polar}\!\!\!
%a_{1} \!=\!\! \begin{bmatrix} 0 \\ 0 \\ 0 \\ v_{j}\!\cos(\theta_{ji}\!-\!\beta_{ji}) \\ v_{j}\!\frac{\sin(\theta_{ji}-\beta_{ji})}{\rho_{ji}} \\ 0 \\0 \\ 0\end{bmatrix} \!\!, \,
%a_{2} \!=\!\! \begin{bmatrix} 0 \\ 0 \\ 0 \\ 0 \\ 0 \\ \omega_{j} \\ 0 \\ 0 \end{bmatrix} \!\!, \, 
%b_{1} \!=\!\! \begin{bmatrix} \cos\theta_{i} \\ \sin\theta_{i} \\ 0 \\ -\cos\beta_{ji} \\ -\frac{\sin\beta_{ji}}{\rho_{ji}} \\ 0 \\0 \\ 0 \end{bmatrix} \!\!, \,
%b_{2} \!=\!\! \begin{bmatrix} 0 \\ 0 \\ 1 \\ 0 \\ 0 \\ -1 \\ 0 \\ 0 \end{bmatrix}
a_{1}^{\rm p} \!=\!\! \begin{bmatrix} 0 \\ 0 \\ 0 \\ v_{j}c(\psi_{ji}) \\ v_{j}\!\frac{s(\psi_{ji})}{\rho_{ji}} \\ 0 \\0 \\ 0\end{bmatrix} \!\!, \,
a_{2}^{\rm p} \!=\!\! \begin{bmatrix} 0 \\ 0 \\ 0 \\ 0 \\ 0 \\ \omega_{j} \\ 0 \\ 0 \end{bmatrix} \!\!, \, 
b_{1}^{\rm p} \!=\!\! \begin{bmatrix} c(\theta_{i}) \\ s(\theta_{i}) \\ 0 \\ - c(\beta_{ji}) \\ -\frac{s(\beta_{ji})}{\rho_{ji}} \\ 0 \\0 \\ 0 \end{bmatrix} \!\!, \,
b_{2}^{\rm p} \!=\!\! \begin{bmatrix} 0 \\ 0 \\ 1 \\ 0 \\ 0 \\ -1 \\ 0 \\ 0 \end{bmatrix}
\end{equation}
for $X_{i}=[x_{i}, y_{i}, \theta_{i}, \rho_{ji}, \beta_{ji}, \theta_{ji}, v_{j}, \omega_{j}]^\top$, that is, the relative pose of the polar coordinates.
%Here, $\psi_{ji}=\theta_{ji}-\beta_{ji}$.
The process noise $W_{i}$ is assumed to follow a Gaussian random process, a common assumption in Kalman filtering-based state estimation. Furthermore, we posit that the kinematics of a neighboring robot are unknown to the ego robot in the context of relative pose estimation. Brownian motion is employed as the motion model, wherein the time derivatives of the linear and angular velocities are considered as white Gaussian noise. 

\begin{remark}\label{remark:processnoise}
{\it
More rigorously, the odometry noise and process noise (or disturbance) can be separately modeled as the following generalization of~\eqref{eq:case:1} and \eqref{eq:case:2}:
\begin{equation}\label{eq:case:general}
\dot{X}_{i} = a(X_{i}) + b(X_{i})(U_{i} +W^{u}_{i}) + W^{d}_{i} 
\end{equation}
Here, $W^{u}_{i}$ denotes the odometry noise and $W^{d}_{i}$ corresponds to the combination of model uncertainty and external disturbances.
In this study, the noise is modeled as lumped noise $W_{i} := b(X_{i})W^{u}_{i} + W^{d}_{i}$. However, the separated noise model described above can also be applied for state estimation using Kalman filtering (KF) and pose-graph optimization (PGO). 
}
\end{remark}

%-----------------------------------------------------------------------------%
\subsection{Observation model}
\label{sec:cooploc:sensor}
%-----------------------------------------------------------------------------%
Three types of sensing and communication information are available for state estimation:
\begin{itemize}
	\item
	proprioceptive sensors (encoders, etc.);
	\item
	exteroceptive sensors (LiDAR, Camera, UWB, etc.);
	\item
	communication network (V2X).
\end{itemize}
We consider the following measurement or sensor model:
\[
Y_{i} = h_{i}(X_i) + V_{i}
\]
where $h_{i}(\cdot)$ defines the relationship between the unknown state and the measurements, and $V_{i}$ is the measurement noise. 

\subsubsection{Measurements of absolute pose using landmark observation model}
The measurement model relates the current (absolute) pose of the ego robot to its
LiDAR range and bearing measurements \([ r_{k(i)}^{l}, \phi_{k(i)}^{l} ]^\top\) for $k \in {\mathcal N}_{i}^{l}$
\begin{equation}\label{eq:meas_model:absolute}
\begin{split}
Y_{ki}^{l}
& \!=\! h_{ki}^{l}(X_{i},X_{k}^{l}) + V_{ki}^{l} \\
& \!=\!\!
\begin{bmatrix}
\!
\sqrt{(x_{k}^{l} \!-\! x_{i} \!-\! d_{i}c\theta_{i})^2 \!+\! (y_{k}^{l} \!-\! y_{i} \!-\! d_{i}s\theta_{i})^2}
\! 
\\
\!
{\tt atan2}\!\left(y_{k}^{l} \!-\! y_{i} \!-\! d_{i}s\theta_{i}, x_{k}^{l} \!-\! x_{i} \!-\! d_{i}c\theta_{i}\right) \!-\! \theta_{i}
\!
\end{bmatrix}
%\begin{bmatrix}
%\sqrt{(x_{k}^l - x_{i} - d_{i}\cos\theta_{i})^2 + (y_{k}^l - y_{i} - d_{i}\sin\theta_{i})^2} \\
%{\tt atan2}\left(y_l - y_k - d\sin\theta_{k},x_l - x_k - d\cos\theta_{k}\right) - \theta_k
%\end{bmatrix}
\!\!+\!
V_{ki}^{l}
\end{split}
\end{equation}
where \(x_{k}^{l}\) and \(y_{k}^{l}\) are the ground truth coordinates of the landmark \(k\) that can be observed by Robot ${\tt R}_{i}$, \(x_{i}\) and \(y_{i}\) and \(\theta_{i}\) represent the current pose of Robot ${\tt R}_{i}$, and \(d_{i}\) is the known distance between robot center and laser rangefinder (LiDAR). For brevity, we use $c\theta_{i}=\cos\theta_{i}$ and $s\theta_{i}=\sin\theta_{i}$.
The landmark measurement noise \(V_{ik}^{l}\) is assumed to be a Gaussian random process for Kalman filtering-based state estimation. 

\subsubsection{Measurements of relative pose using range, bearing, and orientation observation models of neighboring robots}
The measurement model relates the current relative pose of robot ${\tt R}_{j}$ with respect to robot ${\tt R}_{i}$, which is denoted as ${}^{i}\xi_{j}$, to the range, bearing, and orientation measurements \([ \rho_{ji}, \beta_{ji}, \theta_{ji}]^\top\): 
\begin{equation}\label{eq:meas:relativepose}
Y_{ji}^{rbo} = h_{ji}^{rbo}(X_{i}) + V_{ji}^{rbo} = \begin{bmatrix} Y_{ji}^{r} \\ Y_{ji}^{b} \\ Y_{ji}^{o} \end{bmatrix} + V_{ji}^{rbo} \\
\end{equation}
where $Y_{ji}^{r} = h_{ji}^{\rm r}(X_{i}) = \rho_{ji}$, $Y_{ij}^{b} = h_{ji}^{\rm b}(X_{i}) = \beta_{ji}$, $Y_{ij}^{o} =  h_{ji}^{\rm o}(X_{i}) = \theta_{ji}$. The measurement function can be rewritten as follows:
$\rho_{ji} = \sqrt{x_{ji}^2 + y_{ji}^2} = \sqrt{(x_{j}-x_{i})^2+(y_{j}-y_{i})^2}$ and $\beta_{ji} = {\tt atan2}(y_{ji},x_{ji}) = {\tt atan2}(y_{j}-y_{i},x_{j}-x_{i})-\theta_{i}$ in Cartesian coordinates. 
The measurement noise \(V_{ji}^{rbo}\) is assumed to be a Gaussian random process for Kalman filtering-based state estimation. 

\begin{remark}
{\it
The measurement variables that relate the current relative pose of the Robot ${\tt R}_{i}$ with respect to the Robot ${\tt R}_{j}$ satisfy the relations $\rho_{ij} = \rho_{ji}$, $\beta_{ij} = {\tt atan2}(\sin\psi_{ji},\cos\psi_{ji})$,  and $\theta_{ij} = - \theta_{ji}$, where $\psi_{ji}=\theta_{ji}-\beta_{ji}$.
}
\end{remark}

\subsubsection{Communication of odometry data}
The wheel-odometry data \(u_{j}=[v_{j},\omega_{j}]^\top\) of a neighboring robot ${\tt R}_{j}$ are not directly measured but can be transmitted via a communication network.
\begin{equation}
Y_{ji}^{od} = h_{ji}^{\rm od}(u_{j}) =  \begin{bmatrix} v_{j} \\ \omega_{j} \end{bmatrix} + V_{ji}^{od}
\end{equation}
where $V_{ji}^{od}$ corresponds to a combination of communication and measurement noise, which is also assumed to be a Gaussian random process for Kalman filtering-based state estimation.

%-----------------------------------------------------------------------------%
\subsection{Message passing model}
%\subsection{Message exchange model}
%\subsection{Communication model}
\label{sec:cooploc:comm}
%-----------------------------------------------------------------------------%
Let $h_{ji}^{\rm mp}(t)$ denote the message passing from robot ${\tt R}_{j}$ to robot ${\tt R}_{i}$ at time $t$.
Assume that this message (set) $m_{ji}(\cdot)$ belongs to the power set $2^{{\mathcal M}_{ji}}$ with the set ${\mathcal M}_{ji}=\{x_{j}, y_{j}, \theta_{j}, \rho_{ij}, \beta_{ij}, \theta_{ij}, v_{j}, \omega_{j} \} = X_{j} \cup U_{j}$.
For the robot state estimation, this message provides additional information as a model. 
\begin{equation}
Y_{ji}^{mp} = h_{ji}^{\rm mp} (X_{j}, U_{j}) + V_{ji}^{mp}
\end{equation}
where $V_{ji}^{\rm mp}$ is the communication noise or uncertainty, which can also be assumed to be a Gaussian random process.

%-----------------------------------------------------------------------------%
\subsection{Extended Kalman filtering for multi-robot localization}
\label{sec:cooploc:ekf}
%-----------------------------------------------------------------------------%
%\check{X} \ \hat{X}
%\begin{itemize}
%	\item
%	\cite{roumeliotis2002distributed}: transformed the centralized extended Kalman filter into a decentralized form. A “small” Kalman filter runs distributively on each robot. When no observation occurs between robots, each robot estimates its pose as a single robot. When two robots observe each other, they communicate to exchange states and measurements. The robots calculate the Kalman gain based on the synergy information and update their state in the update phase.
%	\item
%	\cite{melnyk2012cooperative,zhu2021cooperative}: multi robot cooperative VIO using MSCKF~\cite{mourikis2007multi}: the MSCKF framework has been used in recent years for filter-based cooperative multi-robot localization
%	\item
%	\cite{chenchana2018range,jia2022fej}: Distributed MSCKF Incorporating Ranging Information
%	\item
%	\cite{olfati2005distributed,olfati2005consensus,khan2008distributing,olfati2009kalman,cattivelli2010diffusion}: Distributed KF (theoretical and algorithmic developments)
%\end{itemize}

Explorations into extensions of EKF-based single-robot state estimation within multirobot environments, such as distributed EKF-based multirobot localization~\cite{roumeliotis2002distributed,huang2011observability}, and SLAM~\cite{sasaoka2016multi,shreedharan2023dkf}, have been the focus of research. In the work by Roumeliotis et al.\cite{roumeliotis2002distributed}, the centralized EKF was adapted into a decentralized form. In this decentralized setup, a small EKF operated independently on each robot, determining the Kalman gain based on concurrent ego measurements and communicative messages containing the states and measurements of neighboring robots. In the context of multirobot cooperative visual-inertial-odometry (VIO), the MSCKF framework\cite{mourikis2007multi} has been recently applied in filter-based cooperative multirobot localization~\cite{melnyk2012cooperative,zhu2021cooperative}. Distributed MSCKF models incorporating inter-robot ranging information have also been explored~\cite{chenchana2018range,jia2022fej}. Readers are also directed to~\cite{olfati2005distributed,olfati2005consensus,khan2008distributing,olfati2009kalman,cattivelli2010diffusion} for foundational theoretical and algorithmic developments of distributed KF.

For the robotic state estimation, we consider a discrete-time version of the motion model
\begin{equation}
\begin{split}
X_{i}(t) 
&\!=\! f_{i}^{\rm dt} ( \hat{X}_{i}(t-1), U_{i}(t)) \\
&\!=\! a_{i}^{\rm dt}({X}_{i}(t-1)) + b_{i}^{\rm dt}({X}_{i}(t-1)) U_{i}(t) + W_{i}(t)
\end{split}
\end{equation}
and the associated sampled observation model. 
\begin{equation}
Y_{i}(t) = h_{i} (X_{i}(t),t) + V_{i}(t)
\end{equation}
It is crucial to note that the measurements available at time $t$ are usually time-varying, given that the set of landmarks ${\mathcal N}_{i}^{l}(t)$ and the set of neighboring robots ${\mathcal N}_{i}(t)$ are subject to change over time. This temporal variation implies that the observation models in the state estimation explicitly depend on time.

\subsubsection{Decentralized EKF}
Multi-robot localization using the decentralized EKF framework presented in~\cite{roumeliotis2002distributed} can be summarized as follows: \\[1mm]
$\triangleright$ \underline{Prediction}
\begin{equation}
\begin{split}
\check{X}_{i}(t) &= f_{i}^{\rm dt} ( \hat{X}_{i}(t-1), U_{i}(t) ) \\
\check{P}_{i}(t) &= F_{i}(t-1) \hat{P}_{i}(t-1) F_{i}^{\top}\!(t-1) + {\Sigma}_{w,i}(t) \\
\check{Y}_{i}(t) &= h_{i} (\check{X}_{i}(t),t)
\end{split}
\end{equation}
where $F_{i} = \partial f_{i}^{\rm dt} / \partial X_i$ denotes the Jacobian matrix of the vector field $f_{i}^{\rm dt}$ evaluated at $(\hat{X}_{i}(t-1), U_{i}(t))$. 
$\hat{P}_{i}(t-1)$ is the error covariance matrix of the posterior probability of the state at the previous time step and ${\Sigma}_{w,i}(t)$ is the second moment of the probabilistic disturbance $W_i(t)$. The observation model $h_{i}$ relates both robot ${\tt R}_{i}$'s onboard sensor measurements and messages passed from neighbors $\mathcal{N}_i(t)$ to robot ${\tt R}_{i}$'s state vector. \\[1mm]
\noindent
$\triangleright$ \underline{Computation of Kalman gains}
\begin{equation}
K_{i}(t) = \check{P}_{i}(t) H_{i}^{\top}\!(t) \!\left( H_{i}(t) \check{P}_{i}(t) H_{i}^{\top}\!(t) + \Sigma_{v,i}(t) \right)^{-1}
\end{equation}
where $H_{i} = \partial h_{i}/ \partial X_i$ is the Jacobian matrix of the measurement function $h_{i}$ evaluated at $\check{X}_{i}(t)$ which is the predicted mean of the state, and ${\Sigma}_{v,i}(t)$ is the second moment of random process noise $V_i(t)$. \\[1mm]
\noindent
$\triangleright$ \underline{Correction}
\begin{equation}
\begin{split}
\hat{X}_{i}(t) &= \check{X}_{i}(t) + K_{i}(t)(Y_{i}(t)-\check{Y}_{i}(t)) \\
\hat{P}_{i}(t) &= (I - K_{i}(t) H_{i}(t)) \check{P}_{i}(t) %{\tt Ricc}(\check{P}_{i}(t))
\end{split}
\end{equation}
where $Y_{i}(t)$ denotes a noisy measurement vector encompassing both the onboard sensing and communication information.
The resulting probabilistic inference of the state estimate is assumed to follow a Gaussian random process.
\begin{equation}
X_{i}(t) \sim {\mathcal N}(\hat{X}_{i}(t),\hat{P}_{i}(t)) \,.
\end{equation}

\subsubsection{Distributed EKF}
In a manner distinct from the centralized EKF, the distributed EKF takes into account coupled constraints and strives for consensus among robots when there are overlapping state variables of estimation. The consensus Kalman filtering, as presented in~\cite{olfati2005distributed,olfati2005consensus}, can be summarized as follows: \\[1mm]

\noindent
$\triangleright$ \underline{Prediction-consensus}
\begin{equation}\label{eq:ckf:pred}
\begin{split}
{\hat{X}}_{i}(t) 
& = \check{X}_{i}(t) + K_{i}(t)(\check{Y}_{i}(t) - Y_{i}(t)) \\
& \quad +\!\!\!\! \sum_{j\in{\mathcal N}_{i}\!(t)}\!\!\!\! \check{K}_{ji}(t) \! \left( C_{ji} \check{X}_{i}(t) -  C_{ij} \check{X}_{j}(t) \right)
\end{split}
\end{equation}
where $K_{i}(t)$ denotes the Kalman gain corresponding to the on-board sensing and $\check{K}_{ji}(t)$ is the distributed Kalman gain corresponding to the consensus constraints of \emph{predicted} relative pose, $\check{\rho}_{ij}(t) = \check{\rho}_{ji}(t)$, $\check{\beta}_{ij}(t)=\check{\theta}_{ji}(t)-\check{\beta}_{ji}(t)$, and $\check{\theta}_{ij}(t) = -\check{\theta}_{ji}(t)$, in the polar coordinates. Matrices $C_{ji}$ and $C_{ij}$ are defined by following the consensus constraints of the relative poses $X_{ji}$ and $X_{ij}$. \\[1mm]
\noindent
$\triangleright$ \underline{Correction-consensus}
\begin{equation}\label{eq:ckf:corr}
\begin{split}
\hat{X}_{i}(t) 
& \leftarrow\! \hat{X}_{i}(t) \!+\!\!\!\!\! \sum_{j\in{\mathcal N}_{i}\!(t)}\!\!\!\!\! \hat{K}_{ji} (t) \!\!\left(\! C_{ji} \hat{X}_{i}(t) \!-\!  C_{ij} \hat{X}_{j}(t)\!\right)
\end{split}
\end{equation}
where $\hat{K}_{ji}(t)$ is the distributed Kalman gain corresponding to the consensus constraints of \emph{corrected} relative pose, $\hat{\rho}_{ij}(t) = \hat{\rho}_{ji}(t)$, $\hat{\beta}_{ij}(t)=\hat{\theta}_{ji}(t)-\hat{\beta}_{ji}(t)$, and $\hat{\theta}_{ij}(t) = -\hat{\theta}_{ji}(t)$, in the polar coordinates. The updated equation~\eqref{eq:ckf:corr}~ is indeed iterative, whereas the updated equation in~\eqref{eq:ckf:pred}~does not necessitate iterations. This also implies that one must consider the convergence analysis of the iteration~\eqref{eq:ckf:corr}, which is linear, so that an eigenvalue or spectral analysis can be applied to investigate the convergence~\cite{boyd2006randomized}.

%-----------------------------------------------------------------------------%
\subsection{Optimization-based state estimation}
\label{sec:cooploc:opt}
%-----------------------------------------------------------------------------%
\begin{table}[t!] \tiny\small
\begin{center}
%\begin{tabular}{ | m{7em} | m{6em}| m{6em} | }
\begin{tabular}{ |@{\,\,} m{4em} @{\,\,\,\,\,} |@{\,\,}  m{11.55em} @{\,\,} |@{\,\,} m{9.5em} | }
  \hline\vspace{1mm}
  \textbf{Function} & \vspace{1mm} \textbf{Loss $\ell(e)$} & \vspace{1mm} \textbf{Weight $\gamma(e)$} \\[.75mm] 
  \hline\hline 
  \vspace{1mm}$L_2$ & \vspace{1mm}$\frac{e^2}{2}$ & \vspace{1mm}$1$ \\[1.5mm] 
  \hline 
  \vspace{1mm}Laplace  &\vspace{1mm}$t |e|$  &\vspace{1mm} $\frac{t}{|e|}$ \\[1.5mm]  
  \hline \vspace{4mm}
  Huber & $\left\{\begin{array}{@{}l@{\,\,\,\,}l} \frac{e^2}{2} &\mbox{for } |e| \!\leq\! t \\[0.5mm] t(|e|-t/2) & \mbox{o.w.}\end{array}\right.$ 
  		& $\left\{\begin{array}{@{}l@{\,\,\,\,}l} 1 &\mbox{for } |e| \!\leq\! t \\[0.5mm] \frac{1}{|e|} & \mbox{o.w.} \end{array}\right.$ \\[4.5mm]  
  \hline 
  \vspace{1mm} Cauchy &\vspace{1mm} $\frac{t^2}{2} \ln (1+\frac{e^2}{t^2})$ &\vspace{1mm} $\frac{t^2}{t^2+e^2}$ \\[1.75mm]  
  \hline \vspace{2mm}
   Fair  & $t^2 \!\left( \frac{t}{|e|} - \ln(1+\frac{t}{|e|}) \!\right)$  & $\frac{t}{t+|e|}$ \\[2mm] 
  \hline \vspace{1mm}
   Geman-McClure  & $\frac{e^2}{2(t+e^2)}$ & $\frac{t^2}{(t+e^2)^2}$ \\[1mm]  
  \hline 
  \vspace{1mm} Welsch  &\vspace{1mm} $\frac{t^2}{2}(1-\exp(-\frac{e^2}{t^2}))$ &\vspace{1mm} $\exp(-\frac{e^2}{t^2})$ \\[1.75mm]   
  \hline \vspace{2mm}
  Switchable-Constraint  & $\left\{\begin{array}{@{}l@{\,\,\,\,}l} \frac{e^2}{2} &\mbox{for } e^2\!\leq\! t \\[0.5mm] \frac{2te^2}{t+e^2} - \frac{t}{2} & \mbox{o.w.}\end{array}\right.$ 
  		& $\left\{\begin{array}{@{}l@{\,\,\,\,}l} 1 &\mbox{for } e^2\!\leq\! t \\[0.5mm] \frac{4t^2}{(t+e^2)^2} & \mbox{o.w.} \end{array}\right.$ \\[3mm] 
  \hline
  \vspace{4mm}Tukey  & $\!\!\left\{\!\begin{array}{@{}l@{\,\,}l} \frac{t^2(1-(1-\frac{e^2}{t^2})^3)}{2} & \mbox{for } |e|\!\leq\! t \\[0.5mm] \frac{t^2}{2} & \mbox{o.w.}\end{array}\right.$ 
  		&\vspace{1mm} $\left\{\begin{array}{@{}l@{\,\,\,\,}l} (1-\frac{e^2}{t^2})^2 &\mbox{for } |e|\!\leq\! t \\[1.mm] 0 & \mbox{o.w.} \end{array}\right.$ \\[6mm]   
  \hline \vspace{4mm}
  Max.Dist.  & $\left\{\begin{array}{@{}l@{\,\,\,\,}l} \frac{e^2}{2} &\mbox{for } |e|\!\leq\! t \\[0.5mm] \frac{t^2}{2} & \mbox{o.w.}\end{array}\right.$ 
  		& $\left\{\begin{array}{@{}l@{\,\,\,\,}l} 1 &\mbox{for } |e|\!\leq\! t \\[0.5mm] 0 & \mbox{o.w.} \end{array}\right.$ \\[5mm]    
  \hline
\end{tabular}\\[2mm]
\caption{A list of robust loss functions for M-estimation\cite{bosse2016robust,barron2019general}.}
\label{tab:1}
\end{center}
\end{table}

\subsubsection{Nonlinear least squares methods}
Consider online optimization methods to estimate the state variables of a nonlinear Markov process in the following form:
\begin{equation}\label{eq:nlsys}
\begin{split}
X_{i}(t) &= f_{i}(X(t\!-\!1)) + W_{i}(t) \,, \\
Y_{i}(t)     &= h_{i}(X(t)) + V_{i}(t)
\end{split}
\end{equation}
where $X_{i} \in {\mathbb R}^n$ refers to the state of the system 
$Y_{i} \in {\mathbb R}^m$ denotes measured output.
$W_{i}\in {\mathbb R}^n$ denotes the disturbance and
$V_{i}\in {\mathbb R}^m$ denotes measurement noise.
Here, the odometry-dependence of $f_{i}$ is hidden for simplicity. 
We further assume that the state, disturbance, and noise belong to compact convex constraint sets:
${\mathbb X}_{i}({t}) \subseteq {\mathbb R}^{n}$,
${\mathbb W}_{i}({t}) \subseteq {\mathbb R}^{n}$,
and ${\mathbb V}_{i}({t}) \subseteq {\mathbb R}^{m}$, respectively, which can be represented as (linear or quadratic) inequalities.  

The estimation problem with full information on $T$ measurement sequences $\{Y_{i}(1), \cdots, Y_{i}(T)\}$ is represented by the following nonlinear programming (NLP):
\begin{mini}|l|
	{\{ X_{i},W_{i},V_{i}\}}{ \!\!\!\! q_0(X_{i}(0)) \!+ \!\sum_{t=1}^{T} q(X_{i}(t\!-\!1), X(t))\,\,\,}
	{}{}
	\label{eq:full}
	\addConstraint{\!\!\!\!\!\!\! X_{i}(t) = f_{i}(X_{i}(t\!-\!1)) + W_{i}(t)} 
	\addConstraint{\!\!\!\!\!\!\! Y_{i}(t) = h_{i}(X_{i}(t)) + V_{i}(t)},
	\addConstraint{\!\!\!\!\!\!\! X_{i}(t) \!\in\! {\mathbb X}_{i}({t}) , W_{i}(t) \!\in\! {\mathbb W}_{i}({t}), V_{i}(t) \!\in\! {\mathbb V}_{i}({t})}
\end{mini}
where $T$ refers to the current time step and the stage loss function is defined as
\begin{equation}\label{eq:lossfn}
\begin{split}
& q(X_{i}(t\!-\!1), X_{i}(t)) \\
& \ = (X_{i}(t) \!-\! f_{i}(X_{i}(t\!-\!1)))^{\! \top} \! \Psi (X_{i}(t) \!-\!  f_{i}(X_{i}(t\!-\!1)))  \\
& \ \ \ \ \ + (Y_{i}(t)\!-\!h_{i}(X_{i}(t)))^{\! \top} \! \Phi (Y_{i}(t)\!-\!h_{i}(X_{i}(t))) \\
& \ = W_{i}^{\top}\!(t) \Psi W_{i}(t) + V_{i}^{\top}\!(t) \Phi V_{i}(t) \\
& \ = q_{w}(W_{i}(t)) + q_{v}(V_{i}(t))
\end{split}
\end{equation}
and $q_0(\cdot)$ denotes the arrival cost function of the initial state $X_{i}(0) \in {\mathbb R}^{n}$.
The arrival cost function $q_0 : {\mathbb X}_{i}(0) \rightarrow \mathbb R$ is used to summarize the prior information or statistics regarding the state at time $t=0$,
and it is assumed to satisfy $q_0(\bar X_{i,0}) = 0$ and $q_0(X_{i,0}) >0$ for all $X_{i,0} \neq \bar X_{i,0}$ where $\bar X_{i,0}$ denotes the best state estimate at time $t=0$. An example of the arrival cost $q_0(\cdot)$ is the quadratic form $q_0(X_{i,0}) = (X_{i,0} \!-\! \bar X_{i,0})^{\!\top}\! P_{i,0}^{-1}(X_{i} - \bar X_{i})$ where $P_{i,0}$ is a positive definite matrix corresponding to the error covariance of the initial state $X_{i}(0) \sim {\mathcal{GP}}(\bar X_{i,0},P_{i,0})$-R@@

%In a MHE framework, we consider a partial history of $h$ measurement sequences $\{y_{\tau_c-h+1}, \cdots, y_{\tau_c}\}$ with $h \leq \tau_c$.
%The corresponding NLP is given by 
%\begin{mini}|l|
%	{\{x_{t}\}, \{w_{t}\}, \{v_{t}\}}{\!\!\!\!\!\!  \phi_{\tau_c-h}(x_{\tau_c - h}) + \!\!\!\!\! \sum_{t=\tau_c-h+1}^{\tau_c}\!\!\!\!\! \ell_t(x_{t-1}, x_{t})}
%	{}{}
%	\label{eq:mhe}
%	\addConstraint{\!\!\!\!\!\! x_{t} = f(x_{t-1})+ w_{t-1}}
%	\addConstraint{\!\!\!\!\!\! y_{t} = g(x_{t}) + v_{t}},
%	\addConstraint{\!\!\!\!\!\! x_{t} \in {\mathbb X}_{t}, w_{t} \in {\mathbb W}_{t}, v_{t} \in {\mathbb V}_{t}}
%\end{mini}
%where the arrival cost function $\phi_{\tau_c-h} : {\mathbb X}_{\tau_c-h} \rightarrow \mathbb R$ plays a crucial role in the performance of MHE because it has a role of an approximate value function in forward dynamic programming
%for recursive state estimation of dynamical systems. 

\subsubsection{Robust M-estimation methods}
Instead of using quadratic forms for the loss functions $q_0(\cdot)$ and $q(\cdot, \cdot)$, robust loss functions for $ M$ estimation can also be used.
\begin{equation}
\begin{split}
\tilde{q}_{0}(X_{i,0})         &= \ell \!\left(\! \sqrt{{q}_{0}(X_{i,0})} \right) , \\
\tilde{q}(W_{i}(t), V_{i}(t)) &= \ell \!\left(\! \sqrt{{q}_{w}(W_{i}(t))} \right) + \ell \!\left(\! \sqrt{{q}_{v}(V_{i}(t))} \right) , \\
\end{split}
\end{equation}
where the function $\ell : {\mathbb R}_{+} \rightarrow  {\mathbb R}_{+}$ is selected as the loss function as shown in Table~\ref{tab:1}.
Owing to the nonconvexity of the {\em robust} loss functions listed in Table~\ref{tab:1}, the resulting nonlinear least-squares problem should be solved iteratively. The iteratively reweighted least-squares (IRLS) method~\cite{bergstrom2014robust}can be used to approximate the loss function $\ell(e(X))$:
\begin{equation}
\ell(e(X)) \approx \gamma(e(\check{X})) e^2(X)
\end{equation}
and iteratively updated as follows: 
\begin{equation}
\hat{X} \leftarrow \arg\min_{X\in {\mathbb X}} \sum_{t=1}^{T} \gamma(e(\check{X}(t))) e^2(X(t)) \ \  \mbox{and} \ \ \check{X} \leftarrow \hat{X}
\end{equation}
where $\check{X}=(\check{X}(t))_{t=1}^{T}$ and $\hat{X}=(\hat{X}(t))_{t=1}^{T}$ are the previous and current guesses of the estimate, respectively, and the function $\gamma : {\mathbb R}_{+} \rightarrow  {\mathbb R}_{+}$ is the weight associated with the loss function. The relations between several loss and weight functions are also listed in Table~\ref{tab:1}.

\begin{figure}[!t]
	\centering
	\includegraphics[width=.95\linewidth]{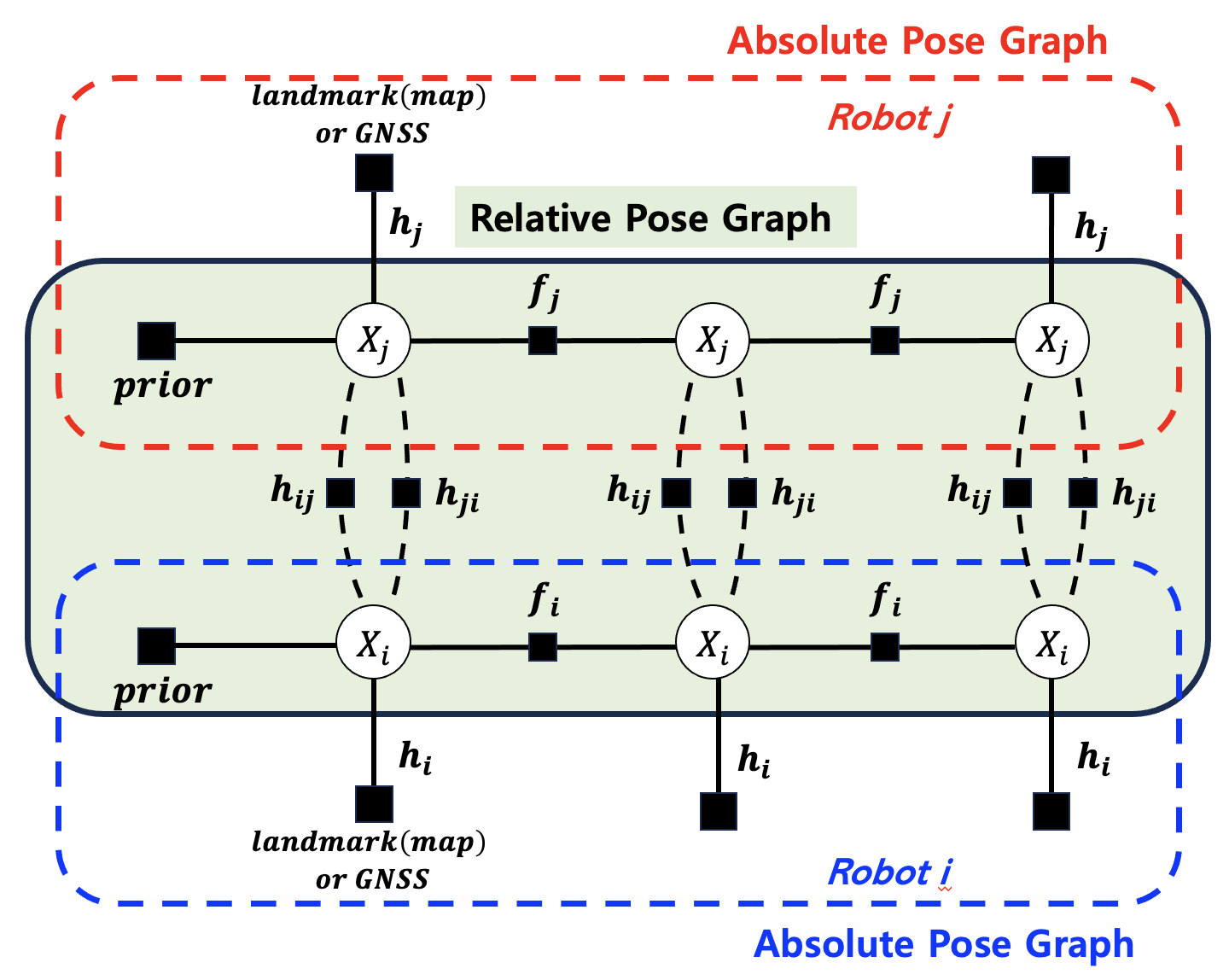}\vspace{-4mm}
	\caption{Factor graph representation of multi-robot absolute and relative pose estimation.}
	\label{fig:fgo1} 
%\end{figure}
\vspace{2mm}
%\begin{figure}[!t]
	\centering
	\includegraphics[width=1.0\linewidth]{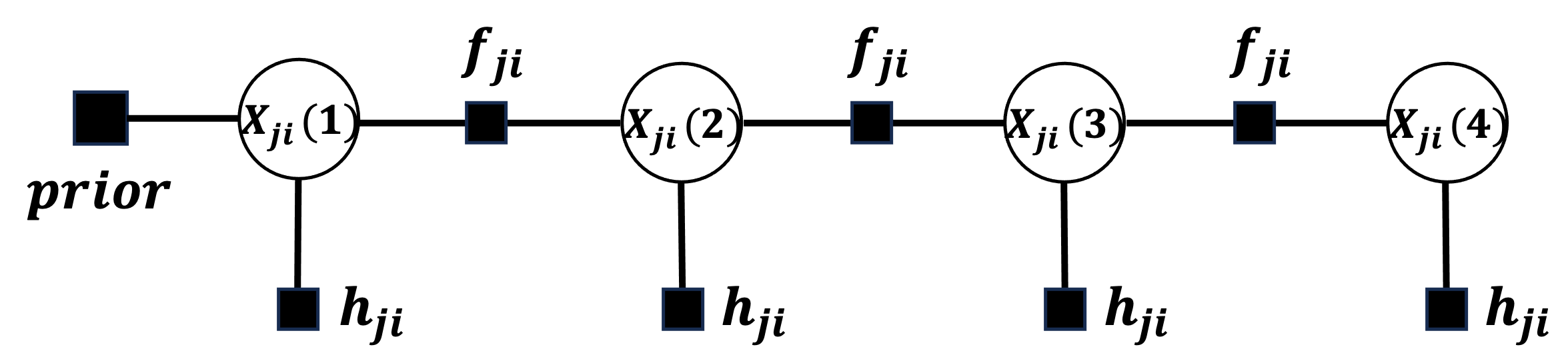}\vspace{-4mm}
	\caption{Factor graph representation of inter-robot relative pose estimation: $f_{ji}$ and $h_{ji}$ refer to the kinematics and measurement functions associated with $X_{ji}$ that is the relative pose of the Robot ${\tt R}_j$ with respect to Robot ${\tt R}_i$.}
	\label{fig:fgo2} 
\end{figure}
%------------------------------------
\subsubsection{Factor graph optimization}
%------------------------------------
Factor graph optimization is widely employed in optimization-based robot state estimation~\cite{grisetti2010tutorial,dellaert2017factor,dellaert2021factor}, particularly in Pose Graph Optimization (PGO) as a back-end for Simultaneous Localization and Mapping (SLAM)~\cite{latif2014robust,mcgann_risam_2023}.
\begin{itemize}
\item
Fig.~\ref{fig:fgo1}~shows a factor graph representation corresponding to multi-robot absolute and relative pose estimation, in which the absolute poses $(X_{i}, X_{j})$, and the relative pose between the robots $(X_{ji}, X_{ij})$) should be estimated by using the relations between inter-time kinematics $(f_{i}, f_{j})$, ego-pose measurements $(h_{i}, h_{j})$, and inter-robot relative measurements $(h_{ji}, h_{ij})$)  
\item
Inter-robot relative pose estimation (shaded by green in Fig.~\ref{fig:fgo1}) can be equivalently represented in the factor graph of Fig.~\ref{fig:fgo2}.
\end{itemize}

For graph optimization for distributed multi-robot localization, consider the three types of error functions: The function associated with the state-transition (forward-dynamics) factor is defined as
\begin{equation}
e_{i}^{f} (X_i,X_{i}^{\prime}) = (X_{i} \!-\! f_{i}(X_{i}^{\prime},U_{i}))^{\!\top} \Sigma_{w,i}^{-1} (X_{i} \!-\! f_{i}(X_{i}^{\prime},U_{i})), 
\end{equation}
The function associated with the map-based absolute position measurements is given by:
\begin{equation}
\begin{split}
& e_{ki}^{l} (X_i,X_{k}^{l}) \\
& \quad = (Y_{ki}^{l} - h_{ki}^{l}(X_{i},X_{k}^{l}))^{\!\top} \Sigma_{l,ki}^{-1} (Y_{ki}^{l} - h_{ki}^{l}(X_{i},X_{k}^{l})),
\end{split}
\end{equation}
The function corresponding to the robot-to-robot measurements and message passing is defined as
\begin{equation}
\begin{split}
&e_{ji}^{r2r} (X_i,X_j) = e_{ji}^{r2r} (X_{ji}) \\
&\quad = (Y_{ji}^{r2r} \!- h_{ji}^{r2r}(X_{ji}))^{\!\top} \Sigma_{r2r,ki}^{-1} (Y_{ji}^{r2r} \!- h_{ji}^{r2r}(X_{ji}))
\end{split}
\end{equation}
where the robot-to-robot measurement vector involves concatenating the range-bearing-orientation measurements and the odometry data for the target or neighboring Robot ${\tt R}_{j}$, denoted as $Y_{ji}^{r2r}=(Y_{ji}^{rbo}, Y_{ji}^{od})^\top$, and the function $h_{ji}^{r2r}(\cdot)$ is defined accordingly.  

The optimization problem is defined as follows:
\begin{equation}
\begin{split}
\min_{(X_i(t))_{t=1}^{T}} \ \ & p(X_{i}(0)) + \sum_{t=1}^{T} e_{i}^{f} (X_i(t),X_{i}(t-1)) \\
& \ \ \ \ + \sum_{t=1}^{T} \sum_{k \in {\mathcal N}_{i}^{l}(t)} \! e_{ki}^{l} (X_{i}(t),X_{k}^{l}) \\
& \ \ \ \ + \sum_{t=1}^{T} \sum_{j \in {\mathcal N}_{i}(t)} \! e_{ji}^{r2r} (X_{ji}(t))
\end{split}
\end{equation}
%\begin{mini}|l|
%	{\{X_{i}\}}{\!\!\!\!\! q_0(X_{i}(0)) \!+ \!\sum_{t=1}^{T} 
%	{}{}
%	\label{eq:full}
%	\addConstraint{\!\!\!\!\!\!\! X_{i}(t) = f_{i}(X_{i}(t\!-\!1)) + W_{i}(t)} 
%	\addConstraint{\!\!\!\!\!\!\! Y_{i}(t) = h_{i}(X_{i}(t)) + V_{i}(t)},
%	\addConstraint{\!\!\!\!\!\!\! X_{i}(t) \!\in\! {\mathbb X}_{i}({t}) , W_{i}(t) \!\in\! {\mathbb W}_{i}({t}), V_{i}(t) \!\in\! {\mathbb V}_{i}({t})}
%\end{mini}
The relative pose estimation problem can be expressed as follows:
\begin{equation}
\begin{split}
\min_{(X_{ji}(t))_{t=1}^{T}}  & p(X_{ji}(0)) +\! \sum_{t=1}^{T} e_{i}^{f} (X_{ji}(t),X_{ji}(t-1)) \\
& \ \ \ \ + \sum_{t=1}^{T}  e_{ji}^{r2r} (X_{ji}(t))
\end{split}
\end{equation}
where $j \in {\mathcal N}_{i}(t)$ is assumed for all $t=1, \ldots, T$.

%\newpage
%==============================================================%
\section{Observability in Relative Pose Estimation}%{Lie Derivatives and Nonlinear Observability}
\label{sec:nonobsanal}
%==============================================================%
This section provides a summary of existing observability analyses for inter-robot planar relative pose estimation in the context of ground mobile robots. Additionally, a novel analysis is presented for scenarios where an inter-robot communication network is unavailable. Given that the considered motion models are nonlinear dynamic systems, the Lie derivative-based observability rank condition is employed. This involves the explicit computation of state-dependent observability spaces for four different information structures, considering both Polar and Cartesian coordinates to represent robot poses in $\text{SE}(2)$. 
%Table~\ref{tab:2} summarizes the observability analysis results provided in this section.

%-----------------------------------------------------------------------------%
\subsection{Observability with global odometry data}
\label{sec:nonobsanal:1}
%-----------------------------------------------------------------------------%
The observability rank condition, based on Lie derivatives, was initially introduced in~\cite{hermann1977nonlinear}. Subsequently, it was applied to the observability analysis of multirobot relative pose estimation~\cite{martinelli2005observability} with inter-robot range-only measurements~\cite{araki2019range} and bearing-only measurements~\cite{sharma2011graph}. Notably, the measurements of the angular speed odometry data $(\omega_{i}, \omega_{j})$ did not change the rank of the observation space. We have rewritten and reinterpreted the results previously presented in~\cite{martinelli2005observability,araki2019range}.

%------------------------------%
\subsubsection{Range-only measurement}
%------------------------------%
We considered two different coordinates of the system representation for observability analysis: Polar and Cartesian coordinates.

%-----------------
\paragraph{Observability analysis in the Polar coordinates} 
%-----------------
For observability analysis of the relative pose estimation between the two mobile robots, we consider the following model equations described in polar coordinates:
\begin{itemize}
\item
Kinematic motion model: $\dot{x} = \sum_{k=1}^{4} g_{k}(x)u_k$ which is a control-linear system given in~\eqref{eq:case:1} and \eqref{eq:case:1:polar} and 
\item
Measurement model: $y = h(x) = \rho_{ji} = x_1$ given in \eqref{eq:meas:relativepose}.
\end{itemize}

By applying an observability analysis method based on Lie derivatives~\cite{hermann1977nonlinear}, we obtain the following sequence of expanding observability subspaces (codistributions):
\begin{equation}\label{eq:obs:rangeonly:polar}
\begin{split}
\mathcal{O}_0 &= {\rm span} \{ \mathcal{L}^{0} h \} = {\rm span} \{ [1,0,0]^\top \} \\
%\mathcal{O}_0 &= {\rm span} \{ \mathcal{L}^{0} h \} = {\rm span} \{ \nabla h \}  = {\rm span} \{ [1,0,0]^\top \} \\
%\mathcal{O}_{k+1} &= \mathcal{O}_{k} + \sum_{\ell=1}^{4} \mathcal{L}_{g_{\ell}} \mathcal{O}_{k} 
\mathcal{O}_1 
&= \mathcal{O}_0 + \sum_{k=1}^{4} {\rm span} \{ \nabla \mathcal{L}_{g_{k}} h \} \\
%&= {\rm span} \{ [1,0,0]^\top, [0,\sin(x_2),0]^\top, [0,\sin(x_{32}), -\sin(x_{32})]^\top \}
&= {\rm span} 
\left\{ 
\left[\begin{array}{@{\,}c@{\,}} 1 \\ 0 \\ 0 \end{array}\right] , 
\left[\begin{array}{@{}c@{}} 0 \\ s(x_2) \\ 0 \end{array}\right] , 
%\left[\begin{array}{@{}c@{}} 0 \\ \sin(x_2) \\ 0 \end{array}\right] , 
\left[\begin{array}{@{\!}c@{\!}} 0 \\ s(x_{32}) \\ -s(x_{32}) \end{array}\right]
%\left[\begin{array}{@{\!}c@{\!}} 0 \\ \sin(x_{32}) \\ -\sin(x_{32}) \end{array}\right]
\right\} \\
\mathcal{O}_{\ell} &= \mathcal{O}_{1} \mbox{ for all $\ell \geq 2$}
\end{split}
\end{equation}
where $x_{32}= x_3 - x_2$ is the angular difference between the orientation and bearing angles of the relative pose, represented by the eco-robot's local coordinates. The rank of observability (sub-)space is given by
\begin{equation}
{\rm rank}\, \mathcal{O}_1 = 3 %\mbox{ unless $x_2=n\pi$ and $x_3=n'\pi$ for $n,n' \in \mathbb{N}$}
\end{equation}
unless $x_2=n\pi$ or $x_2 - x_3=\bar{n}\pi$ for $n,\bar{n} \in \mathbb{N}$.
This implies that the corresponding system is \emph{locally observable} at almost every location.

%-----------------
\paragraph{Observability analysis in the Cartesian coordinates:}
%-----------------
Under the same information structure but with a different coordinate system of pose representations, we consider 
\begin{itemize}
\item
Kinematic motion model: $\dot{x} = \sum_{k=1}^{4} g_{k}(x)u_k$ which is a control-linear system represented in \eqref{eq:case:1} and \eqref{eq:case:1:cartesian} and 
\item
Measurement model: 
%$y = h(x) = \frac{1}{2}(x^2_{ji} + y^2_{ji}) = \frac{1}{2}(x_1^2 + x_2^2)$
$y = h(x) = \frac{1}{2}(x_1^2 + x_2^2)$
This formulation is equivalent to the model represented in \eqref{eq:meas:relativepose}. The introduction of this measurement function serves primarily for mathematical convenience and was considered in \cite{araki2019range} specifically for observability analysis purposes.
\end{itemize}
Similar to the case of polar coordinates, we apply the Lie derivative-based observability analysis and obtain the following sequence of observability codistributions:
\begin{equation}\label{eq:obs:rangeonly:cartesian}
\begin{split}
\mathcal{O}_0 &= {\rm span} \{ \mathcal{L}^{0} h \} = {\rm span} \{ [x_1,x_2,0]^\top \} \\
\mathcal{O}_1 
&= \mathcal{O}_0 + \sum_{k=1}^{4} {\rm span} \{ \nabla \mathcal{L}_{g_{k}} h \} \\
%&= {\rm span} \{ [1,0,0]^\top, [0,\sin(x_2),0]^\top, [0,\sin(x_{32}), -\sin(x_{32})]^\top \}
&= {\rm span} 
\left\{\!
\left[\begin{array}{@{\,}c@{\,}} x_1 \\ x_2 \\ 0 \end{array}\right] \!\!, \!  
\left[\begin{array}{@{}c@{}} -1 \\ 0 \\ 0 \end{array}\right] \!\!, \! 
%\left[\begin{array}{@{\!}c@{\!}} \cos(x_3) \\ \sin(x_{3}) \\ x_2\cos(x_{3})-x_1\sin(x_{3}) \end{array}\right]
\left[\begin{array}{@{\!}c@{\!}} c(x_3) \\ s(x_{3}) \\ x_2 c(x_{3})-x_1 s(x_{3}) \end{array}\right]
\!\right\} \\
\mathcal{O}_{\ell} &= \mathcal{O}_{1} \mbox{ for all $\ell \geq 2$} \,.
\end{split}
\end{equation}
The rank condition is given by
\begin{equation}
{\rm rank}\, \mathcal{O}_1 = 3 %\mbox{ unless $x_2=n\pi$ and $x_3=n'\pi$ for $n,n' \in \mathbb{N}$}
\end{equation}
unless $x_2=0$ or ${\rm atan}(x_2/x_1) - x_3= n\pi$ for $n \in \mathbb{N}$. This coincides with the result obtained from the Cartesian coordinate pose representation with the same information structure.

%------------------------------%
\subsubsection{Bearing-only measurement}
%------------------------------%
Similar to range-only measurements, we consider both Polar and Cartesian coordinates.

%-----------------
\paragraph{Observability analysis in the Polar coordinates:}
%-----------------
First, we consider the following models of relative motion and bearing-only measurements in polar coordinates:
\begin{itemize}
\item
Kinematic motion model: $\dot{x} = \sum_{k=1}^{4} g_{k}(x)u_k$ which is a control-linear system represented in \eqref{eq:case:1} and \eqref{eq:case:1:polar} and 
\item
Measurement model: $y = h(x) = \beta_{ji} = x_2$ represented in \eqref{eq:meas:relativepose}.
\end{itemize}
By applying the method of observability analysis based on Lie derivatives, we obtain a sequence of codistributions
\begin{equation}\label{eq:obs:bearingonly:polar}
\begin{split}
\mathcal{O}_0 &= {\rm span} \{ \mathcal{L}^{0} h \} = {\rm span} \{ [0,1,0]^\top \} \\
%\mathcal{O}_0 &= {\rm span} \{ \mathcal{L}^{0} h \} = {\rm span} \{ \nabla h \}  = {\rm span} \{ [1,0,0]^\top \} \\
%\mathcal{O}_{k+1} &= \mathcal{O}_{k} + \sum_{\ell=1}^{4} \mathcal{L}_{g_{\ell}} \mathcal{O}_{k} 
\mathcal{O}_1 
&= \mathcal{O}_0 + \sum_{k=1}^{4} {\rm span} \{ \nabla \mathcal{L}_{g_{k}} h \} \\
%&= {\rm span} \{ [1,0,0]^\top, [0,\sin(x_2),0]^\top, [0,\sin(x_{32}), -\sin(x_{32})]^\top \}
&= {\rm span} 
\left\{\!
\left[\begin{array}{@{\,}c@{\,}} 0 \\ 1 \\0 \end{array}\right] \!\!, \!
\left[\begin{array}{@{}c@{}} {s(x_2)}/{x_1^2} \\ -{c(x_2)}/{x_1^2} \\ 0 \end{array}\right] \!\!, \! 
\left[\begin{array}{@{\!}c@{\!}} {s(x_{23})}/{x_1^2} \\ -{c(x_{23})}/{x_1} \\ {c(x_{23})}/{x_1} \end{array}\right]
%\left[\begin{array}{@{\,}c@{\,}} 0 \\[3mm] 1 \\[3mm] 0 \end{array}\right] , 
%\left[\begin{array}{@{}c@{}} \frac{s(x_2)}{x_1^2} \\[3mm] -\frac{c(x_2)}{x_1^2} \\[3mm] 0 \end{array}\right] , 
%\left[\begin{array}{@{\!}c@{\!}} \frac{s(x_23)}{x_1^2} \\[2mm] -\frac{c(x_2)}{x_1} \\[2mm] \frac{c(x_2)}{x_1} \end{array}\right]
\!\right\} \\
\mathcal{O}_{\ell} &= \mathcal{O}_{1} \mbox{ for all $\ell \geq 2$}
\end{split}
\end{equation}
where a nontrivial displacement is assumed; that is, $x_1 \neq 0$. 
The rank of observability (sub)space is given by:
\begin{equation}
{\rm rank}\, \mathcal{O}_1 = 3 %\mbox{ unless $x_2=n\pi$ and $x_3=n'\pi$ for $n,n' \in \mathbb{N}$}
\end{equation}
almost every, unless $x_2=n\pi$ or $c(x_2)=c(x_{23})=0$.
Therefore, we conclude that the corresponding system with bearing-only measurements is \emph{locally observable}.

%-----------------
\paragraph{Observability analysis in the Cartesian coordinates:}
%-----------------
Consider the following models of relative motion and bearing-only measurements in Cartesian coordinates:
\begin{itemize}
\item
Kinematic motion model: $\dot{x} = \sum_{k=1}^{4} g_{k}(x)u_k$ which is a control-linear system represented in \eqref{eq:case:1} and \eqref{eq:case:1:cartesian} and 
\item
Measurement model: $y = h(x) = \tan^{-1}(x_1/x_2)$ represented in \eqref{eq:meas:relativepose}.
\end{itemize}
To apply the method of observability analysis based on Lie derivatives, we computed a sequence of observability subspaces: 
\begin{equation}\label{eq:obs:bearingonly:cartesian}
\begin{split}
\!\!\!\!\mathcal{O}_0 \!&= {\rm span} \{ \mathcal{L}^{0} h \} = {\rm span} \{ [{x_2}/{d^2},-{x_1}/{d^2},0]^\top \} \\
%\mathcal{O}_0 &= {\rm span} \{ \mathcal{L}^{0} h \} = {\rm span} \{ [\frac{x_2}{x_1^2+x_2^2},-\frac{x_1}{x_1^2+x_2^2},0]^\top \} \\
\!\!\!\!\mathcal{O}_1 
\!&= \mathcal{O}_0 \!+ \!\sum_{k=1}^{4} {\rm span} \{ \nabla \mathcal{L}_{g_{k}} h \} \\
%&= {\rm span} \{ [1,0,0]^\top, [0,\sin(x_2),0]^\top, [0,\sin(x_{32}), -\sin(x_{32})]^\top \}
&= {\rm span} 
\left\{ \!\!
\left[\begin{array}{@{\,}c@{\,}} {x_2}/{d^2} \\ -{x_1}/{d^2} \\0 \end{array}\right] \!\!, \! 
\left[\begin{array}{@{}c@{}} {2 x_1 x_2)}/{d^4} \\ -{2 x_2^2)}/{d^4} \\ 0 \end{array}\right] \!\!, \! \right. \\
&\
\left.
\left[\begin{array}{@{\!}c@{\!}} -{2 x_1 x_2 c(x_3))}/{d^4} - 2x_1^2 s(x_3)/d^4 \\ -{2 x_2^2 c(x_3))}/{d^4} + {c(x_3)}/d^2 -{2 x_1 x_2 s(x_3))}/{d^4} \\ -{x_2 s(x_3)}/{d^2} + x_1 c(x_3)/d^2 \end{array}\right]
%\left[\begin{array}{@{\,}c@{\,}} 0 \\[3mm] 1 \\[3mm] 0 \end{array}\right] , 
%\left[\begin{array}{@{}c@{}} \frac{s(x_2)}{x_1^2} \\[3mm] -\frac{c(x_2)}{x_1^2} \\[3mm] 0 \end{array}\right] , 
%\left[\begin{array}{@{\!}c@{\!}} \frac{s(x_23)}{x_1^2} \\[2mm] -\frac{c(x_2)}{x_1} \\[2mm] \frac{c(x_2)}{x_1} \end{array}\right]
\!\!\right\} \\
\!\!\!\!\mathcal{O}_{\ell} \!&= \mathcal{O}_{1} \mbox{ for all $\ell \geq 2$} 
\end{split}
\end{equation}
where $d = \sqrt{x_1^2 + x_2^2}$ is the distance in the 2D space between the two robots. 
The rank of observability (sub-)space is given as
\begin{equation}
{\rm rank}\, \mathcal{O}_1 = 3 %\mbox{ unless $x_2=n\pi$ and $x_3=n'\pi$ for $n,n' \in \mathbb{N}$}
\end{equation}
unless $x_2=0$ or $x_3={n}\pi$ for $n \in \mathbb{N}$, implying that the corresponding system is \emph{locally observable}. 

%------------------------------%
\subsubsection{Orientation-only measurement}
%------------------------------%
For observability analysis of a system with orientation-only measurements, either Polar or Cartesian coordinate system representations can be considered.
\begin{itemize}
\item
Kinematic motion model: $\dot{x} = \sum_{k=1}^{4} g_{k}(x)u_k$ which is a control-linear system represented in \eqref{eq:case:1} and \eqref{eq:case:1:cartesian} (or \eqref{eq:case:1:polar}), and 
\item
Measurement model: $y = h(x) = x_3$ represented in \eqref{eq:meas:relativepose}.
\end{itemize}

Similar to previous cases, the observability subspaces can be computed as follows:
\begin{equation}\label{eq:obs:orientonly}
\begin{split}
\mathcal{O}_0 &= {\rm span} \{ \mathcal{L}^{0} h \} = {\rm span} \{ [0,0,1]^\top \} \\
%\mathcal{O}_1 
%&= \mathcal{O}_0 + \sum_{k=1}^{4} {\rm span} \{ \nabla \mathcal{L}_{g_{k}} h \} \\
\mathcal{O}_{\ell} &= \mathcal{O}_{1} \mbox{ for all $\ell \geq 1$} 
\end{split}
\end{equation}
and the rank of observability (sub)space is. 
\begin{equation}
{\rm rank}\, \mathcal{O}_1 = 1 \,,
\end{equation}
regardless of the state $x$ and the input $u$. It can be concluded that the relative position between robots cannot be estimated using only relative orientation estimation, even though the velocities are known.

%-----------------------------------------------------------------------------%
\subsection{Observability without global odometry data}
\label{sec:nonobsanal:2}
%-----------------------------------------------------------------------------%
In Section~\ref{sec:nonobsanal:1}, we operate under the assumption that wheel odometry or kinematic velocity data can be shared through a communication network. When the odometry inputs of a neighboring robot are unavailable, range-only or bearing-only measurements fail to guarantee observability in inter-robot relative pose estimation. Our study demonstrates that with both range and bearing measurements available, observability in inter-robot relative pose estimation can be assured without the need for communicating odometry data. This analysis has relevance to nonlinear unknown input observability~\cite{martinelli2018nonlinear,martinelli2022nonlinear,maes2019observability} and disturbance observer literature~\cite{chen2015disturbance,veluvolu2009high,radke2006survey}. We maintain focus on the observability analysis method introduced in Section~\ref{sec:nonobsanal:1}, considering the augmented state-space model in which the velocities of a neighboring robot are treated as additional unknown states.

For the nonlinear observability analysis, we assume that odometry inputs of a neighboring robot follow Brownian motion, as given in~\eqref{eq:case:2}, using polar coordinates. Similar to Section~\ref{sec:nonobsanal:1}, we employ a nonlinear observability method based on Lie derivatives. 
The following augmented state–space model and range-bearing measurement equations are considered:

\begin{itemize}
\item
Kinematic motion model: $\dot{x} = \bar{g}_0(x) + \sum_{k=1}^{2} \bar{g}_{k} (x) u_k$ which is a control-affine system represented in \eqref{eq:case:2} and \eqref{eq:case:2:polar}, where $x= [\rho_{ji}, \beta_{ji}, \theta_{ji}, v_{j}, \omega_{j}] \in \mathbb{R}^2 \times \mathbb{S} \times \mathbb{R}^2$ and 
\[
\bar{g}_0(x) \!=\!\! \begin{bmatrix} x_{4} c(x_{32}) \\ x_{4}\!\displaystyle\frac{s(x_{32})}{x_{1}} \\ x_{5} \\0 \\ 0 \end{bmatrix}\!\!, \,
\bar{g}_1(x) \!=\!\! \begin{bmatrix} - c(x_{2}) \\ -\displaystyle\frac{s(x_{2})}{x_{1}} \\ 0 \\0 \\ 0 \end{bmatrix}\!\!, \,
\bar{g}_2(x) \!=\!\! \begin{bmatrix} 0 \\ 0 \\ -1 \\ 0 \\ 0 \end{bmatrix}\!\!, \,
\]
\item
Measurement model: $y_1 = h_1(x) = \rho_{ji} = x_1$ and $y_2 = h_2(x) = \beta_{ji} = x_2$ represented in \eqref{eq:meas:relativepose}.
\end{itemize}

Using the Lie derivatives, we compute the vectors spanning the state-dependent observability space as
\[
O_{01} = \nabla \mathcal{L}^{0} h_1 \!=\!\! \begin{bmatrix} 1 \\ 0 \\ 0 \\ 0 \\ 0 \end{bmatrix}\!\!, \, 
O_{02} = \nabla \mathcal{L}^{0} h_2 \!=\!\! \begin{bmatrix} 0 \\ 1 \\ 0 \\ 0 \\ 0 \end{bmatrix}\!\!, \, 
\]
\[
\begin{split}
O_{110} &= \nabla \mathcal{L}^{1}_{\bar{g}_0} h_1 
= \nabla(\nabla h_1 \cdot \bar{g}_{0}) 
%= \nabla x_{4}c(x_{32})
\!=\!\! \begin{bmatrix} 0 \\ x_{4}s(x_{32}) \\  -x_{4}s(x_{32}) \\ c(x_{32}) \\ 0 \end{bmatrix}\!\!, \, \\
\end{split}
\]
\[
\begin{split}
O_{120} &= \nabla \mathcal{L}^{1}_{\bar{g}_0} h_2 
= \nabla(\nabla h_2 \cdot \bar{g}_{0}) 
%= \nabla x_{4}\displaystyle \frac{s(x_{32})}{x_{1}}
 \!=\!\! \begin{bmatrix} -\displaystyle \frac{x_{4}s(x_{32})}{x_{1}^2} \\ \displaystyle \frac{x_{4}c(x_{32})}{x_{1}} \\ 0 \\ \displaystyle \frac{s(x_{32})}{x_{1}} \\ 0 \end{bmatrix}\!\!, \, \\
 \end{split}
\]
\[
 \begin{split}
O_{111} &= \nabla \mathcal{L}^{1}_{\bar{g}_1} h_1 
= \nabla(\nabla h_1 \cdot \bar{g}_{1}) 
%= - \nabla c(x_{2})
\!=\!\! \begin{bmatrix} 0 \\ s(x_{2}) \\  0 \\ 0 \\ 0 \end{bmatrix}\!\!, \, \\
 \end{split}
\]
\[
 \begin{split}
O_{121} &= \nabla \mathcal{L}^{1}_{\bar{g}_1} h_2 
= \nabla(\nabla h_2 \cdot \bar{g}_{1}) 
%= - \nabla \displaystyle \frac{s(x_{2})}{x_{1}}
 \!=\!\! \begin{bmatrix} \displaystyle \frac{s(x_{2})}{x_{1}^2} \\ -\displaystyle \frac{c(x_{2})}{x_{1}} \\ 0 \\ 0 \\ 0 \end{bmatrix}\!\! .
\end{split}
\]
In addition to the first-order Lie derivatives and associated observability subspaces, we are interested in a nontrivial higher-order Lie derivative as follows:
%\begin{figure*}
\[
\begin{split}
O_{2100} 
= \mathcal{L}^{2}_{\bar{g}_0\bar{g}_1} h_1 
&= \nabla \left( \nabla \mathcal{L}^{1}_{\bar{g}_0} h_1 \cdot \bar{g}_0 \right) \\
%& = \nabla
%\begin{bmatrix} 0 \\ x_{4}s(x_{32}) \\  -x_{4}s(x_{32}) \\ c(x_{32}) \\ 0 \end{bmatrix}
%\cdot
%\begin{bmatrix} x_{4} c(x_{32}) \\ x_{4}\!\displaystyle\frac{s(x_{32})}{x_{1}} \\ x_{5} \\0 \\ 0 \end{bmatrix} \\
%& =
%\nabla \left( x_4^2 s^2(x_{32})/x_{1} - x_{4}x_{5}s(x_{32}) \right) \\
&=
\left[
\begin{array}{c} 
\displaystyle \frac{x_4^2 s^2(x_{32})}{x_{1}^2} \\[3mm]
 -\displaystyle \frac{x_4^2 s(x_{32}) c(x_{32})}{x_{1}} +x_{4}x_{5}c(x_{32}) \\[3mm] 
\displaystyle \frac{-x_4^2 s(x_{32}) c(x_{32})}{x_{1}} -x_{4}x_{5}c(x_{32}) \\[3mm] 
 \displaystyle \frac{2 x_4 s^2(x_{32})}{x_{1}} - x_{5}s(x_{32})  \\[3mm] 
 - x_{4}s(x_{32})
 \end{array}
 \right]
\end{split}
\]
%\end{figure*}

\begin{table}[t!]
\begin{center}
\centerline{\includegraphics[width=.925\linewidth]{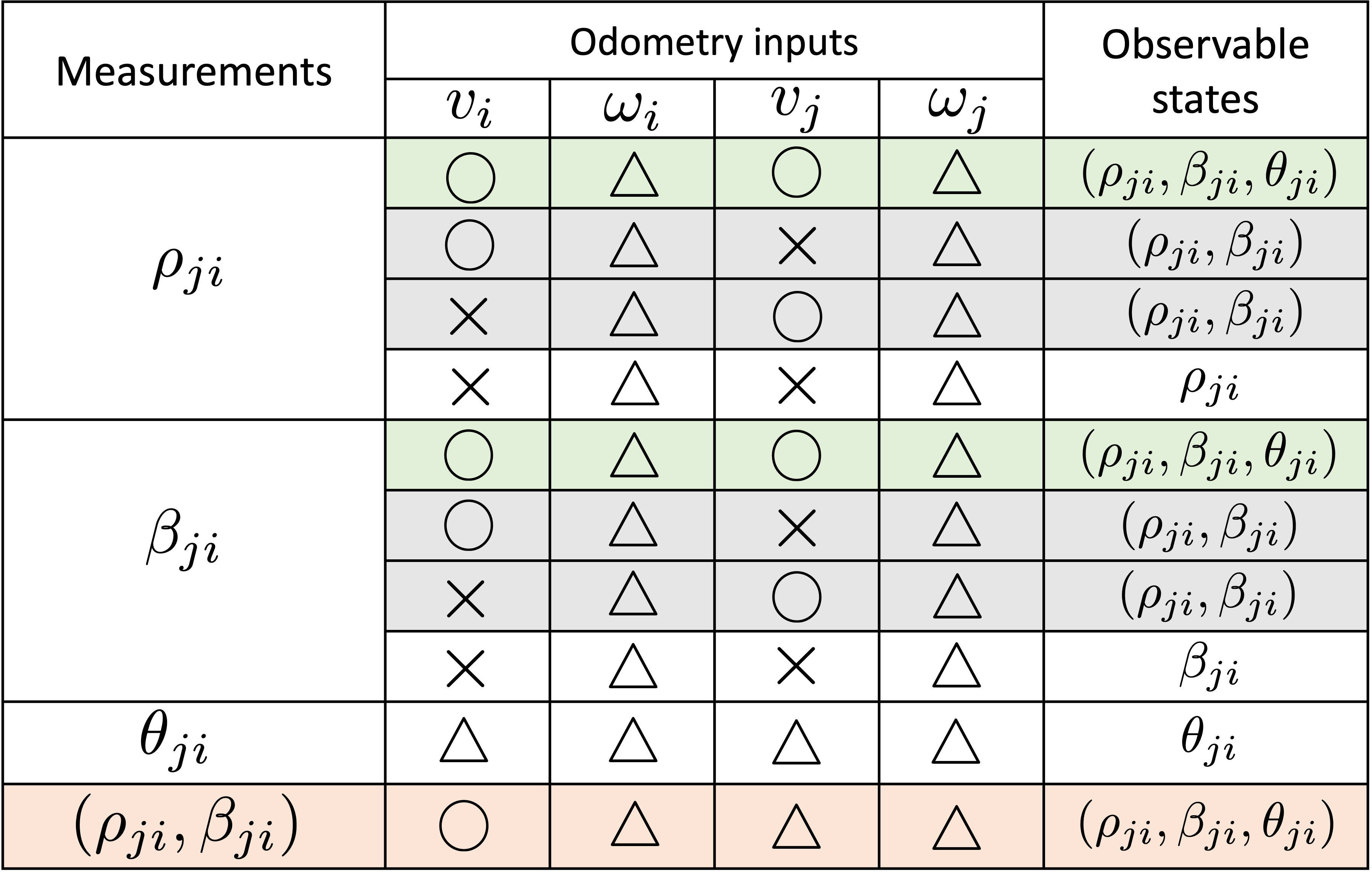}}
\caption{Observability subspaces associated with measurements and odometry data. (For odometry inputs, $\bigcirc$ means {\em non-zero}, $\times$ means {\em zero}, and $\bigtriangleup$ means either zero or non-zero.)}
\label{tab:2}
\end{center}
\end{table}

Thus, we have the following sequence of observability subspaces
\[
\begin{split}
\mathcal{O}_0 &= {\rm span} \{ O_{01}, O_{02} \} , \\
\mathcal{O}_1 &= \mathcal{O}_0 + {\rm span} \{ O_{110}, O_{120}, O_{1101}, O_{121} \} , \\
\mathcal{O}_2 & \supseteq \mathcal{\tilde O}_2 : = \mathcal{O}_1 + {\rm span} \{ O_{2100} \} 
\end{split}
\]
and the rank condition
\[
{\rm rank}\, \mathcal{\tilde O}_2 = 5 \ \Rightarrow \ {\rm rank}\, \mathcal{O}_2 = 5 \,,
\]
which implies that an inter-robot system with both range and bearing measurements is \emph{locally observable} almost everywhere. 

The observability analyses with different information structures are summarized in Table~\ref{tab:2}. The new result indicates that the system of inter-robot relative pose dynamics is observable when both range and bearing measurements are available, even in the absence of velocity information from the neighboring robot.

\begin{remark}[Velocity tracking]
In addition to indirect velocity estimation through position (range and bearing) measurements, the direct measurement and computation of velocity based on the Doppler effect in radar~\cite{chen2006micro,chen2019micro} and LiDAR~\cite{ma2019moving} have been extensively explored for applications such as target tracking~\cite{ma2019moving,sun2021vessel,kellner2014instantaneous} and ego-motion estimation~\cite{kellner2013instantaneous,kellner2013egomotion}. Intuitively, such direct velocity tracking can ensure observability even with range-only or bearing-only measurements, obviating the need to communicate wheel-odometry data.
\end{remark}

%==============================================================%
\section{ROS/Gazebo Simulation Results: Two Mobile Robot Cases}
\label{sec:gazebo}
%==============================================================%
\begin{figure}[!t]
	\centering
	\includegraphics[width=1.0\linewidth]{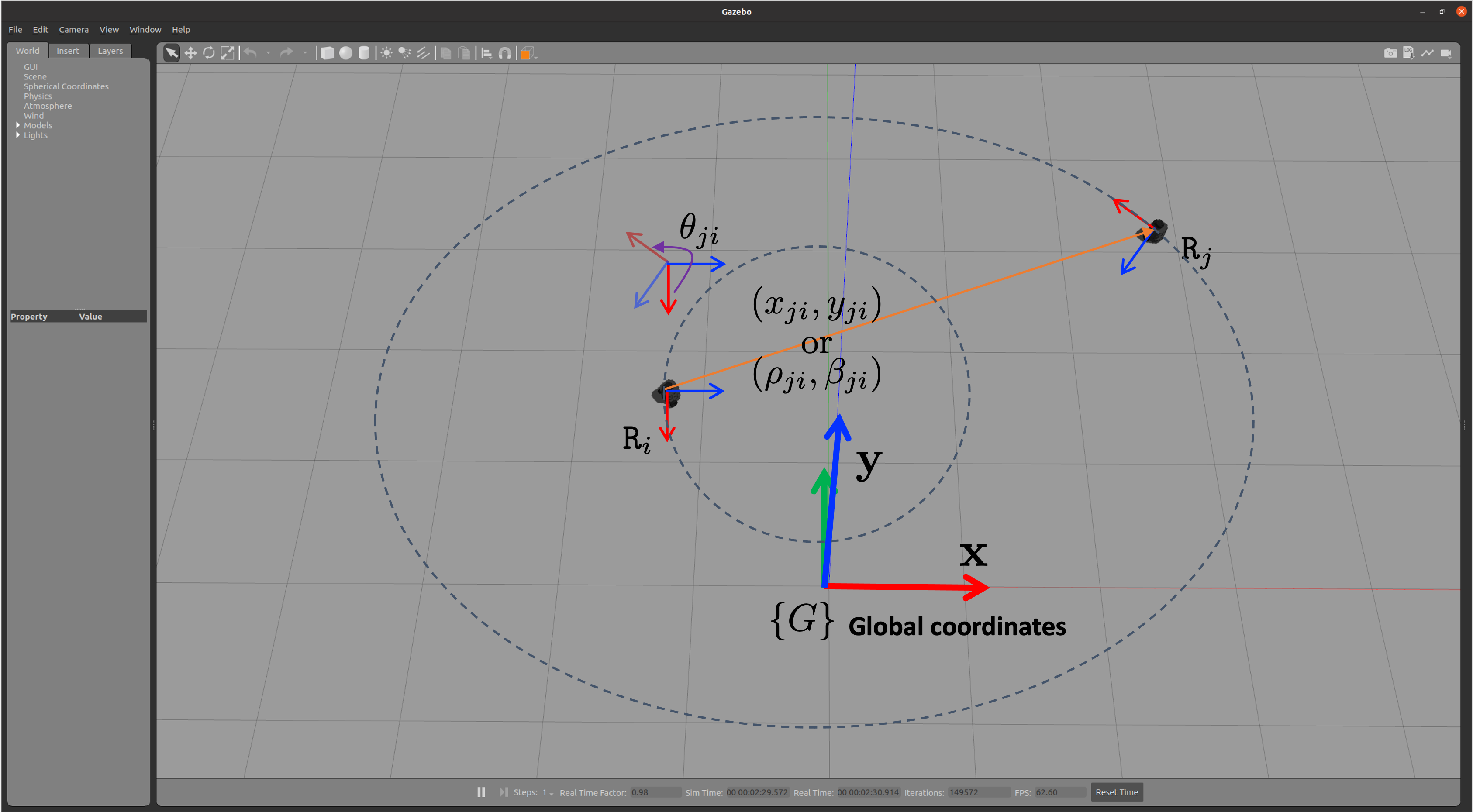}\\[-2mm]
	\caption{ROS/Gazebo simulation environments for inter-robot relative pose estimation of two Turtlebot3 robots.}
	\label{fig:ros-gazebo} 
\end{figure}

This section presents the simulation results of inter-robot relative pose estimation using EKF and PGO under various information structures. 
For the simulation experiments demonstrating observability analysis and estimation, the robot operating system (ROS) and Gazebo simulation environments were employed, as depicted in Fig.~\ref{fig:ros-gazebo}. These environments enable seamless transitions from simulation to real hardware implementation. The estimation source code and simulations were executed on a laptop equipped with an Intel i7-9700K CPU (3.6 GHz), 32-GB LPDDR5 RAM, 512-GB PCIe NVMe x2 SSD onboard, and an Ubuntu 20.04 LTS operating system.

For Gazebo, we created two Turtlebot3 robot objects and performed relative pose estimations between them. Gazebo provides the ground-truth data of robot pose $X=[p_x , p_y , p_z , q_x , q_y , q_z , q_w ]$ (3D position and 3D orientation using quaternions)\footnote{Since our work considers the planar motion and pose in $\text{SE}(2)$, we only need to export data of 2D position $(p_x,p_y)$ and 1D orientation, the yaw angle, from the quaternions as $\theta = {\rm atan2}(2(q_y q_z + q_w q_x), 1 - 2(q_x^2 + q_y^2))$.} for each Turtlebot3 object in the simulation based on the origin of the map, that is, global coordinates. Using this information on the absolute pose of each robot, we generated distance, bearing, and orientation observation data with artificial measurement noise between the two robot objects. The linear and angular velocities $U=[v_i , w_i , v_j , w_j ]$ which are the kinematic control inputs of the two robots ${\tt R}_i$ and ${\tt R}_j$, are set as constants $U=[0.2, 0.1, 0.4, 0.09]$. Therefore, the two robots moved in concentric circles with radii of 2 m and 4 m.
%It is assumed that there is no noise in the odometry $(v_i , w_i )$ and the network odometry $(v_j , w_j )$.

We considered four different experimental scenarios employing distinct information structures for inter-robot state estimation.
\begin{itemize}
\item 
For Case 1, we assumed the availability of an inter-robot communication network, onboard sensing of wheel-encoder, and inter-robot ranging measurements.
\item 
For Case 2, we assumed the availability of an inter-robot communication network, onboard sensing of wheel-encoder, and inter-robot bearing measurements.
\item 
For Case 3, we assumed the availability of an inter-robot communication network, onboard sensing of wheel-encoder, and both inter-robot ranging and bearing measurements.
\item 
For Case 4, we assumed only the availability of onboard sensor measurements for inter-robot ranging and bearing without an inter-robot communication network.
\end{itemize} 

\begin{figure}[!t]
	\centerline{\includegraphics[width=.98\linewidth]{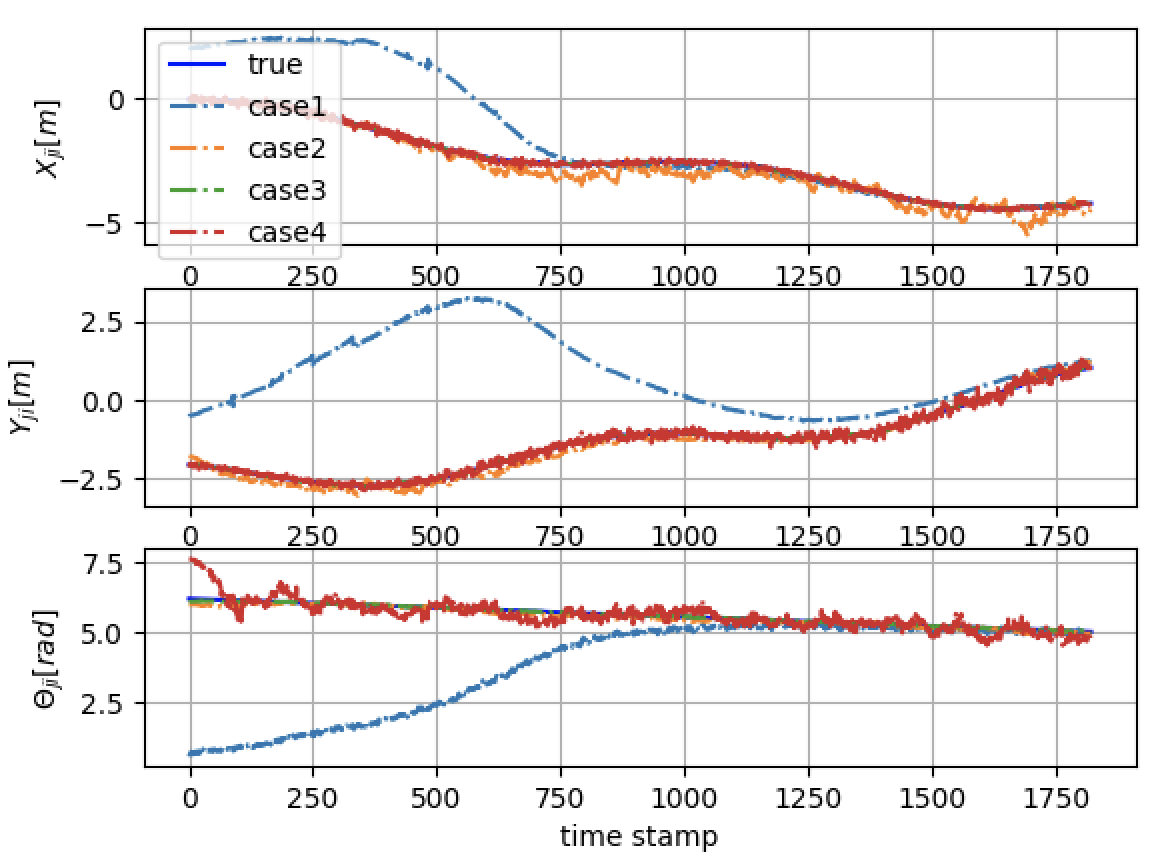}}\vspace{-3.5mm}
	\caption{EKF-based relative pose estimation results with different information structures (Cases 1$\sim$4).}
	\label{fig:EKFresult}
%\end{figure}
\vspace{3mm}
%\begin{figure}[!t]
	\centerline{\includegraphics[width=.99\linewidth]{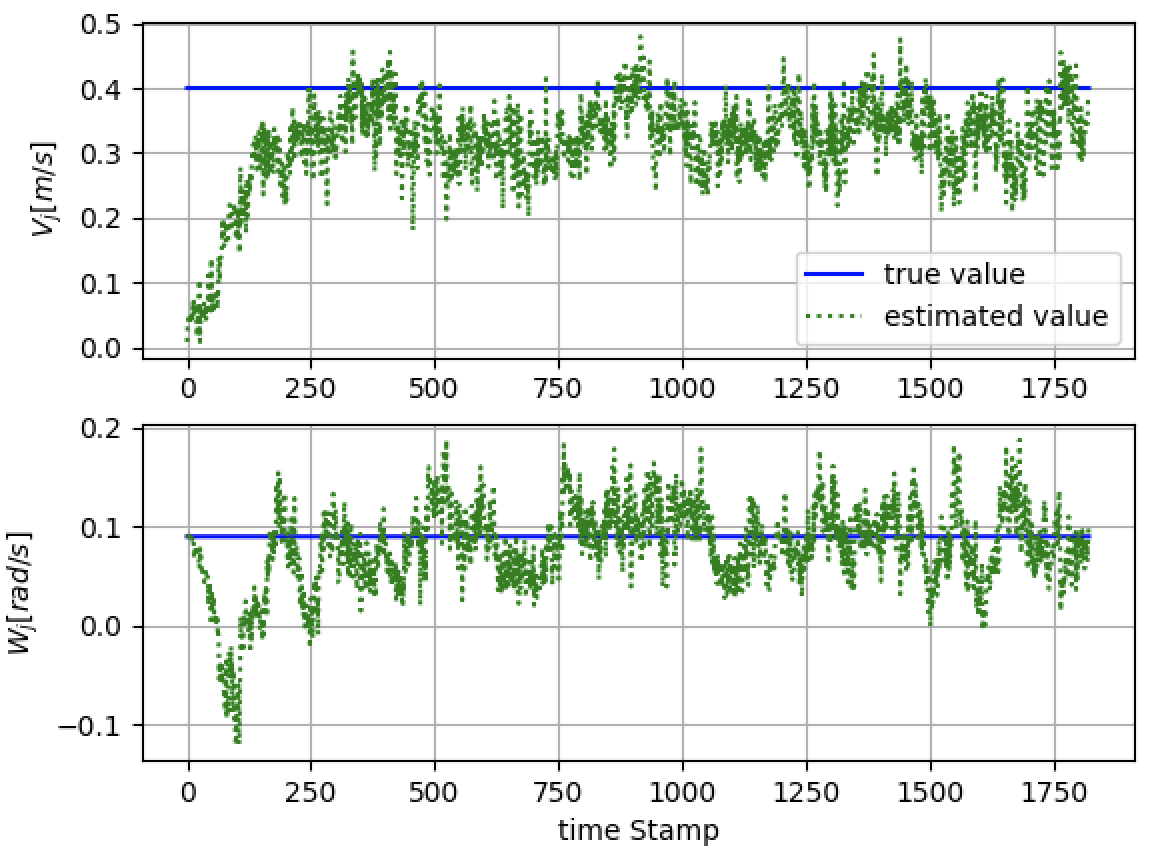}}\vspace{-3.5mm}
	\caption{EKF-based velocity ($v_j, w_j$) estimation results for Case 4.}
	\label{fig:EKFresult_Case4}
\end{figure}

%-----------------------------------------------------------------------------%
\subsection{EKF estimation}
%-----------------------------------------------------------------------------%
%After acquiring observation data for 2 minutes in the simulation, an Extended Kalman Filter estimation experiment was performed for each scenario.

Fig.\ref{fig:EKFresult} displays the results of EKF-based estimation and ground truth for four different cases of information structures in ROS/Gazebo simulations. In Case 1, where only distance measurements were available, the initial estimation accuracy was low but improved over time. Notably, Case 2, where only bearing information was measured, exhibited higher estimation accuracy. In Case 4, where wheel-odometry data or velocities ($v_j,w_j$) of a neighboring robot were not provided but estimated alongside the relative pose to the robot, the estimation results are shown in Fig.\ref{fig:EKFresult_Case4}. Table~\ref{tab:ekfRmse} compares the estimation accuracy using RMSE values of the EKF-based state estimation for the four cases with different measurements and information. Case 1, which relied solely on range measurements for inter-robot relative pose estimation, exhibited the largest RMSE. In Case 4, where there was no communication network between the robots, the RMSE for the orientation estimation $\theta_{ji}$ was high; however, the estimation of the position 
$(x_{ji},y_{ji})$ showed relatively accurate results.

\begin{table}[b!]
    \centering
    \caption{RMSE of state (relative pose in planar motions) estimation for each case of information fusion using EKF}\vspace{-2mm}
    \begin{tabular}{|c|c|c|c|}
    	\hline
         & $x_{ji}$ & $y_{ji}$ & $\theta_{ji}$ \\[.5mm] 
         \hline
         Case 1 & 1.6827 & 2.8620 & 2.6385 \\
         Case 2 & 0.3133 & 0.1601 & 0.0906 \\
         Case 3 & 0.0213 & 0.0331 & 0.0198 \\
         Case 4 & 0.0706 & 0.0994 & 0.2971 \\
         \hline
    \end{tabular}
    \label{tab:ekfRmse}
\end{table}

%-----------------------------------------------------------------------------%
\subsection{Nonlinear least-square estimation}
%-----------------------------------------------------------------------------%
%In the least squares method, estimation was performed using sensor data for 1 minute, which is half the simulation time of the Kalman filter estimation for time saving. 
\begin{figure}[!t]
	\centerline{\includegraphics[width=.8\linewidth]{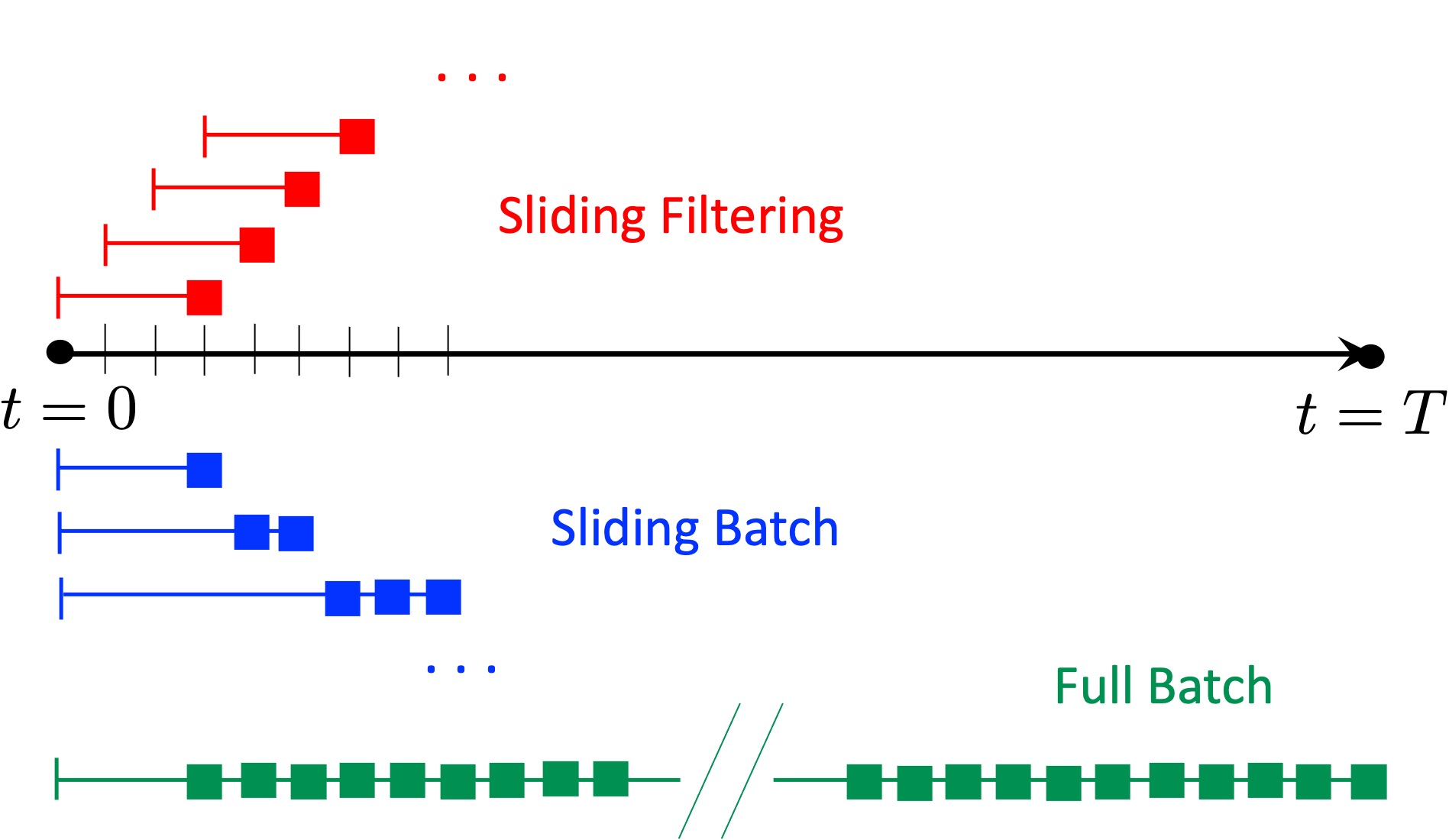}}\vspace{-3.5mm}
	\caption{Diagram of NLS-based PGO for state estimation with different data processing strategies: Sliding Filtering, Sliding Batch, and Full Batch. The colored solid lines refer to the horizon of measurement data considered for optimization-based estimation, while the colored square boxes correspond to the state estimates resulting from the applied methods.}
	\label{fig:nls-slidingwindow}
\end{figure}

Similar to the EKF-based estimation, simulation experiments of NLS estimation for inter-robot relative pose were performed for each scenario of the information structure. The Ceres Solver~\cite{AgarwalCeresSolver2022} was employed for numerical optimization. Three methods were implemented for data processing and numerical optimization. The first is the Sliding Filtering (SF) method, which solves NLS problems using a fixed-size sliding window for semi-batch optimization. Only the last state estimation over the estimation horizon is updated, resulting in partial smoothing for filtering. The second method is the Sliding Batch (SB) method, where the size of the sliding window is gradually increased, and associated batch optimization problems are solved. In the SB method, only the state estimates over the increased sliding window are updated, leading to full smoothing for partial smoothing. Finally, the Full Batch (FB) method considers all measurements, estimating the state trajectory over the runtime simultaneously.

\begin{table}[b!]
    \centering
    \caption{Comparison of total RMSE values of state (relative pose in planar motions) estimation using EKF and NLS methods}\vspace{-2mm}
    \begin{tabular}{|c|c|c|c|c|}
    	\hline
          & EKF & SF & SB & FB \\[.5mm]
         \hline
         Case 1 & 2.3944 & 1.0543 & 1.3821 & 0.0321 \\
         Case 2 & 0.1880 & 0.7783 & 0.0618 & 0.0197 \\
         Case 3 & 0.0247 & 0.1050 & 0.0105 & 0.0090 \\
         Case 4 & 0.1242 & 0.0267 & 0.0132 & 0.0061 \\
         \hline
    \end{tabular}
    \label{tab:rmse}
\end{table}

\begin{figure}[!t]
	\centering
	\includegraphics[width=1.0\linewidth]{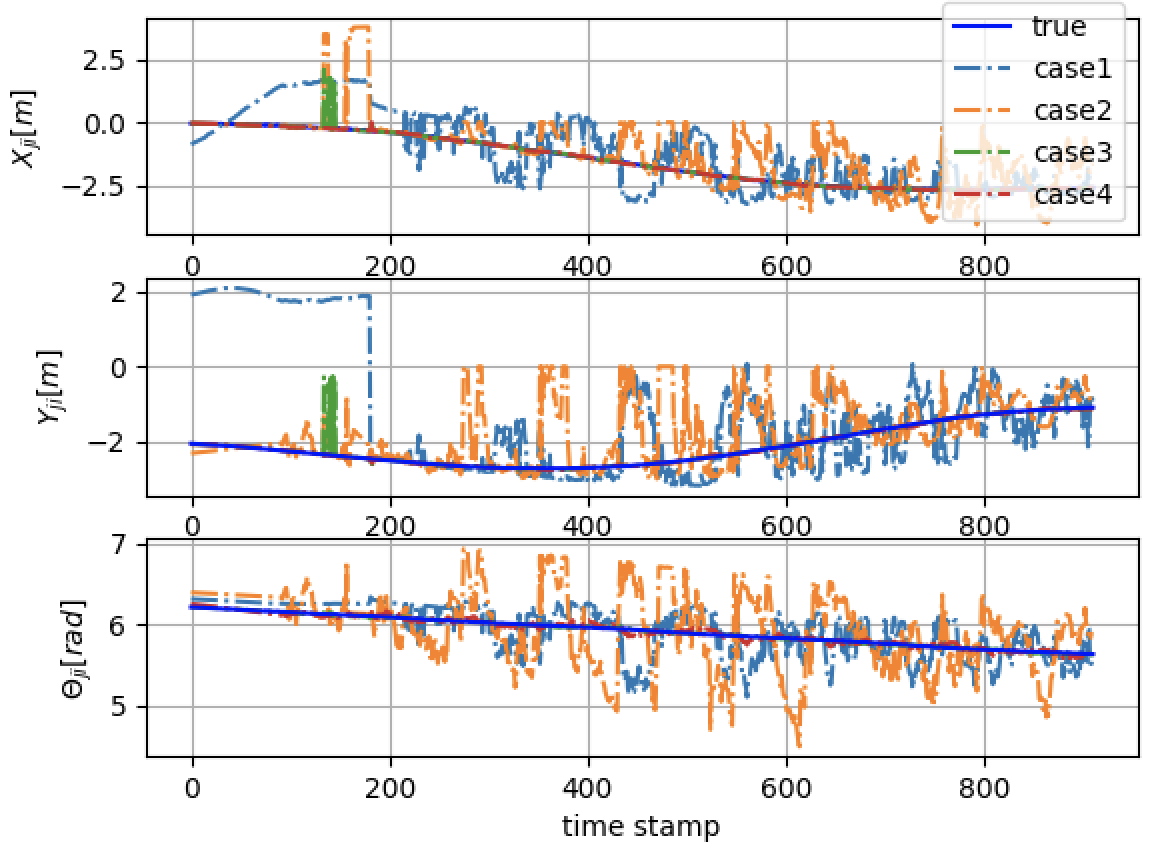}\vspace{-3.5mm}
	\caption{Relative pose estimation result of each cases in SF}
	\label{fig:sliding_nls_filtering_estimation} 
%\end{figure}
\vspace{4mm}
%\begin{figure}[!t]
	\centering
	\includegraphics[width=1.0\linewidth]{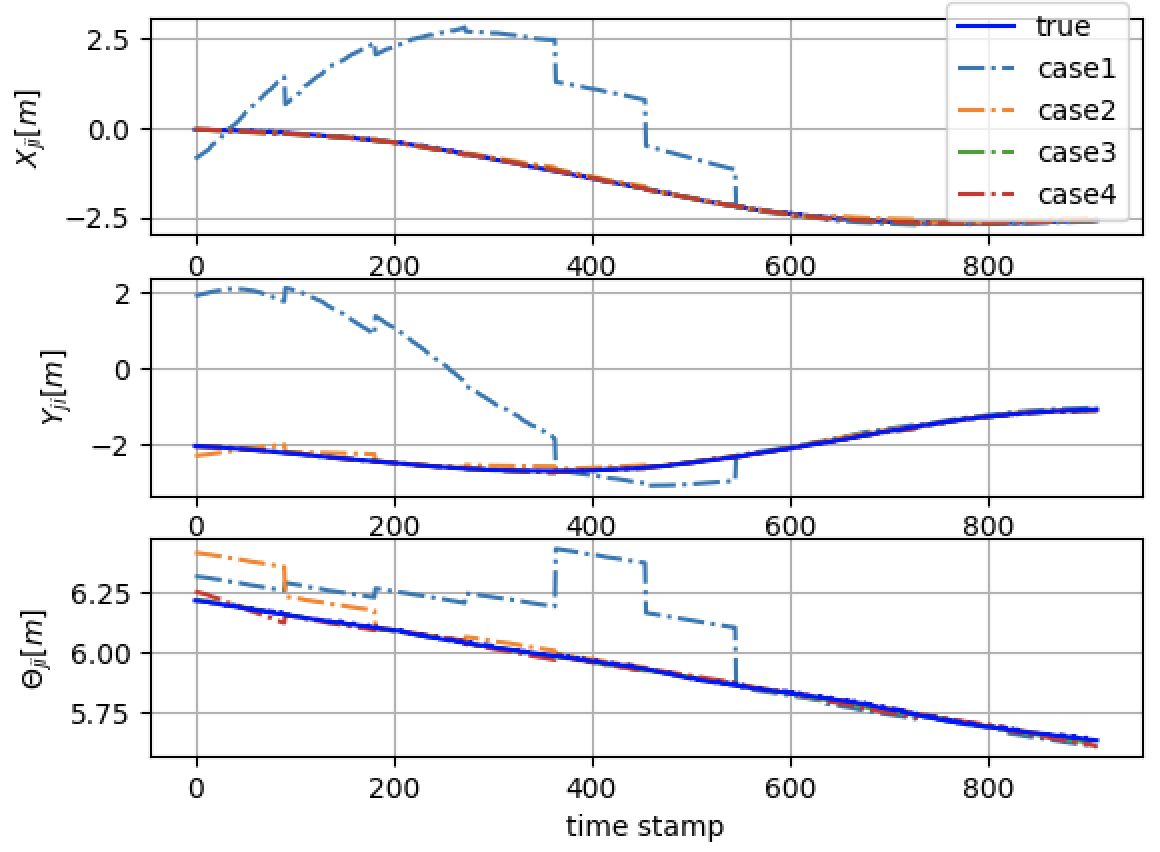}\vspace{-3.5mm}
	\caption{Relative pose estimation result of each cases in SB}
	\label{fig:sliding_nls_estimation} 
\end{figure}

\begin{figure}[!t]
	\centering
	\includegraphics[width=1.0\linewidth]{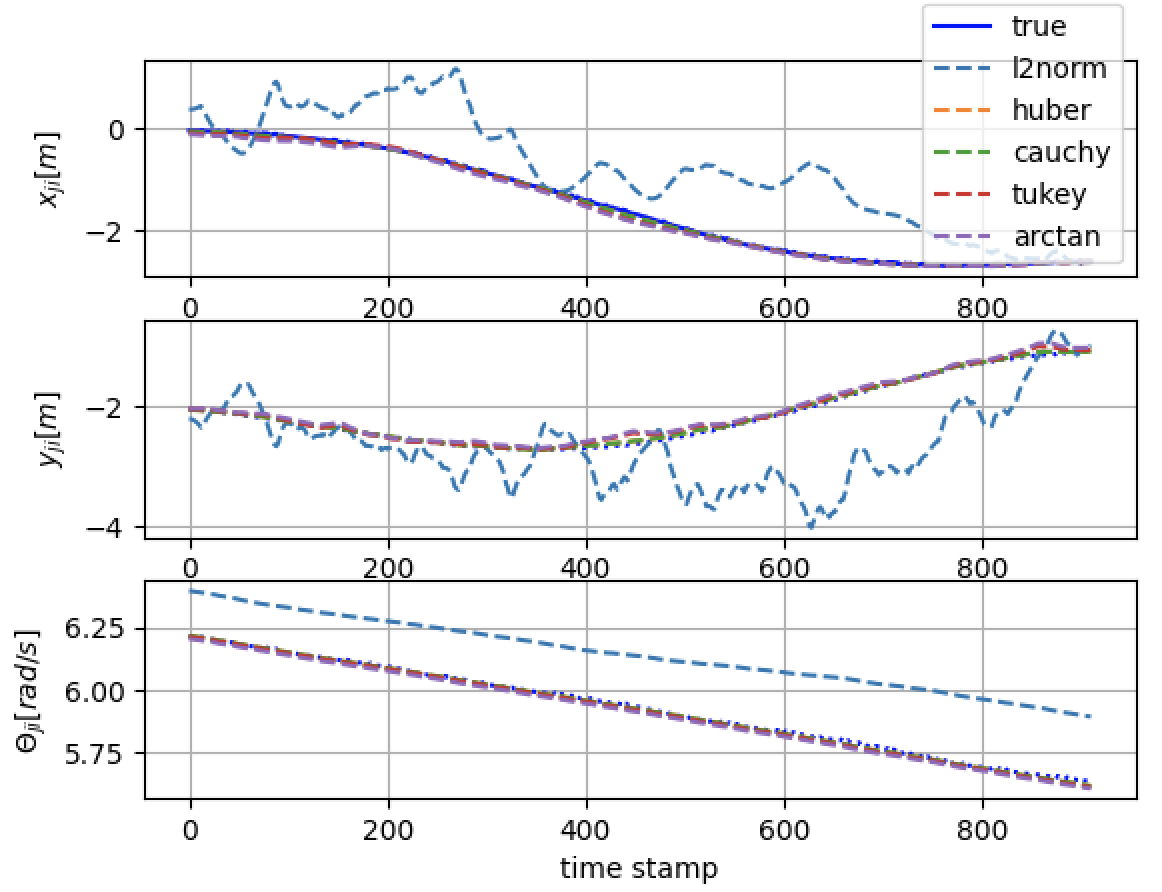}\vspace{-3.5mm}
	\caption{ Relative pose estimation results of M-estimation using different kernel functions are showcased for Case 4, wherein no communication between robots occurs, and only range and bearing measurements are utilized for estimation.}
	\label{fig:mEstimatorCase4} 
%\end{figure}
\vspace{3mm}
%\begin{figure}[!t]
	\centering
	\includegraphics[width=1.0\linewidth]{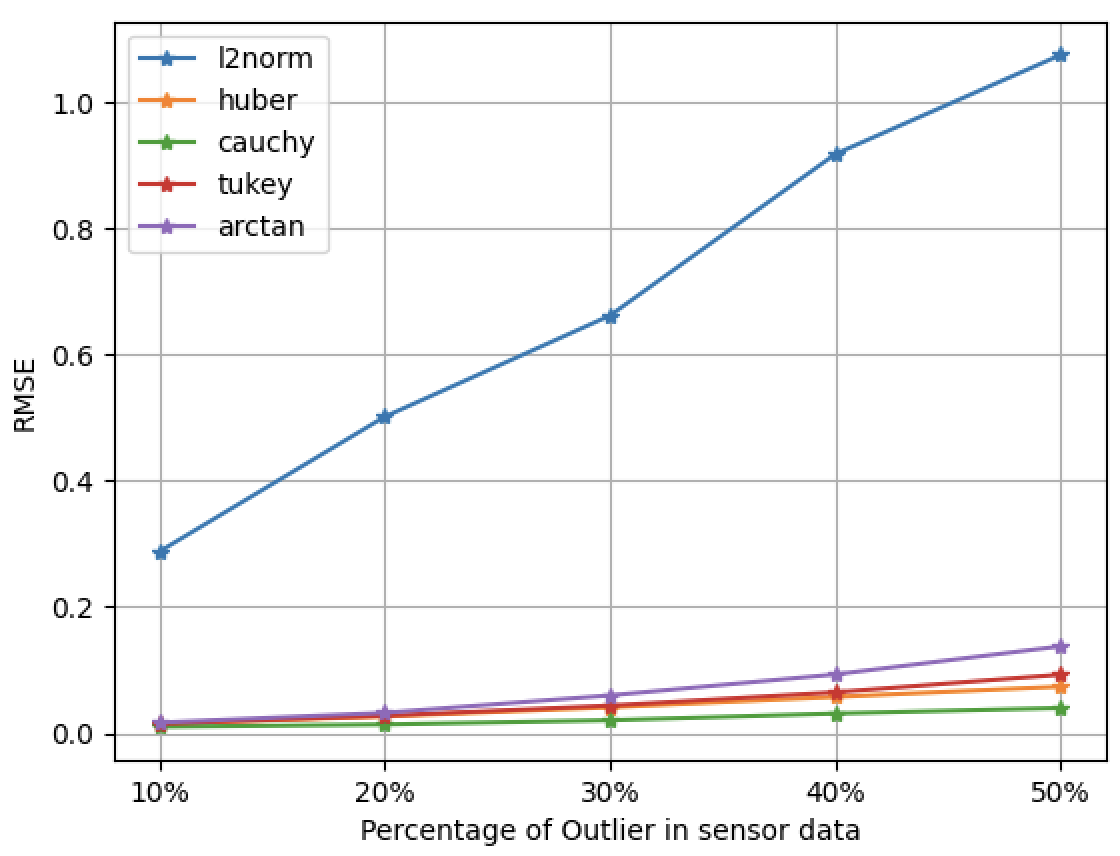}\vspace{-3.5mm}
	\caption{RMSE of M-estimation results using different kernel functions with varying ratios of outliers for Case 4.}
	\label{fig:mEstimatorCase410to50} 
\end{figure}

Table~\ref{tab:rmse} provides a comparative analysis of RMSE values obtained through the Kalman filter and those derived from the optimization method. The batch method exhibited the smallest estimation error. Figs.~\ref{fig:sliding_nls_filtering_estimation} and \ref{fig:sliding_nls_estimation} depict the results of relative pose estimation using SF and SB, respectively. 

%-----------------------------------------------------------------------------%
\subsection{Robust M-estimation}
%-----------------------------------------------------------------------------%
For the robust M-estimation experiment, outliers were introduced to the sensor data using the interquartile range (IQR) method~\cite{rousseeuw2011robust}. The experiment involved incrementally increasing the ratio of outliers to the sensor data from 10\% to 50\% in 10\% intervals, employing various kernel functions (Huber, Cauchy, Tukey, Arctan) provided by the Ceres solver. Estimation experiments were exclusively conducted for Case 4 (Odometry + range and bearing sensors). Fig.~\ref{fig:mEstimatorCase4} illustrates the estimation results using each kernel function when the sensor data contained 30\% outliers. The L2norm represents the result of estimation using the least-squares method without applying a kernel function. It is evident that the estimation results deviate from the true values due to the influence of outlier data. In contrast, the estimation results obtained using the kernel function were observed to be more accurate than those obtained using the least-squares method. 

Fig.~\ref{fig:mEstimatorCase410to50} compares the calculated Root Mean Square Error (RMSE) obtained by applying various kernel functions while increasing the outlier rate in Case 4. The threshold value of each kernel function was empirically determined, setting it to Huber (0.5), Cauchy (0.5), Tukey (3.0), and Arctan (3.0); the tuning parameter value $t$ in Table~\ref{tab:1} was selected experimentally. It is notable that the RMSE of all kernel functions increases as the proportion of outliers increases. However, compared to the rapid increase in RMSE in L2norm, the RMSE gradually increases when the kernel function is applied.

%==============================================================%
\section{Hardware Experiments}\label{sec:hardware}
%==============================================================%
A series of experiments was conducted to evaluate the precision of inter-robot relative pose estimation and its real-world hardware viability. Specifically considering the information structure of Case 4 from the previous section, which assumes only onboard sensor measurements of inter-robot ranging and bearing are available without any communication network between robots. This successful outcome inherently validates the efficacy and observability of other information structures. Moreover, four distinct motion scenarios of the two robots' movements were designed, including the mirrored motion scenario carried out in the simulation, reflecting potential relative positional dynamics inherent in inter-robot systems. Similar to the simulation, both EKF and PGO were applied to the estimation method and compared for each scenario. A video of the experiment is provided in the Supplementary Material.

%-----------------------------------------------------------------------------%
\subsection{Experimental setup}
%-----------------------------------------------------------------------------%
Two Turtlebot3 robots, designated as ego-robots and neighbor-robots, were employed for experimental trials. A UWB module from NoopLoop provided precise mutual range and bearing measurements, calculated based on the time of arrival (TOA) and angle of arrival (AOA) of received signals. Each robot was equipped with a tag and anchor pair, delivering data at a frequency of 200 Hz with an accuracy of 5 cm ranging and $5^{\circ}$ for direction-angle finding. The operational range of the tag was within $\pm90^{\circ}$ relative to the anchor coordinates, with a noted decrease in measurement accuracy as it approached the $\pm90^{\circ}$ extremes.

For EKF-based estimation, the algorithm ran on a LattePanda Alpha, a single-board computer (SBC) equipped with an Intel Core m3-7Y30 CPU (2.6GHz) and 8 GB of Dual Channel RAM onboard. PGO-based estimation computations were performed on a laptop equipped with an Intel i7-9700k CPU (3.5 GHz), 32 GB LPDDR5 RAM, and 512 GB PCLe NVMe x2 SSD onboard. The entire code was based on ROS and C++, operating on the Ubuntu 20.04 LTS operating system.
The indoor experiments were implemented with the provision of robots’ position and orientation as ground truth, measured in the absolute coordinate system defined by the motion capture (MoCap) system. This system is equipped with ten Qualisys ARQUS A5 cameras, which capture objects at a frame rate of $700\!\!\sim\!\!1400$ fps and with $1\!\!\sim\!\!5$ MP of pixels, featuring a 3D tracking resolution of 0.06 mm and a camera latency of 1.4 ms.
 
The experimental design comprised four scenarios, each meticulously crafted with sensor configurations to analyze performance variations concerning the robots' relative positions, linear and angular velocities.
\begin{itemize}
\item
For Scenario 1, the two robots move in concentric circles at constant velocities with radii of 1.5 m and 2.0 m, respectively.
\item
For Scenario 2, the robots’ motion pattern was similar to Scenario 1, but with varying speeds over time.
\item
For Scenario 3, the neighbor-robot moves in a circle with a radius of 0.8 m, while the ego-robot remains stationary at a point outside the circle.
\item
For Scenario 4, while the neighbor-robot moves similarly to Scenario 3, the ego-robot out of the circle moves back and forth toward the center of the circle.
\end{itemize}

\begin{figure}[!t]
	\centerline{\includegraphics[width=1.0\linewidth]{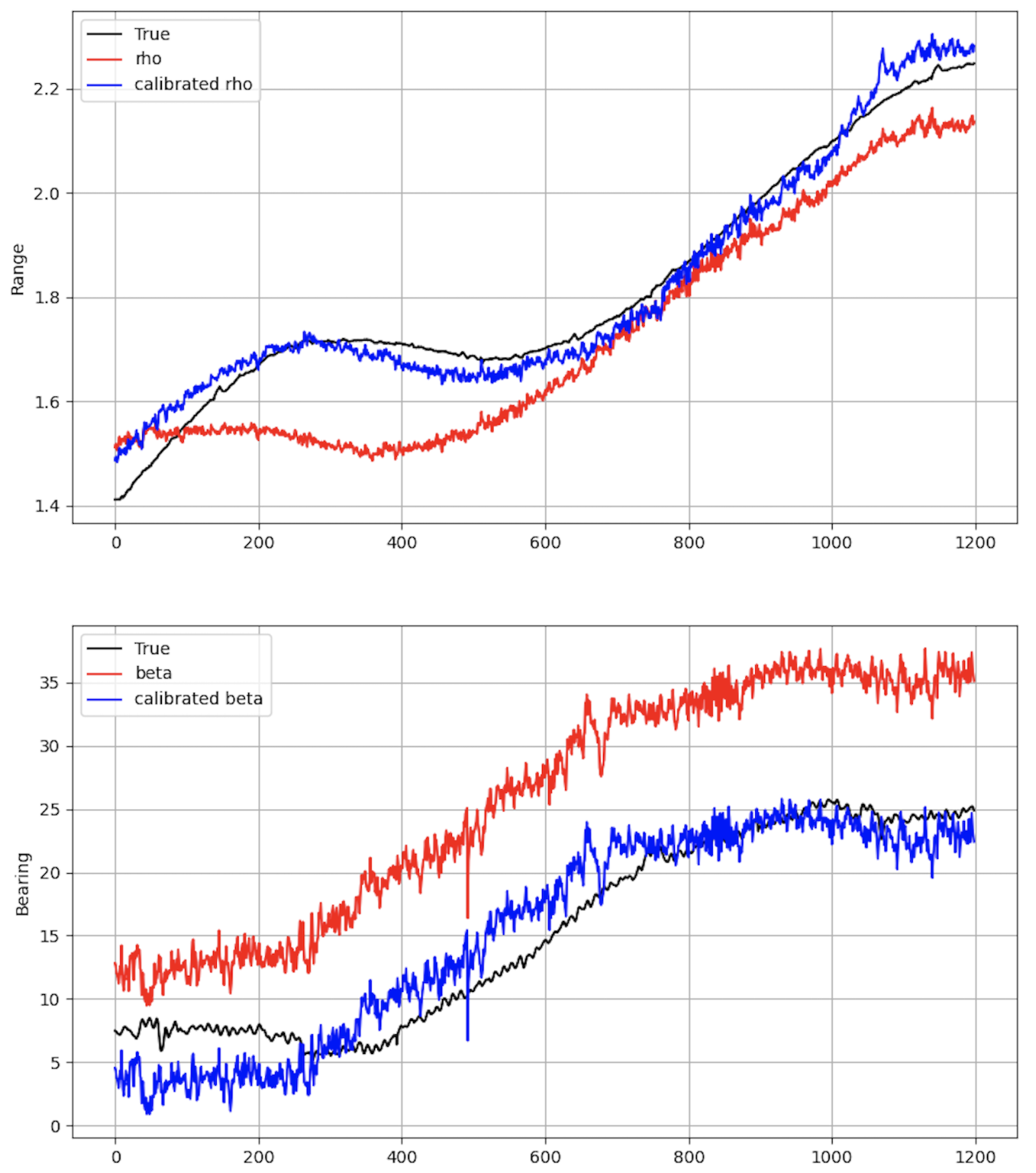}}\vspace{-3.5mm}
	\caption{Results of UWB-AOA calibration for range and bearing: the black solid line represents the ground truth, the red solid line signifies raw UWB measurements, and the blue solid line depicts calibrated measurements as per\eqref{eq:calib_meas}, which are utilized for estimation.}
	\label{fig:UWBcalibration}
%\end{figure}
\vspace{3mm}
%\begin{figure}[!t]
	\centerline{\includegraphics[width=.905\linewidth]{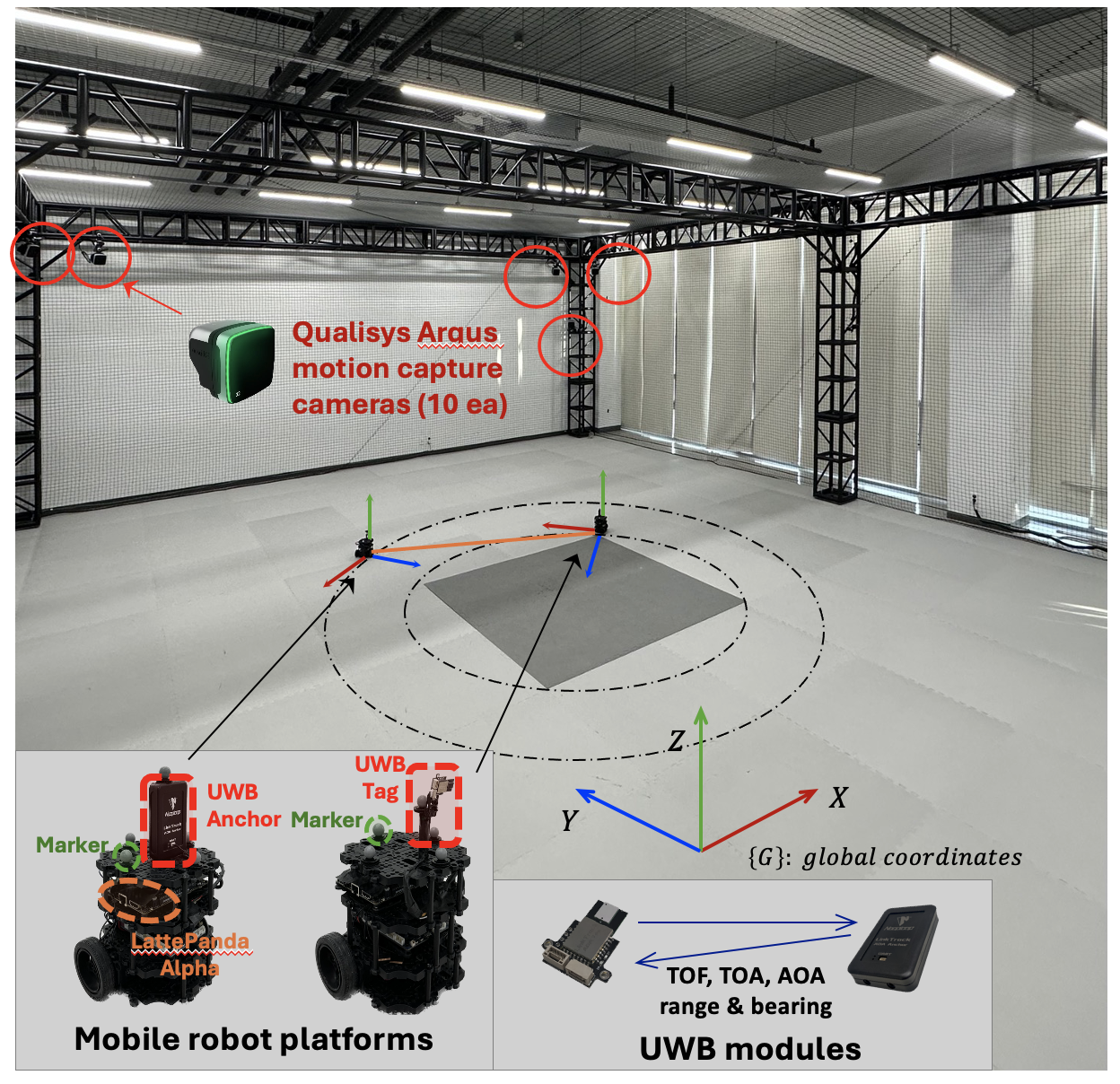}}\vspace{-3.0mm}
	\caption{Photo of the MoCap facility and robot platforms used for hardware experiments.}
	\label{fig:MoCap}
\end{figure}

Furthermore, we conducted a meticulous sensor calibration between UWB and MoCap systems and the outcomes of which are illustrated in Fig.~\ref{fig:UWBcalibration}. Recognizing and adjusting for offsets, excluding noise, between MoCap-tracked data and UWB-AOA measurements are crucial for ensuring high fidelity in our estimation. The least-squares method was employed as a robust technique for refining distance and orientation measurements, considering the following equations:
\begin{equation}
	z_{ij}^{\rho} = \rho_{ji} + \chi^{\rho}(X_{ji}, Y_{ji}) + \nu_{ji}^{\rho}
\end{equation}
where $z_{ij}^{\rho}$ is the range measured by MoCap, $\rho_{ji}$ is the range measured by UWB, and $\nu_{ji}^{\rho}$ is the noise or uncertainty of the range measurements. The regressor $\chi^{\rho}(X_{ji}, Y_{ji})$ is a range calibration factor defined as $\chi^{\rho}(X_{ji}, Y_{ji}) = a_0 + a_1 X_{ji} + a_2 Y_{ji}$ and the parameters are determined using the least-squares method with offline experiments of UWB sensing. 
%\begin{equation}
%\chi^{\rho}(X_{ji}, Y_{ji}) \!=\!\! \begin{bmatrix} 1 & X_{ji}^{(1)} & Y_{ji}^{(1)}  \\ 1 & X_{ji}^{(2)} & Y_{ji}^{(2)}  \\ \vdots & \vdots & \vdots \\ 1 & X_{ji}^{(T)} & Y_{ji}^{(T)}  \end{bmatrix}\!\! \!\!\begin{bmatrix} a_{0} \\ a_{1} \\ a_{2} \\ \end{bmatrix}\!\!
%\end{equation}
Similarly, the bearing measurements can be represented as
\begin{equation}
	z_{ij}^{\beta} = \beta_{ji} + \chi^{\beta}(X_{ji}, Y_{ji}) + \nu_{ji}^{\beta}
\end{equation}
%\begin{equation}
%\chi^{\beta}(X_{ji}, Y_{ji}) \!=\!\! \begin{bmatrix} 1 & X_{ji}^{(1)} & Y_{ji}^{(1)}  \\ 1 & X_{ji}^{(2)} & Y_{ji}^{(2)}  \\ \vdots & \vdots & \vdots \\ 1 & X_{ji}^{(T)} & Y_{ji}^{(T)}  \end{bmatrix}\!\! \!\!\begin{bmatrix} b_{0} \\ b_{1} \\ b_{2} \\ \end{bmatrix}\!\!
%\end{equation}
where $z_{ij}^{\beta}$ is bearing measured by MoCap, $\beta_{ji}$ is range measured by UWB, $\nu_{ji}^{\beta}$ is a noise or uncertainty of bearing, and the regressor $\chi^{\beta}(X_{ji}, Y_{ji})$ is a bearing calibration factor defined as $\chi^{\beta}(X_{ji}, Y_{ji}) = b_0 + b_1 X_{ji} + b_2 Y_{ji}$. The parameters $(b_0, b_1, b_2)$ are determined using the least-squares method with offline experiments of UWB sensing.

For hardware demonstrations, the actual measurements used for the state estimation are
\begin{equation}\label{eq:calib_meas}
\begin{split}
\tilde\rho_{ji} &= \rho_{ji} + \chi^{\rho}(X_{ji}, Y_{ji}) + \nu_{ji}^{\rho}\,, \\
\tilde\beta_{ji} &= \beta_{ji} + \chi^{\beta}(X_{ji}, Y_{ji}) + \nu_{ji}^{\beta}
\end{split}
\end{equation}
where $\rho_{ji}$ and $\beta_{ji}$ are the measurements from the UWB anchor-tag module, and $\chi^{\rho}(\cdot)$ and $\chi^{\beta}(\cdot)$ are the pre-calibrated factors of range and bearing sensing, respectively. Fig~\ref{fig:UWBcalibration} shows a comparison of  before and after calibration in UWB sensor measurements. 

Fig.~\ref{fig:MoCap} is an image of the MoCap experimental environment used for hardware implementation and validation. Ground-truth data were collected using an optoelectronic marker-based motion capture system equipped with ten Qualisys ARQUS A5 cameras. These cameras were synchronized via the Qualisys Track Manager in space and time, providing a unified global frame and a local reference frame at the center of each robot platform. Special attention was given to arranging the placement of marker and sensor modules to comply with prescribed specifications for resolution, focus, and exposure time.

%-----------------------------------------------------------------------------%
\subsection{EKF estimation}
\begin{figure}[!t]
	\centerline{\includegraphics[width=1.\linewidth]{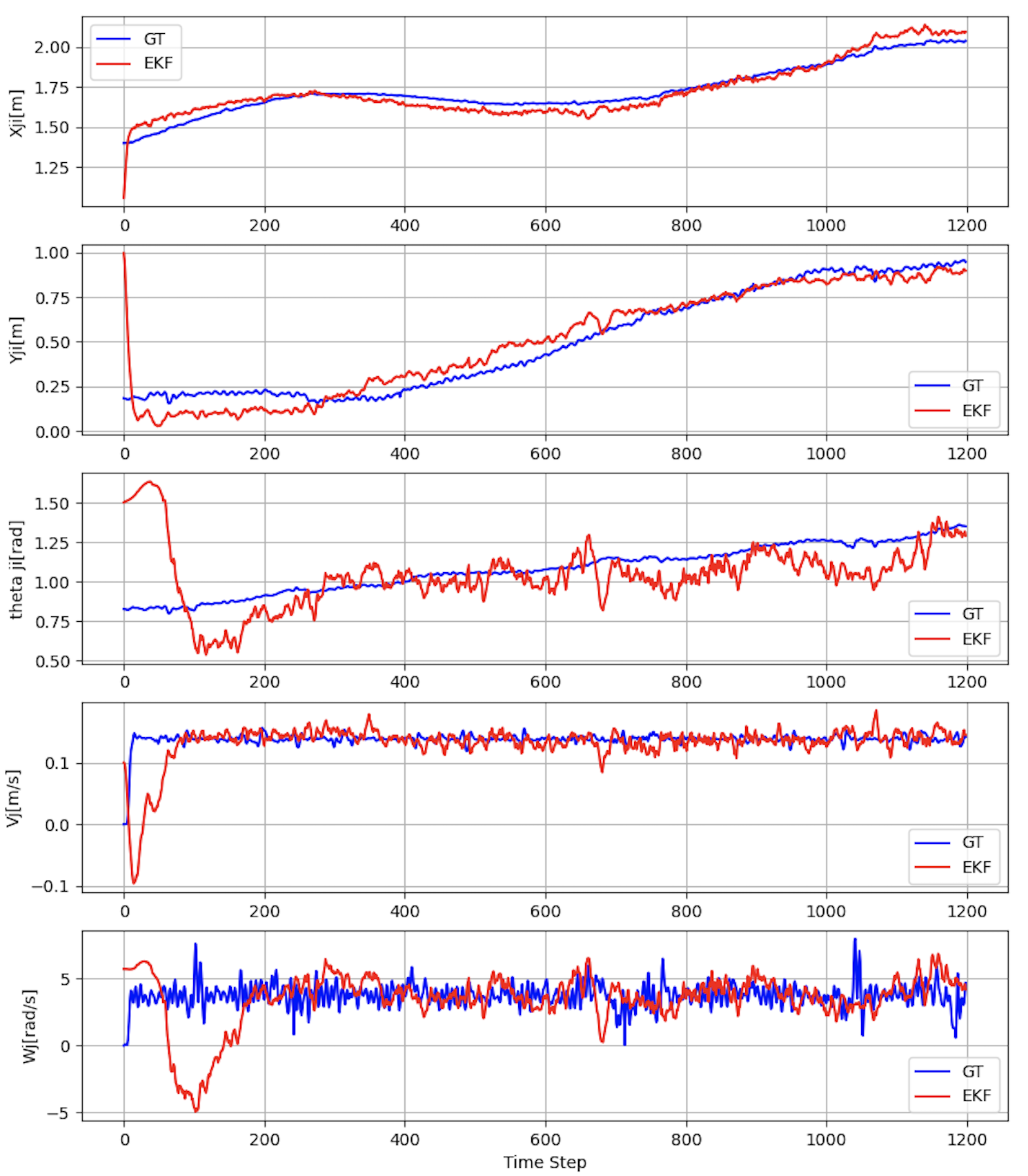}}\vspace{-3.5mm}
	\caption{EKF-based relative pose $(x_{ji}, y_{ji}, \theta_{ji})$, and velocity ($v_j, w_j$) estimation results with Scenario 1.}
	\label{fig:EKF_XjiYjiTjiVjWj_Sc1_1}
\end{figure}

\begin{figure}[!t]
	\centerline{\includegraphics[width=1.\linewidth]{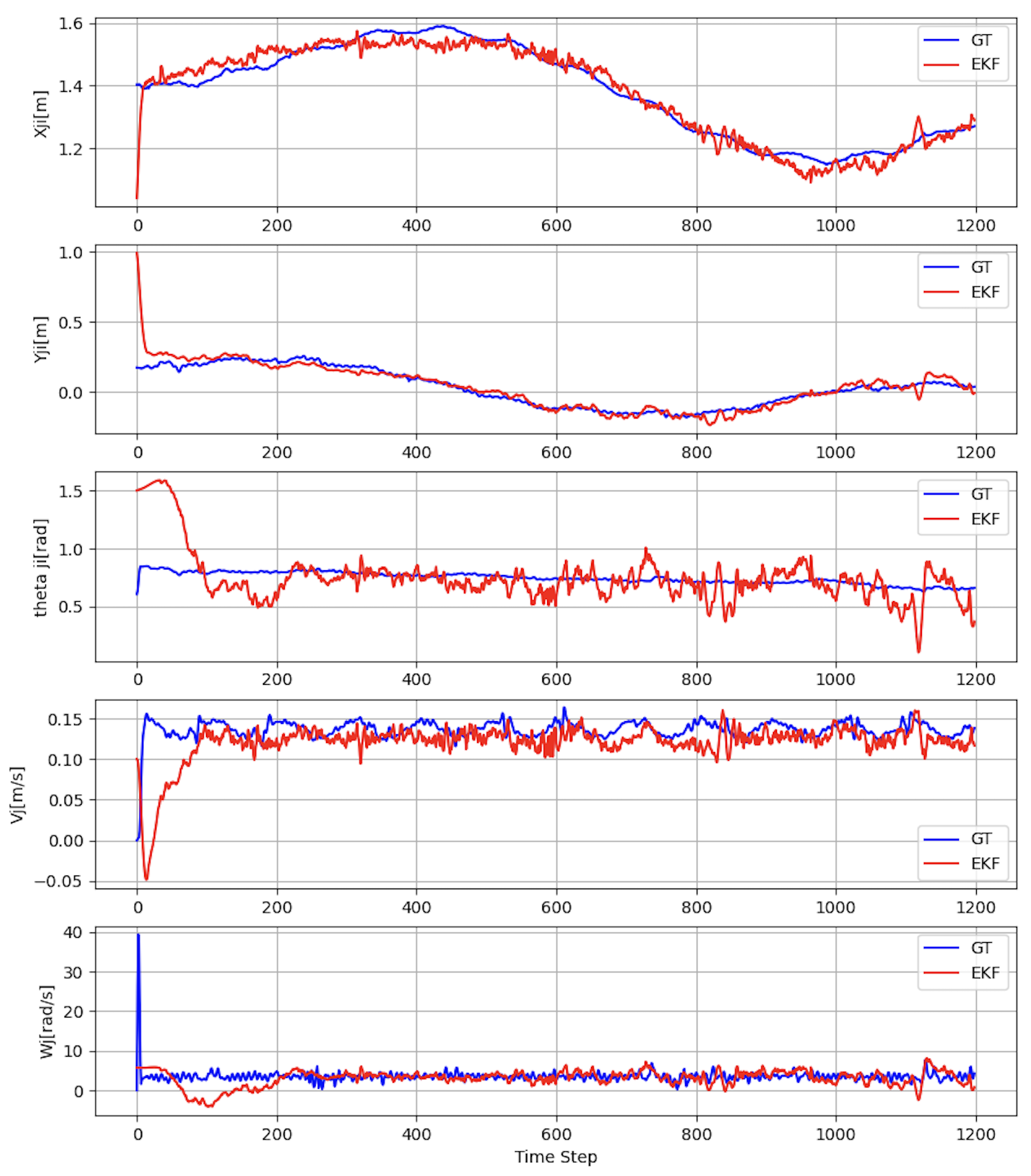}}\vspace{-3.5mm}
	\caption{EKF-based relative pose $(x_{ji}, y_{ji}, \theta_{ji})$ ,and velocity ($v_j, w_j$) estimation results with Scenario 2.}
	\label{fig:EKF_XjiYjiTjiVjWj_Sc2_1}
\end{figure}

%-----------------------------------------------------------------------------%
Hardware experiments were conducted to assess the EKF-based estimation performance in various motion scenarios. In particular, Scenario 1, resembling the simulated environment, provided an opportunity to evaluate the inter-robot relative pose estimation in the real world. Figs.~\ref{fig:EKF_XjiYjiTjiVjWj_Sc1_1} and~\ref{fig:EKF_XjiYjiTjiVjWj_Sc2_1} illustrate the EKF-based estimation results and the corresponding ground truth for both Scenarios 1 and 2. In the hardware implementation, it is evident that the estimation results for both Scenarios 1 and 2 align closely with the expected values, even in the absence of informed communication regarding wheel-odometry data or velocities from the neighboring robot. While reduced estimation precision is noticeable in Scenario 2, where robots move at varying speeds over time, the estimated trajectory closely mirrors the actual values compared to Scenario 1.

\begin{figure}[!t]
	\centerline{\includegraphics[width=1.\linewidth]{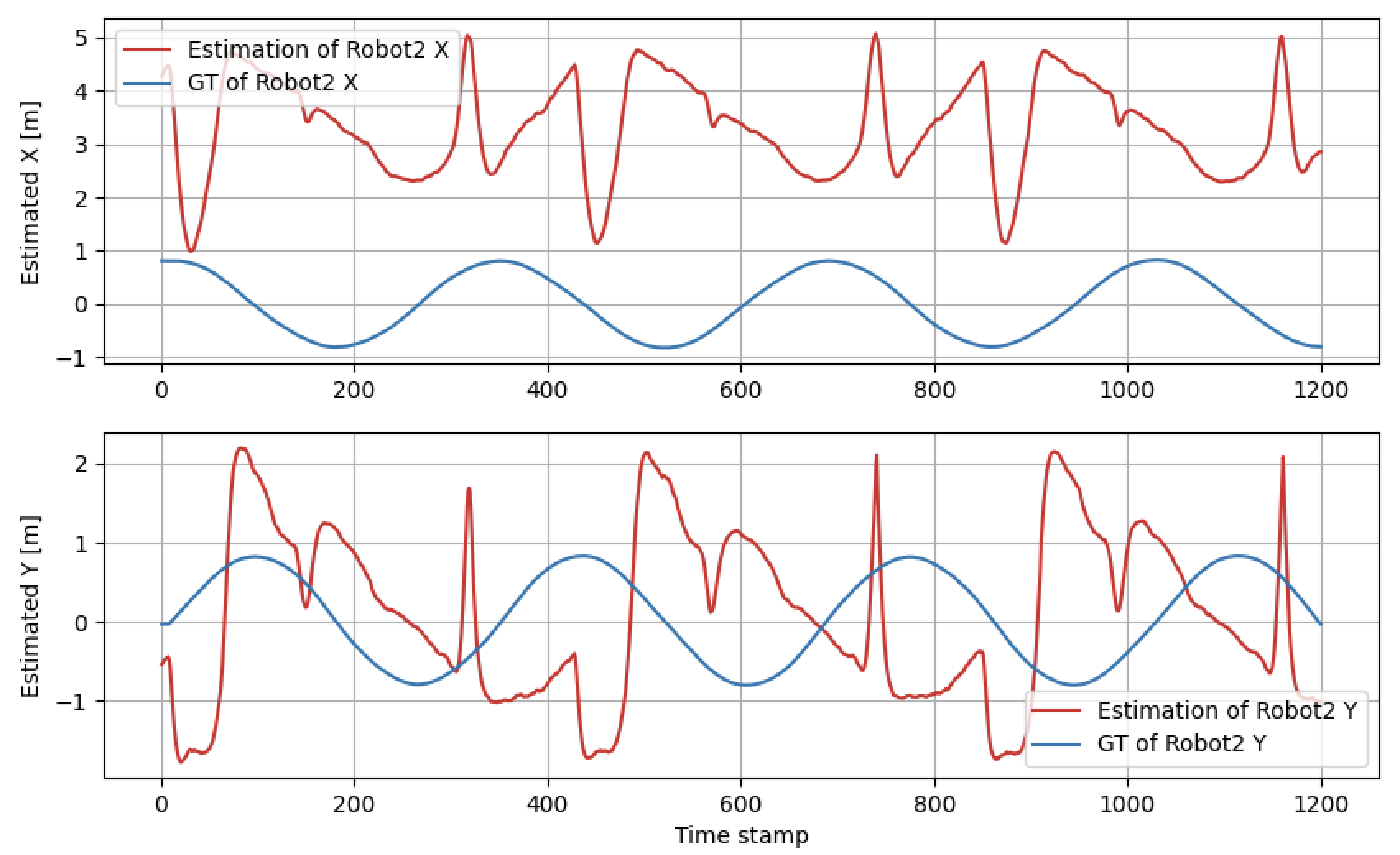}}\vspace{-3.5mm}
	\caption{EKF-based relative position $(x_{ji}, y_{ji})$ estimation results with Scenario 3.}
	\label{fig:EKF_XjiYjiTji_Sc3}
%\end{figure}
\vspace{3mm}
%\begin{figure}[!t]
	\centerline{\includegraphics[width=1.\linewidth]{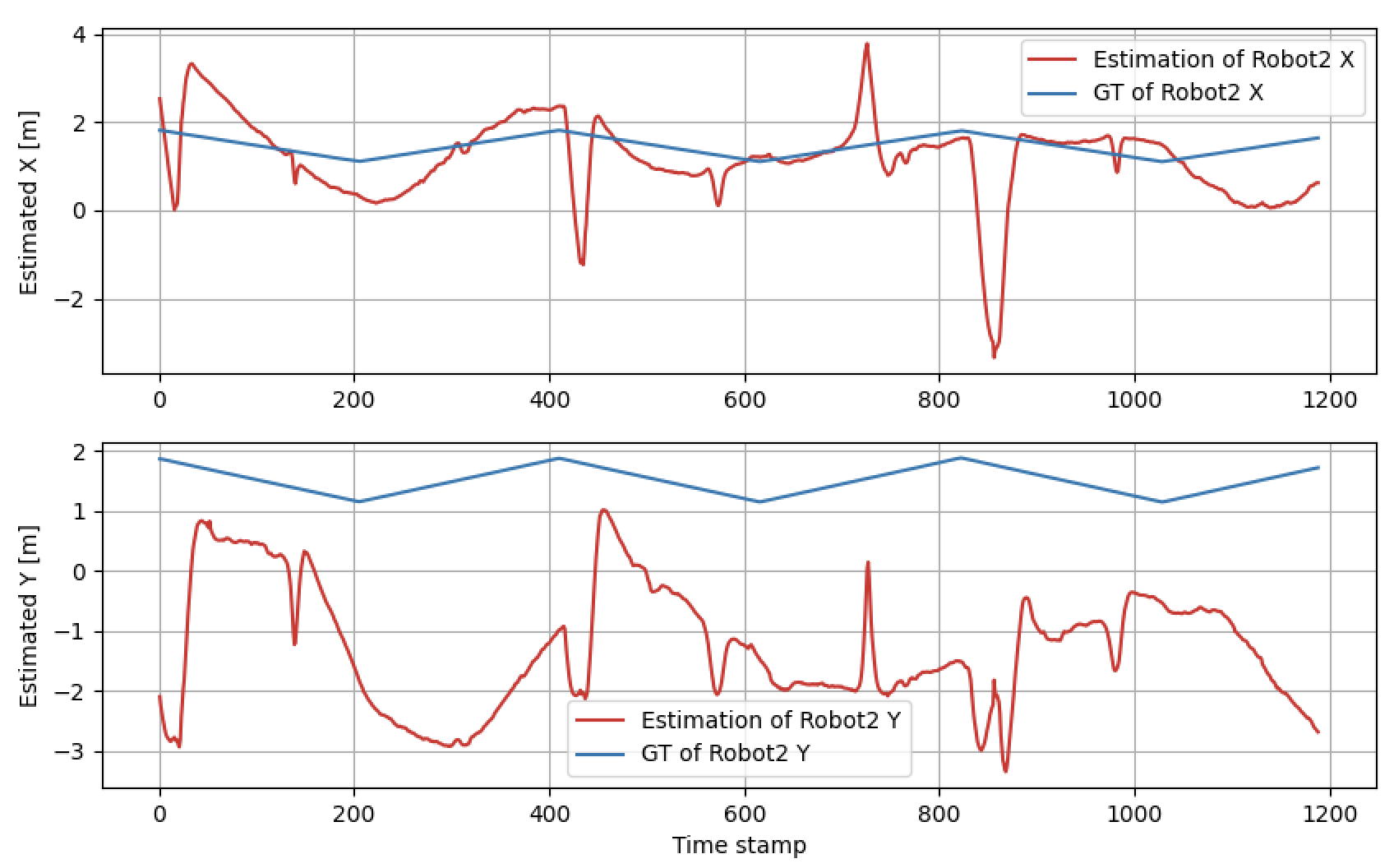}}\vspace{-3.5mm}
	\caption{EKF-based relative position $(x_{ji}, y_{ji})$ estimation results with Scenario 4.}
	\label{fig:EKF_XjiYjiTji_Sc4}
\end{figure}

The estimation algorithm demonstrated resilience to irregular pulses or outliers, possibly arising from uneven surfaces and the robot's vibrations in the experimental space. Table~\ref{tab:xyt RMSE of S1sim and S1hw and S2hw} compares the estimation accuracy using the RMSE values of the EKF-based state estimation for Scenarios 1, 2, and 4 in the simulation study. Although the RMSE for orientation estimation $\theta_{ji}$ was relatively high, the estimation of positions $x_{ji}$ and $y_{ji}$ showed relatively accurate results. In Scenarios 3 and 4, as depicted in Figs.~\ref{fig:EKF_XjiYjiTji_Sc3} and~\ref{fig:EKF_XjiYjiTji_Sc4}, the limitations of the Field of View (FOV) of UWB-AOA delivering measurements of range and bearing adversely affected the accuracy of the estimations. We posit that employing alternative sensors capable of more precise measurements could enhance this accuracy.


\begin{table}[b!]
	\centering
	\caption{RMSE values of state (relative pose in planar motions) estimation for simulation case4 and each scenario of hardware implementation using EKF}\vspace{-2mm}
	\begin{tabular}{|c|c|c|c|}
	\hline
         & $x_{ji}$ & $y_{ji}$ & $\theta_{ji}$ \\[.5mm] 
	\hline
	Simulation Case 4 & 0.0706 & 0.0994 & 0.2971  \\
	Hardware Scenario 1  & 0.0509 & 0.0889 & 0.2142  \\
	Hardware Scenario 2  & 0.0356 & 0.0631 & 0.2061  \\
	\hline
	\end{tabular}
	\label{tab:xyt RMSE of S1sim and S1hw and S2hw}
\end{table}

%-----------------------------------------------------------------------------%
\subsection{Nonlinear least-square estimation}
\begin{figure}[t]
	\centerline{\includegraphics[width=.985\linewidth]{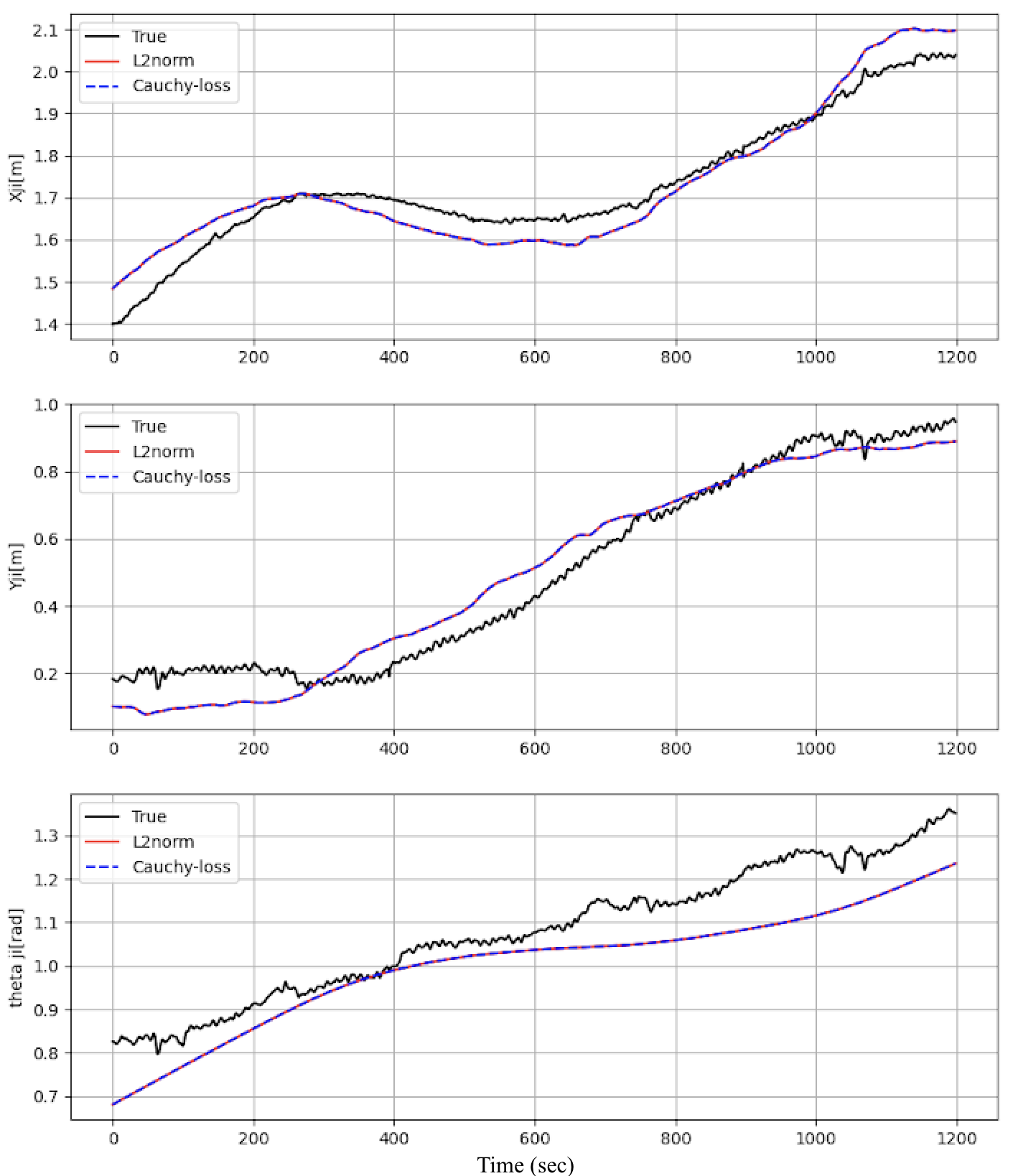}}\vspace{-3.0mm}
	\caption{Relative pose estimation result of Scenario 1 in FB.}
	\label{fig:NLS_XjiYjiTji_FB_Sc1_1}
%\end{figure}
\vspace{3mm}
%\begin{figure}[t]
	\centerline{\includegraphics[width=.985\linewidth]{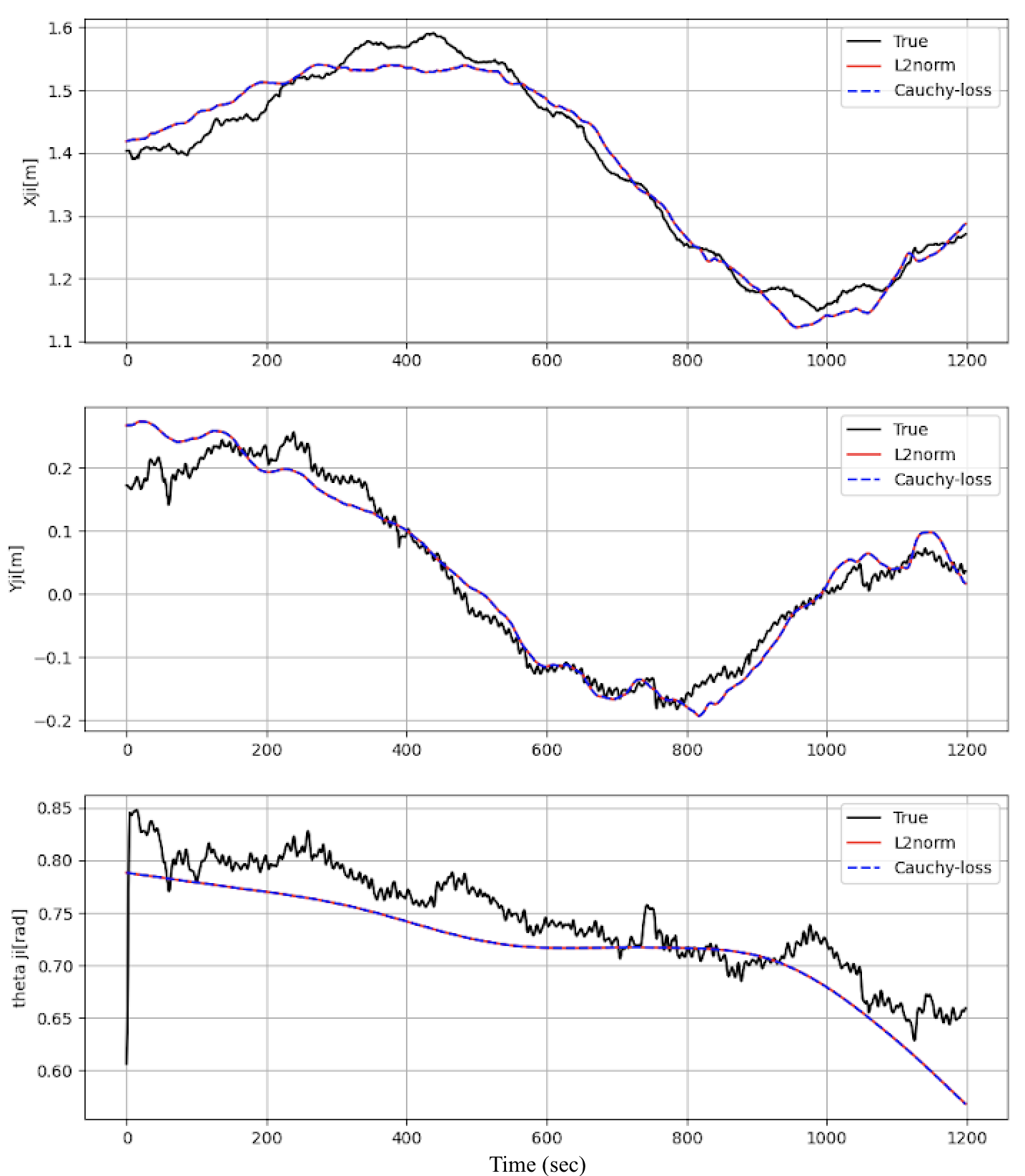}}\vspace{-3.0mm}
	\caption{Relative pose estimation result of Scenario 2 in FB.}
	\label{fig:NLS_XjiYjiTji_FB_Sc2_1}
\end{figure}

%-----------------------------------------------------------------------------%
Similar to the EKF-based estimation scenarios, we conducted a series of nonlinear least-squares (NLS) experiments for Scenarios 1 and 2, utilizing the Ceres Solver for numerical optimization. For the hardware implementation, we applied full-batch methods with L2norm and CauchyLoss kernel functions as robust M-estimation techniques to handle measurement noise. Fig.~\ref{fig:NLS_XjiYjiTji_FB_Sc1_1} and Fig. \ref{fig:NLS_XjiYjiTji_FB_Sc2_1} depict that NLS-based estimation of position $x_{ji}$, $y_{ji}$, and orientation $\theta_{ji}$ gradually demonstrates reduced error and smoother trajectories in both Scenarios 1 and 2 compared to EKF-based estimation. Although the precision of these estimations inherently relies on measurement accuracy, the results exhibit considerable effectiveness across all examined scenarios. Particularly in Scenario 2, it can be observed that NLS-based estimation exhibits strength in handling high deviations from actual values. A comparison of the RMSE values obtained from the computation results of the EKF and the three different NLS optimization batch methods is provided in Table~\ref{tab:S1 and S2 total RMSE of EKF and SF/SB/FB (hw)}.

\begin{table}[b!]
	\centering
	\caption{Total RMSE values of state (relative pose in planar motions) estimation.}\vspace{-2mm}
	\begin{tabular}{|c|c|c|c|c|}
		\hline
		& EKF  & FB with L2norm & FB with Cauchy \\[.5mm]
		\hline
		Hardware Scenario 1  & 0.3540 & 0.2428 & 0.2053 \\
		Hardware Scenario 2  & 0.3048 & 0.0925 & 0.0925 \\
		\hline
	\end{tabular}
	\label{tab:S1 and S2 total RMSE of EKF and SF/SB/FB (hw)}
\end{table}

%==============================================================%
\section{Discussion and Future Directions}\label{sec:discussion}
%==============================================================%
%-----------------------------------------------------------------------------%
\subsection{Uncertainty propagation in distributed data fusion}
\label{sec:disc:2}
%-----------------------------------------------------------------------------%
%\begin{itemize}
%	\item
%	Distributed data fusion (DDF)~\cite{hollinger2014distributed}
%	\item
%	DDF-based multi-robot SLAM~\cite{cunningham2010ddf,cunningham2013ddf}
%	\item
%	Propagation of relative pose estimation errors in multi-robot multi-target tracking~\cite{smith1990,peterson2023motlee}
%\end{itemize}

\begin{figure}[!t]
	\centering
	\includegraphics[width=.925\linewidth]{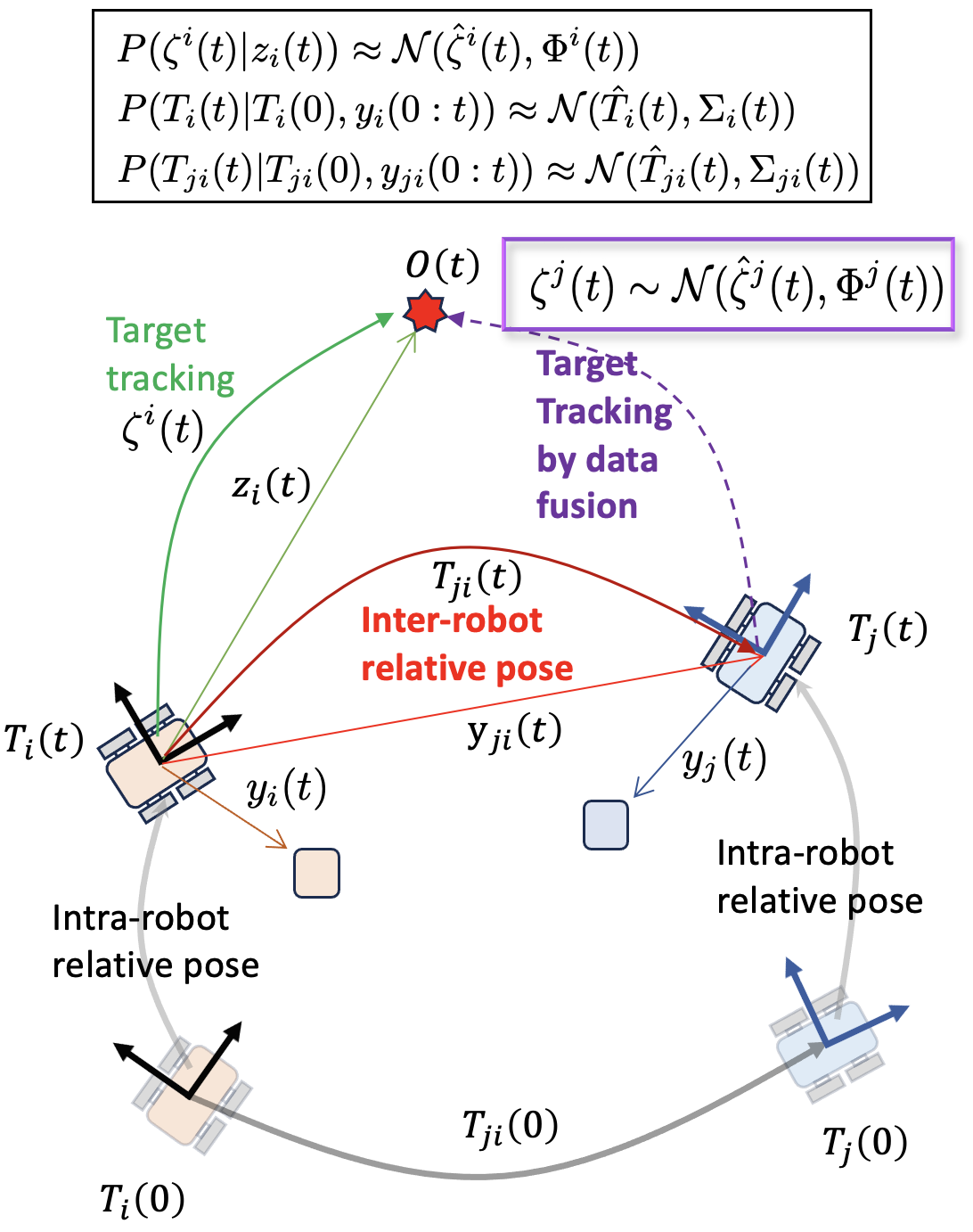}\vspace{-3mm}
	\caption{ The representation and propagation of spatial uncertainty in a two-robot distributed data fusion for single-target tracking are illustrated in which Robot ${\tt R}_{j}$ doesn't need to have a direct observation of the target ${\tt O}$ but uses the transformations of spatial information given in~\eqref{eq:spatial-transform}. The uncertainties of (a) target estimation, (b) intra-robot pose estimation, and (c) inter-robot relative pose estimation propagate through the process of distributed data fusion for target tracking. The propagated accumulative uncertainty in data fusion is quantified and approximated as the covariance $\Phi^{j}!(t) = \text{C}(\zeta^{j}!(t))$ at each time-step $t$.}
	\label{fig:spatial-uq} 
\end{figure}

Another important application of inter-robot relative pose estimation is distributed data fusion (DDF).
that can be used for 
\begin{itemize}
\item
Collaborative multi-robot target tracking~\cite{peterson2023motlee}, where the objective is to estimate the state of one or multiple targets by merging data from a team of robots;
\item
Collaborative multi-robot SLAM~\cite{cunningham2010ddf,cunningham2013ddf}, aiming to perform localization estimates and mapping of a mission space collectively;
\item
Collaborative multi-robot localization~\cite{fox2000probabilistic,roumeliotis2004propagation,carrillo2013decentralized}, encompassing intra-robot (time-to-time or event-to-event) and inter-robot (robot-to-robot) relative pose estimation; and
\item
Multi-robot SLAM utilizing inter-robot loop closure~\cite{Boroson2020,Liu2022}, employing the relative pose (loop closure) between two robots by minimizing UWB ranging, enabling robots to localize and build a map collaboratively even without visual loop closures.
\end{itemize}

Mapping, by definition, is the process of constructing estimates of object locations or poses concerning the world or local frame. For clarity and simplicity, we consider a collaborative two-robot single-static-target tracking problem. Suppose there are probabilistic estimates of the absolute pose of Robot ${\tt R}_{i}$, denoted as $T_{i} = (x_{i}, y_{i}, \theta_{i}) \sim \mathcal{N}(\hat{T}_{i} , \Sigma_{i})$, and the target location or pose, $\zeta^{i} = ( \zeta_{x}^{i}, \zeta_{y}^{i}, \zeta_{\theta}^{i}) \sim \mathcal{N}(\hat{\zeta}^{i} , \Phi^{i})$, represented in Robot ${\tt R}_{i}$'s local coordinates.

%
%where the target is considered as a point-mass such that there is no orientation specifications, but analysis results presented in this section does not change for more general target pose estimation and the associated uncertainty propagation. 
%
Probabilistic estimates need not be Gaussian, but our focus lies on uncertainty quantification and propagation by approximating the first and second moments of random variables (positions or poses). Additionally, we assume a probabilistic estimate of the relative pose of Robot ${\tt R}_{j}$ with respect to ${\tt R}_{i}$ as $T_{ji} = (x_{ji}, y_{ji}, \theta_{ji}) \sim \mathcal{N}(\hat{T}_{ji} , \Sigma_{ji})$.

Fig.~\ref{fig:spatial-uq} illustrates a team of two mobile robots employing DDF to track a single target. Assuming Robot ${\tt R}_{j}$ lacks a direct measurement or estimation of the target, Robot ${\tt R}_{i}$ shares the spatial information of target ${\tt O}$, and Robot ${\tt R}_{j}$ transforms this spatial information into its own local coordinates by considering the inter-robot relative pose $T_{ji}$ (or $T_{ij}$). 
In DDF-based target tracking, the primary challenge is to compute or estimate the probability distribution of the target pose $\zeta^{j}$ in the local coordinates of Robot ${\tt R}_{j}$. In other words, the goal of DDF is to compute the transformation of spatial information, defined as 
\begin{equation}\label{eq:spatial-transform}
\zeta^{j} = \ominus T_{ji} \oplus \left( T_{i} \oplus \zeta^{i} \right)
\end{equation}
where $T_{i} \oplus \zeta^{i}$ is a head-to-tail pose composition, $\ominus T_{ji}=T_{ij}$ is the inverse transformation, and $\ominus T_{ji} \oplus \tilde{\zeta}^{i}$ is tail-to-tail pose composition with $\tilde{\zeta}^{i} := T_{i} \oplus \zeta^{i}$. They define spatial relationships, that is, the relationships between spatial information and variables.

As shown in Fig.~\ref{fig:spatial-uq}, given the first and second moments of the uncertain spatial variables $(T_{ji}, T_{i},  \zeta^{i})$, approximating the first and second moments of the transformed data $\zeta^{j}$ for which the mean and covariance are estimated is possible as follows ~\cite{smith1986representation,smith1990}:
\begin{equation}\label{eq:spatial-uncertainty-prop}
\begin{split}
\hat{\zeta}^{j} 
& \approx 
\ominus \hat{T}_{ji} \oplus \left( \hat{T}_{i} \oplus \hat{\zeta}^{i} \right) 
\\
\text{C}({\zeta}^{j}) 
& \approx 
J_{\oplus} \left[\begin{array}{@{}c@{\,\,}c@{}} J_{\ominus} & 0 \\ 0 & I \end{array}\right]
\left[\begin{array}{@{}c@{\,\,}c@{}} \Sigma_{ji} & 0 \\ 0 & \text{C}(\tilde{\zeta}^{i}) \end{array}\right]
\left[\begin{array}{@{}c@{\,\,}c@{}} J_{\ominus} & 0 \\ 0 & I \end{array}\right]^{\top}\!\! J_{\oplus}^{\top}
\end{split}
\end{equation}
where $\hat{\zeta}$ and $C(\zeta)$ denote the mean and covariance of a random variable $\zeta$, respectively. The tively and the compound relation and associated uncertainty or error propagation can be computed as
\[
\begin{split}s
\tilde{\zeta}^{i} 
& =\! \begin{bmatrix} \tilde{\zeta}_{x}^{i} \\ \tilde{\zeta}_{y}^{i} \\ \tilde{\zeta}_{\theta}^{i}\end{bmatrix} \!\!
= T_{i} \oplus \zeta^{i} =\! \begin{bmatrix} \zeta_{x}^{i} \cos \theta_{i} - \zeta_{y}^{i} \sin \theta_{i} + x_{i} \\ 
\zeta_{x}^{i}\sin \theta_{i} + \zeta_{y}^{i} \sin \theta_{i} + y_{i} \\
\theta_{i} + \zeta_{\theta}^{i}  \end{bmatrix} , \\
\text{C}(\tilde{\zeta}^{i}) 
& = \text{C}(T_{i} \oplus \zeta^{i}) \approx 
\tilde{J}_{\oplus} \left[\begin{array}{@{}c@{\,\,}c@{}} \Sigma_{i} & 0 \\ 0 & \Phi^{i} \end{array}\right] \tilde{J}_{\oplus}^{\,\top}
\end{split}
\]
with Jacobians given as 
\[
\begin{split}
\tilde{J}_{\oplus}
& =\! \frac{\partial \tilde{\zeta}^{i}}{\partial (T_{i}, \zeta^{i})} 
=\! \left[\begin{array}{@{}c@{\,\,\,\,\,\,}c@{\,\,}c@{\,\,}c@{\,\,}c@{\,\,}c@{}} 
1 & 0 & -(\zeta_{y}^{i} - y_{i}) & \cos\theta_{i} & -\sin\theta_{i} & 0 \\
0 & 1 & (\zeta_{x}^{i} - x_{i}) & \sin\theta_{i} & \cos\theta_{i} & 0 \\
0 & 0 & 1 & 0 & 0& 1
\end{array}\right] , \\
{J}_{\ominus}
& =\! \frac{\partial T_{ij}}{\partial T_{ji}}
= \! \left[\begin{array}{@{}c@{\,\,}c@{\,\,}c@{}}
-\cos\theta_{ji} & -\sin\theta_{ji} & y_{ji} \\
\sin\theta_{ji} & -\cos\theta_{ji} & -x_{ji} \\
0 & 0 & -1
\end{array}\right] , \\
{J}_{\oplus} & =\! \frac{\partial \zeta^{j}}{\partial (T_{ji},\tilde{\zeta}^{j})}
=\! \left[\begin{array}{@{}c@{\,\,\,\,\,\,}c@{\,\,}c@{\,\,}c@{\,\,}c@{\,\,}c@{}} 
1 & 0 & -(\tilde\zeta_{y}^{i} - y_{ji}) & \cos\theta_{ji} & -\sin\theta_{ji} & 0 \\
0 & 1 & (\tilde\zeta_{x}^{i} - x_{ji}) & \sin\theta_{ji} & \cos\theta_{ji} & 0 \\
0 & 0 & 1 & 0 & 0& 1
\end{array}\right] \\
\end{split}
\]
evaluated at the mean values of the spatial representations $(\hat{T}_{ji}, \hat{T}_{i},  \hat{\zeta}^{i})$.

\begin{remark}[Uncertainty propagation]
From the previous computations that are crucial for a robust DDF, one can see the importance of the inter-robot relative pose estimation $T_{ji} = (x_{ji}, y_{ji}, \theta_{ji})$ as well as the intra-robot relative pose estimation $T_{i}=(x_{i}, y_{i}, \theta_{i})$. In addition, spatial uncertainty of a transformed information $\zeta^{j}$ quantified as its covariance $C(\zeta^{j})$ can be approximated as a function of the mean $(\hat{T}_{ji}, \hat{T}_{i},  \hat{\zeta}^{i})$ and the covariance $(\Sigma_{ji}, \Sigma_{i}, \Phi^{i})$, as shown in~\eqref{eq:spatial-uncertainty-prop}.
\end{remark}

\begin{remark}[Symmetries and Perturbation Map]
In the realm of distributed data fusion (DDF), the exploration extends to a probabilistic representation encompassing both mobile robot localization and map features with interdependent relationships, known as the symmetry and perturbation map (SPmap)~\cite{castellanos2012mobile} in multi-sensor SLAM literature.
\end{remark}

\begin{remark}[KF-based probabilistic inference vs. NLP-based PGO]
Essentially, robot state estimation considerably benefits from methods that explicitly and accurately consider uncertainty. Techniques based on Kalman filter (KF) recursive probabilistic inference, such as Error-State EKF~\cite{sola2017quaternion} and Invariant EKF~\cite{barrau2018invariant}, Unscented Kalman Filter (UKF)\cite{hauberg2013unscented,brossard2020code}, and Particle Filter (PF)\cite{zhang2017feedback,li2020unscented} on manifolds, inherently compute estimation uncertainty as the covariance matrix. In contrast, Nonlinear Programming (NLP)-based Pose Graph Optimization (PGO) methods, like those used in our approach, do not explicitly compute uncertainty quantification. Nonetheless, they exhibit robustness against outliers, as discussed in Section~\ref{sec:disc:3} and demonstrated in simulation and hardware experiments (Sections~\ref{sec:gazebo} and \ref{sec:hardware}). Future work should explore an iterative EKF that incorporates robust kernel functions in the correction step~\cite{tao2023}, extending it to robot state estimation on manifolds. This could offer a promising alternative solution, combining the advantages of KF-based recursive probabilistic inference and NLP-based (semi-)batch PGO, particularly concerning uncertainty quantification and propagation in robot state estimation and spatial information.
\end{remark}

%-----------------------------------------------------------------------------%
\subsection{Outlier robust relative pose estimation}
\label{sec:disc:3}
%-----------------------------------------------------------------------------%
%\begin{itemize}
%	\item
%	Robust PGO using M-estimation~\cite{bosse2016robust,barron2019general}
%	\item
%	Adaptive PGO~\cite{agamennoni2015self,pfeifer2019expectation,pfeifer2021advancing,ramezani2022aeros}
%	\item 
%	Smoothing with sliding window measurements (partial information) or full batch measurements (full information)
%	\item
%	Moving horizon estimation (MHE) with dynamic programming (DP)~\cite{Allan2019}: Cost-to-arrive (aka Cost-to-come)
%	\item
%	RANSAC + KF~\cite{civera20101,s17102164}
%	\item
%	Iterative EKF with robust loss functions~\cite{Jazwinski2007,skoglund2015,tao2023} of M-estimation that adopts the iteratively reweighted least squares (IRLS) optimization framework in the correction step for being outlier-robust
%\end{itemize}

The presence of outliers in sensor measurements can considerably degrade estimation performance. As demonstrated in Section~\ref{sec:gazebo}, nonlinear least-squares methods and vanilla EKF are susceptible to outliers, while PGO using M-estimation of robust cost functions~\cite{bosse2016robust,barron2019general} provides high-fidelity and reliable robotic state estimation. The performance of M-estimators depends on tuning parameters in a robust cost function, necessitating the development of adaptive tuning methods for solving iteratively reweighted least squares (IRLS) optimization. Several adaptive M-estimator parameter-tuning strategies have been proposed~\cite{agamennoni2015self,pfeifer2019expectation,pfeifer2021advancing,ramezani2022aeros}.

Beyond the choice of M-estimator and its tuning parameters, the time-horizontal size of processing measurements concurrently influences accuracy performance and computing time in PGO-based estimation, given its smoothing nature. To balance accuracy and computing time, robust PGO-based smoothing can consider either sliding-window measurements (partial information) or full-batch measurements (full information). However, selecting an appropriate sliding window size can be challenging and often involves a trial-and-error process. Alternatively, adopting the ideas of dynamic programming-based moving horizon estimation (MHE)~\cite{Allan2019,rao2003constrained,rawlings2012optimization} can provide better performance with a short sliding window horizon for robust PGO-based smoothing in robot state estimation.

In addition to PGO integrating M-estimators, random sample consensus (RANSAC) stands out as a popular iterative method for estimating unknown parameters and system states from outlier-contaminated measurement data. Past studies~\cite{civera20101,s17102164} combined RANSAC with KF for robust state estimation of dynamical systems. Recently, an iterative EKF incorporating robust loss functions of M-estimation, adopting the IRLS optimization framework in the correction step for outlier robustness, has been proposed~\cite{Jazwinski2007,skoglund2015,tao2023}. This approach investigates the connection between the Kalman gain and Gauss-Newton iterations, providing another avenue for robust state estimation in the presence of outliers.

%-----------------------------------------------------------------------------%
\subsection{Range-aided multi-robot state estimation}
\label{sec:disc:4}
%-----------------------------------------------------------------------------%

%% range-aided robot state estimation 
%\begin{itemize}
%	\item
%	PGO-based Range-Aided SLAM (RA-SLAM)~\cite{torres2018range,funabiki2020range,lajoie2020door,boroson2020inter,rodrigues2021online}, 
%	\item
%	PGO-based Certifiable RA-SLAM~\cite{papalia2023certifiably,papalia2023score,dumbgen2023toward,holmes2023semidefinite,goudar2023optimal}, 
%	\item
%	PGO-based RA-SLAM using UWB ranging~\cite{wang2017ultra,song2019uwb,nguyen2021viral,liu2022distributed},
%	\item
%	MSCKF-based RA localization using UWB ranging~\cite{chenchana2018range,jia2022fej},
%	\item
%	Camera-IMU-UWB-based robot state estimation (localization~\cite{nguyen2021range,nguyen2021flexible,zhang2022lidar,nguyen2022ntu,delama2023uvio,wang2023rvio} and indoor navigation~\cite{hu2023robust,hu2023tightly,lin2023gnss,goudar2023range}),
%	\item
%	PF-based multi-robot SLAM using inter-robot range and bearing measurements exchanged over ad-hoc communication networks
%\end{itemize}

Recently, considerable attention has focused on range-aided (RA) multi-robot state estimation, where inter-robot range measurements play a pivotal role in cooperative localization, mapping, and Simultaneous Localization and Mapping (SLAM). Particle filter (PF)-based methods for multi-robot SLAM, utilizing inter-robot range and bearing measurements exchanged over ad hoc communication networks, have been introduced in~\cite{carlone2010rao,carlone2011simultaneous,saeedi2016multiple}. RA cooperative localization using a multistate constrained Kalman filter (MSCKF) with Ultra-Wideband (UWB) range measurements is explored in~\cite{chenchana2018range,jia2022fej}. Range-Aided SLAM (RA-SLAM) based on Pose Graph Optimization (PGO)~\cite{torres2018range,funabiki2020range,lajoie2020door,boroson2020inter,rodrigues2021online} has gained popularity in multi-robot SLAM.

In RA-SLAM, the nonconvex nonlinear optimization in the back-end can be relaxed to second-order conic programming (SOCP) or semidefinite programming (SDP)-based convex optimization. Certifiable RA-SLAM is introduced, offering a lower bound and a suboptimal solution for the original RA-SLAM~\cite{papalia2023certifiably,papalia2023score,dumbgen2023toward,holmes2023semidefinite,goudar2023optimal}. This suboptimal solution serves as an initial estimate for an iterative method to solve the original back-end nonlinear least-squares optimization. Commonly used UWB devices and signals for inter-robot range measurements in RA-SLAM are discussed in~\cite{wang2017ultra,song2019uwb,nguyen2021viral,liu2022distributed}. Additionally, Camera-IMU-UWB-based sensor fusion is considered for indoor localization~\cite{nguyen2021range,nguyen2021flexible,zhang2022lidar,nguyen2022ntu,delama2023uvio,wang2023rvio} and navigation~\cite{hu2023robust,hu2023tightly,lin2023gnss,goudar2023range}.

For cooperative multi-robot range-aided localization using the PGO framework, the associated \emph{centralized} manifold optimization is defined as follows:
\begin{equation}\label{eq:PGO_RA-SLAM:central}
\begin{split}
\mbox{min} & \sum_{(i_{t},j_{s})\in\mathcal{E}} 
 w^{\rm rot}_{j_{s}i_{t}} \! \| R_{j_{s}} \!\!-\! R_{i_{t}} \hat{R}_{j_{s}i_{t}}\! \|_{\rm F}^{2} \\[-3mm]
& \qquad\qquad + w^{\rm tran}_{j_{s}i_{t}}\! \| \tau_{j_{s}} \!\!-\! \tau_{i_{t}} \!\!-\! R_{i_{t}} \hat{\tau}_{j_{s}i_{t}}\! \|_2^2 \\[1mm]
& \qquad\qquad + w^{\rm rang}_{j_{s}i_{t}} ( \|\tau_{j_{s}} \!\!-\! \tau_{i_{t}}\|_2 - \tilde {\rho}_{j_{s}i_{t}})^2 \\[2mm]
\mbox{s.t.} &\ \  R_{i_{t}} \in \textrm{SO}(d), \, \tau_{i_{t}} \!\in \mathbb{R}^{d}, \, \forall i, \, \forall t 
\end{split}
\end{equation}
where both inter-robot spatial relations are encoded in a graph $\mathcal{G} = (\mathcal{V}, \mathcal{E})$. Each variable node in $\mathcal{V}$ corresponds to a single pose $T_{i_{t}}=(R_{i_{t}},\tau_{i_{t}}) \in \text{SE}(d)$ owned by a robot ${\tt R}_{i}$ at time $t$ and an edge $(i_{t},j_{s}) \in \mathcal{E}$ is formed, if there is relative spatial information between robots such as relative pose estimates $\hat{T}_{j_{s}i_{t}} = (\hat{R}_{j_{s}i_{t}}, \hat{\tau}_{j_{s}i_{t}})$ or range measurements $\tilde{\rho}_{j_{s}i_{t}}$ from $T_{i_{t}}$ to $T_{j_{s}}$.

Central manifold optimization~\eqref{eq:PGO_RA-SLAM:central} can be decomposed into the following distributed incremental optimization for each robot ${\tt R}_{i}$:
%%
%\begin{equation}\label{eq:PGO_RA-SLAM:async}
%\begin{split}
%\mbox{min} &\!\!\!\! \sum_{(i_{t},j_{s})\in\mathcal{E}} 
%\!\! w^{\rm rot}_{i_{t},j_{s}} \! \| R_{j_{s}}^{-} \!\!-\! (R_{i_{t}}^{-} \boxplus \Delta R_{i_{t}} ) \hat{R}_{j_{s}i_{t}}\! \|_{\rm F}^{2} \\[-3mm]
%& \qquad\quad + w^{\rm tran}_{i_{t},j_{s}}\! \| \tau_{j_{s}}^{-} \!\!-\! (\tau_{i_{t}}^{-} \boxplus \Delta \tau_{i_{t}} ) \!\!-\! (R_{i_{t}}^{-} \boxplus \Delta R_{i_{t}} )\hat{\tau}_{j_{s}i_{t}}\! \|_2^2 \\[1mm]
%& \qquad\quad + w^{\rm rang}_{i_{t},j_{s}} ( \|\tau_{j_{s}}^{-} \!\!-\! (\tau_{i_{t}}^{-} \boxplus \Delta \tau_{i_{t}} )\|_2 - \tilde {\rho}_{j_{s}i_{t}})^2 \\[2mm]
%\mbox{s.t.} &\ \  R_{i_{t}} \in \textrm{SO}(d), \, \tau_{i_{t}} \!\in \mathbb{R}^{d}, \, \forall i, \, \forall t 
%\end{split}
%\end{equation}
%%
\begin{equation}\label{eq:PGO_RA-SLAM:async}
\begin{split}
\mbox{min} & \sum_{\{j_{s}:(i_{t},j_{s})\in\mathcal{E}\}}
 w^{\rm rot}_{j_{s}i_{t}} \! \| R_{j_{s}}^{+} \!\!-\! R_{i_{t}} \hat{R}_{j_{s}i_{t}}\! \|_{\rm F}^{2} \\[-3mm]
& \qquad\qquad\qquad + w^{\rm tran}_{j_{s}i_{t}}\! \| \tau_{j_{s}}^{+} \!\!-\! \tau_{i_{t}} \!\!-\! R_{i_{t}} \hat{\tau}_{j_{s}i_{t}}\! \|_2^2 \\[1mm]
& \qquad\qquad\qquad + w^{\rm rang}_{j_{s}i_{t}} ( \|\tau_{j_{s}}^{+} \!\!-\! \tau_{i_{t}}\|_2 - \tilde {\rho}_{j_{s}i_{t}})^2 \\[2mm]
\mbox{s.t.} &\  R_{i_{t}} \!\!=\! R_{i_{t}}^{-} \boxplus \delta R_{i_{t}} \!\in\! \textrm{SO}(d), \, \tau_{i_{t}} \!\!=\! \tau_{i_{t}}^{-} \boxplus \delta \tau_{i_{t}} \!\in\! \mathbb{R}^{d}, \, \forall t \\
&\ \|\delta R_{i_{t}}\|_{\rm F} \leq \epsilon_{R}, \, \|\delta \tau_{i_{t}}\| \leq \epsilon_{\tau} 
\end{split}
\end{equation}
where $d\in\{2,3\}$ refers to the dimension and $\boxplus : \mathcal{M}(d) \times \mathbb{R}^{d} \rightarrow  \mathcal{M}(d)$ defines an infinitesimal addition preserving the corresponding manifold structure, $(\epsilon_{R},\epsilon_{\tau}) >0$ characterize the trust region optimization (TRO) for updates of orientation and translation variables, respectively. The superscripts $(\cdot)^{-}$ and $(\cdot)^{+}$ denote the prior and post-updates of the variables, respectively.

It is assumed that a neighboring robot ${\tt R}_{j}$ with $j \in \mathcal{N}_{i}$ updated his or her poses over time and shared the current guess of these pose estimates $T_{j_{s}}^{+} = (R_{j_{s}}^{+}, \tau_{j_{s}}^{+})$ for all time $s$ with Robot ${\tt R}_{i}$. This is only for the clarity of presentation, and the updates can be either synchronous or asynchronous. Once the optimization~\eqref{eq:PGO_RA-SLAM:async} is solved with a solution $(\delta R_{i}^{*}, \delta \tau_{i}^{*})$, then it is updated as $R_{i}^{+} \leftarrow R_{i}^{-}\boxplus \delta R_{i}^{*}$ and $\tau_{i}^{+} \leftarrow \tau_{i}^{-} \boxplus \delta \tau_{i}^{*}$.
Note that the distributed TRO on manifolds can be equivalently rewritten as a Euclidean optimization with a proper parameterization of the perturbed rotation matrix $\delta R_{i} \in \mathbb{R}^{d}$.

%-----------------------------------------------------------------------------%
\subsection{Formation control using the inter-robot relative pose estimation}
\label{sec:disc:5}
%-----------------------------------------------------------------------------%
\begin{figure}[!t]
	\centering
	\includegraphics[width=.925\linewidth]{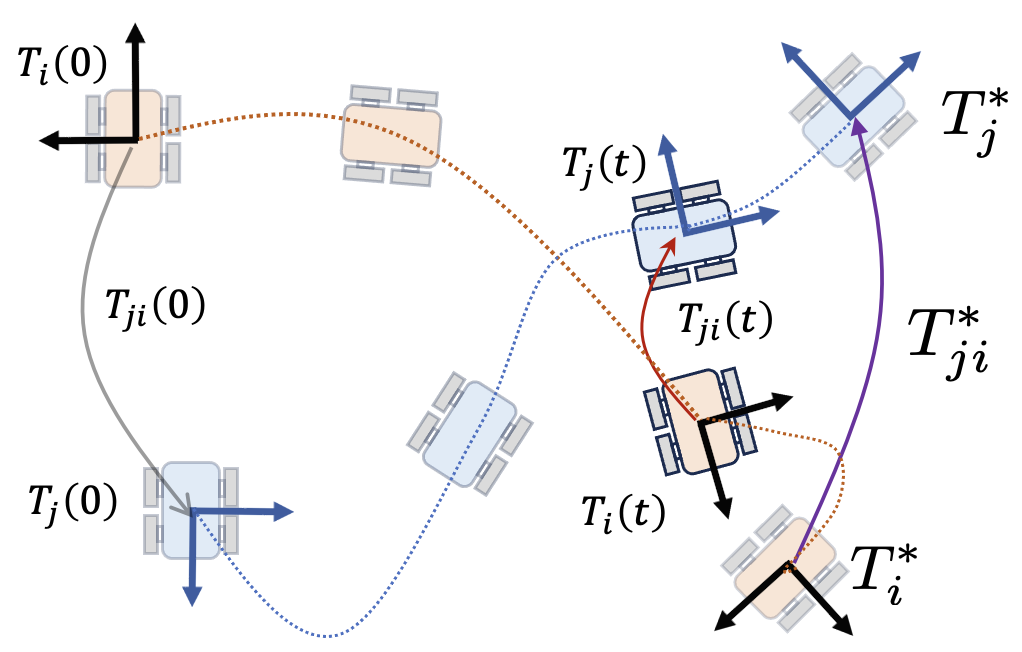}\vspace{-3mm}
	\caption{Formation control of two-robot to achieve the desired absolute pose $T_{i}^{*}$ ($T_{j}^{*}$) and the desired relative pose $T_{ji}^{*}$ ($T_{ij}^{*}$) for Robot ${\tt R}_i$ (${\tt R}_j$).}
	\label{fig:formation} 
\end{figure}

In the realm of cooperative coordination for mobile multi-robot systems, significant industrial and academic efforts have been directed towards enhancing efficiency, robustness, scalability, and reliability across various applications. These applications include sensor network localization, search and rescue, object detection and tracking, environment mapping, and surveillance.
Cooperative coordination problems can be categorized based on the desired configuration of a multi-robot system. These categories encompass (a) position-based, (b) relative position-based, (c) range-based, (d) bearing-based, and (e) orientation-based formation control. This paper focuses on the most general coordination problem, absolute and relative pose-based formation control, which subsumes specific cases of formation problems (a)$\sim$(e). For a comprehensive overview of multi-robot formation control, readers are directed to recent monographs~\cite{sakurama2021generalized} and reviews~\cite{verma2021multi}. 

%-------------------%
\subsubsection{Estimation-based output feedback formation control}
%-------------------%
Using the sensing measurement $y_{ji}(t) = h_{ji}(T_{ji}(t))$ related with the inter-robot relative pose, the formation control input of the robot ${\tt R}_{i}$ can be represented as a function of the relative measurement
\[
u_{ji}(t) = \pi_{ji}(y_{ji}(t)-y_{ji}^{*},t)
\]
or the relative pose estimate $\hat{T}_{ji}(t)$
\[
u_{ji}(t) = \pi_{ji}(\hat{T}_{ji}(t)-{T}_{ji}^{*},t)
\]
where $y_{ji}^{*}$ and ${T}_{ji}^{*}$ define the desired configurations of robots ${\tt R}_{i}$ and ${\tt R}_{j}$ in the ${\tt R}_{i}$'s local frame that shown in Fig. ~\ref{fig:formation}.
From a practical perspective, the actual relative measurement is different from that extracted from the relative pose estimation, that is, $y_{ji}(t) \neq \hat{y}_{ji}(t) : = h_{ji}(\hat{T}_{ji}(t))$, owing to measurement noise. In practice, it is better to use the relative pose estimate $\hat{T}_{ji}(t)$ or the associated relative measurement estimate $\hat{y}_{ji}(t) = h_{ji}(\hat{T}_{ji}(t))$ for feedback control of cooperative formation, even when the desired formation is defined in terms of the relative measurement, $y_{ji}^{*}$.

Similarly, the absolute pose-based control of robot ${\tt R}_{i}$ can be represented as
\[
u_{ii}(t) = \pi_{ii}(\hat{T}_{i}(t)-{T}_{i}^{*},t)
\]
where ${T}_{i}^{*}$ denotes the desired pose described in the local frame of robot ${\tt R}_{i}$, which is generally defined as its initial pose, that is, ${T}_{i}(0) = 0$ without loss of generality. As stated in~\cite{brooks1985visual}, it is not appropriate for mobile robots to use a single global frame; however, a set of local frames is more suitable for mobile robot estimation and control problems. 

The output feedback formation control is represented as a weighted sum of the absolute and relative pose-based control inputs.
\begin{equation}
  u_{i}(t) = \lambda_{ii}(t) u_{ii}(t) + \sum_{j \in \mathcal{N}_{i}} \lambda_{ji}(t)  u_{ji}(t) 
\end{equation}
where the time-varying weights satisfy the conditions of unit sum and non-negativity, $\sum_{k\in\mathcal{N}_{i} \cup\{i\}}\lambda_{ki}(t) = 1$ and $\lambda_{ki}(t) \in [0,1]$ for all $(i,k)$ and $t$.

%==============================================================%
\section{Conclusions}\label{sec:conclusion}
%==============================================================%
In this paper, we provide an overview of observability analysis and estimation methods for wheeled-mobile multi-robot localization. A novel observability analysis of inter-robot relative pose estimation, without information exchange, concludes that wheel-odometry information exchange is unnecessary as long as inter-robot range and bearing measurements are available. Both Extended Kalman Filter (EKF) and optimization-based M-estimation are applied and compared in ROS/Gazebo simulations of two-robot relative pose and velocity estimation. Robust PGO-based estimation demonstrates higher reliability than EKF-based estimation, especially in the presence of outliers. The same estimation methods are implemented in hardware experiments using two Turtlebot3 robots. 

%-------------------%%-------------------%
\section*{Acknowledgement}
%-------------------%%-------------------%
The authors would like to thank JunGee Hong and Dr. Muhammad Kazim for their support with the hardware experiments and valuable discussions.

\bibliographystyle{IEEEtran}
\bibliography{observability_cooperative_localization}

\begin{thebibliography}{100}
\providecommand{\url}[1]{#1}
\csname url@samestyle\endcsname
\providecommand{\newblock}{\relax}
\providecommand{\bibinfo}[2]{#2}
\providecommand{\BIBentrySTDinterwordspacing}{\spaceskip=0pt\relax}
\providecommand{\BIBentryALTinterwordstretchfactor}{4}
\providecommand{\BIBentryALTinterwordspacing}{\spaceskip=\fontdimen2\font plus
\BIBentryALTinterwordstretchfactor\fontdimen3\font minus
  \fontdimen4\font\relax}
\providecommand{\BIBforeignlanguage}[2]{{%
\expandafter\ifx\csname l@#1\endcsname\relax
\typeout{** WARNING: IEEEtran.bst: No hyphenation pattern has been}%
\typeout{** loaded for the language `#1'. Using the pattern for}%
\typeout{** the default language instead.}%
\else
\language=\csname l@#1\endcsname
\fi
#2}}
\providecommand{\BIBdecl}{\relax}
\BIBdecl

\bibitem{siegwart2011introduction}
R.~Siegwart, I.~R. Nourbakhsh, and D.~Scaramuzza, \emph{Introduction to
  Autonomous Mobile Robots}.\hskip 1em plus 0.5em minus 0.4em\relax Cambridge,
  Massachusetts: MIT Press, 2011.

\bibitem{Yang2008decentralized}
P.~Yang, R.~A. Freeman, and K.~M. Lynch, ``Multi-agent coordination by
  decentralized estimation and control,'' \emph{IEEE Transactions on Automatic
  Control}, vol.~53, no.~11, pp. 2480--2496, 2008.

\bibitem{sakurama2021generalized}
K.~Sakurama, T.~Sugie \emph{et~al.}, ``Generalized coordination of multi-robot
  systems,'' \emph{Foundations and Trends{\textregistered} in Systems and
  Control}, vol.~9, no.~1, pp. 1--170, 2021.

\bibitem{roumeliotis2002distributed}
S.~Roumeliotis and G.~Bekey, ``Distributed multirobot localization,''
  \emph{IEEE Transactions on Robotics and Automation}, vol.~18, no.~5, pp.
  781--795, 2002.

\bibitem{howard2006multi}
A.~Howard, ``Multi-robot simultaneous localization and mapping using particle
  filters,'' \emph{The International Journal of Robotics Research}, vol.~25,
  no.~12, pp. 1243--1256, 2006.

\bibitem{thrun2005multi}
S.~Thrun and Y.~Liu, ``Multi-robot slam with sparse extended information
  filers,'' in \emph{Robotics Research. The Eleventh International Symposium},
  2005, pp. 254--266.

\bibitem{Queralta2020collaborative}
J.~P. Queralta, J.~Taipalmaa, B.~Can~Pullinen, V.~K. Sarker, T.~Nguyen~Gia,
  H.~Tenhunen, M.~Gabbouj, J.~Raitoharju, and T.~Westerlund, ``Collaborative
  multi-robot search and rescue: {P}lanning, coordination, perception, and
  active vision,'' \emph{IEEE Access}, vol.~8, pp. 191\,617--191\,643, 2020.

\bibitem{martinelli2005multi}
A.~Martinelli, F.~Pont, and R.~Siegwart, ``Multi-robot localization using
  relative observations,'' in \emph{IEEE International Conference on Robotics
  and Automation (ICRA)}, 2005, pp. 2797--2802.

\bibitem{miller2020cooperative}
A.~Miller, K.~Rim, P.~Chopra, P.~Kelkar, and M.~Likhachev, ``Cooperative
  perception and localization for cooperative driving,'' in \emph{IEEE
  International Conference on Robotics and Automation (ICRA)}, 2020, pp.
  1256--1262.

\bibitem{wang2023distributed}
S.~Wang, Y.~Wang, D.~Li, and Q.~Zhao, ``Distributed relative localization
  algorithms for multi-robot networks: A survey,'' \emph{Sensors}, vol.~23,
  no.~5, p. 2399, 2023.

\bibitem{Desai2001formationcontrol}
J.~Desai, J.~Ostrowski, and V.~Kumar, ``Modeling and control of formations of
  nonholonomic mobile robots,'' \emph{IEEE Transactions on Robotics and
  Automation}, vol.~17, no.~6, pp. 905--908, 2001.

\bibitem{das2002vision}
A.~K. Das, R.~Fierro, V.~Kumar, J.~P. Ostrowski, J.~Spletzer, and C.~J. Taylor,
  ``A vision-based formation control framework,'' \emph{IEEE Transactions on
  Robotics and Automation}, vol.~18, no.~5, pp. 813--825, 2002.

\bibitem{garcia2021certifiable}
M.~Garcia-Salguero, J.~Briales, and J.~Gonzalez-Jimenez, ``Certifiable relative
  pose estimation,'' \emph{Image and Vision Computing}, vol. 109, p. 104142,
  2021.

\bibitem{kim2023}
J.-H. Kim and K.-K.~K. Kim, ``Vision-based markerless robot-to-robot relative
  pose estimation using {RGB-D} data,'' \emph{Journal of Institute of Control,
  Robotics and Systems}, vol.~29, no.~6, pp. 495--0503, 2023.

\bibitem{Fang2022adaptive}
S.~Fang, H.~Li, and M.~Yang, ``Adaptive cubature split covariance intersection
  filter for multi-vehicle cooperative localization,'' \emph{IEEE Robotics and
  Automation Letters}, vol.~7, no.~2, pp. 1158--1165, 2022.

\bibitem{haynes1970nonlinear}
G.~Haynes and H.~Hermes, ``Nonlinear controllability via {L}ie theory,''
  \emph{SIAM Journal on Control}, vol.~8, no.~4, pp. 450--460, 1970.

\bibitem{hermann1977nonlinear}
R.~Hermann and A.~Krener, ``Nonlinear controllability and observability,''
  \emph{IEEE Transactions on automatic control}, vol.~22, no.~5, pp. 728--740,
  1977.

\bibitem{van1982controllability}
A.~van~der Schaft, ``Controllability and observability for affine nonlinear
  {H}amiltonian systems,'' \emph{IEEE Transactions on Automatic Control},
  vol.~27, no.~2, pp. 490--492, 1982.

\bibitem{bicchi1998observability}
A.~Bicchi, D.~Prattichizzo, A.~Marigo, and A.~Balestrino, ``On the
  observability of mobile vehicle localization,'' in \emph{Theory and Practice
  of Control and Systems}.\hskip 1em plus 0.5em minus 0.4em\relax World
  Scientific, 1998, pp. 142--147.

\bibitem{conticelli2000observability}
F.~Conticelli, A.~Bicchi, and A.~Balestrino, ``Observability and nonlinear
  observers for mobile robot localization,'' \emph{IFAC Proceedings Volumes},
  vol.~33, no.~27, pp. 663--668, 2000.

\bibitem{lorussi2001optimal}
F.~Lorussi, A.~Marigo, and A.~Bicchi, ``Optimal exploratory paths for a mobile
  rover,'' in \emph{IEEE International Conference on Robotics and Automation
  (ICRA)}, vol.~2, 2001, pp. 2078--2083.

\bibitem{martinelli2005observability}
A.~Martinelli and R.~Siegwart, ``Observability analysis for mobile robot
  localization,'' in \emph{IEEE/RSJ International Conference on Intelligent
  Robots and Systems (IROS)}, 2005, pp. 1471--1476.

\bibitem{Huang2010consistentEKF}
G.~P. Huang, A.~I. Mourikis, and S.~I. Roumeliotis, ``Observability-based rules
  for designing consistent {EKF SLAM} estimators,'' \emph{The International
  Journal of Robotics Research}, vol.~29, no.~5, pp. 502--528, 2010.

\bibitem{huang2011observability}
G.~P. Huang, N.~Trawny, A.~I. Mourikis, and S.~I. Roumeliotis,
  ``Observability-based consistent {EKF} estimators for multi-robot cooperative
  localization,'' \emph{Autonomous Robots}, vol.~30, no.~1, pp. 99--122, 2011.

\bibitem{Martinelli2011calibration}
A.~Martinelli, ``State estimation based on the concept of continuous symmetry
  and observability analysis: {T}he case of calibration,'' \emph{IEEE
  Transactions on Robotics}, vol.~27, no.~2, pp. 239--255, 2011.

\bibitem{araki2019range}
B.~Araki, I.~Gilitschenski, T.~Ogata, A.~Wallar, W.~Schwarting, Z.~Choudhury,
  S.~Karaman, and D.~Rus, ``Range-based cooperative localization with nonlinear
  observability analysis,'' in \emph{IEEE Intelligent Transportation Systems
  Conference (ITSC)}, 2019, pp. 1864--1870.

\bibitem{Sharma2012graph}
R.~Sharma, R.~W. Beard, C.~N. Taylor, and S.~Quebe, ``Graph-based observability
  analysis of bearing-only cooperative localization,'' \emph{IEEE Transactions
  on Robotics}, vol.~28, no.~2, pp. 522--529, 2012.

\bibitem{hesch2014cameraimu}
J.~A. Hesch, D.~G. Kottas, S.~L. Bowman, and S.~I. Roumeliotis,
  ``{Camera-IMU-based} localization: {O}bservability analysis and consistency
  improvement,'' \emph{The International Journal of Robotics Research},
  vol.~33, no.~1, pp. 182--201, 2014.

\bibitem{panahandeh2013observability}
G.~Panahandeh, C.~X. Guo, M.~Jansson, and S.~I. Roumeliotis, ``Observability
  analysis of a vision-aided inertial navigation system using planar features
  on the ground,'' in \emph{IEEE/RSJ International Conference on Intelligent
  Robots and Systems (IROS)}, 2013, pp. 4187--4194.

\bibitem{yang2019aided}
Y.~Yang and G.~Huang, ``Aided inertial navigation: {U}nified feature
  representations and observability analysis,'' in \emph{International
  Conference on Robotics and Automation (ICRA)}, 2019, pp. 3528--3534.

\bibitem{panahandeh2016planar}
G.~Panahandeh, S.~Hutchinson, P.~H{\"a}ndel, and M.~Jansson, ``Planar-based
  visual inertial navigation: {O}bservability analysis and motion estimation,''
  \emph{Journal of Intelligent \& Robotic Systems}, vol.~82, pp. 277--299,
  2016.

\bibitem{paul2018alternating}
M.~K. Paul and S.~I. Roumeliotis, ``Alternating-stereo {VINS}: {O}bservability
  analysis and performance evaluation,'' in \emph{Proceedings of the IEEE
  Conference on Computer Vision and Pattern Recognition}, 2018, pp. 4729--4737.

\bibitem{yang2019observability}
Y.~Yang and G.~Huang, ``Observability analysis of aided ins with heterogeneous
  features of points, lines, and planes,'' \emph{IEEE Transactions on
  Robotics}, vol.~35, no.~6, pp. 1399--1418, 2019.

\bibitem{huai2022observability}
J.~Huai, Y.~Lin, Y.~Zhuang, C.~K. Toth, and D.~Chen, ``Observability analysis
  and keyframe-based filtering for visual inertial odometry with full
  self-calibration,'' \emph{IEEE Transactions on Robotics}, vol.~38, no.~5, pp.
  3219--3237, 2022.

\bibitem{hesch2012observability}
J.~A. Hesch, D.~G. Kottas, S.~L. Bowman, and S.~I. Roumeliotis,
  ``Observability-constrained vision-aided inertial navigation,''
  \emph{University of Minnesota, Dept. of Comp. Sci. \& Eng., MARS Lab, Tech.
  Rep}, vol.~1, p.~6, 2012.

\bibitem{gomaa2020observability}
M.~A. Gomaa, O.~De~Silva, G.~K. Mann, and R.~G. Gosine,
  ``Observability-constrained vins for mavs using interacting multiple model
  algorithm,'' \emph{IEEE Transactions on Aerospace and Electronic Systems},
  vol.~57, no.~3, pp. 1423--1442, 2020.

\bibitem{liu2022variable}
C.~Liu, C.~Jiang, and H.~Wang, ``Variable observability constrained
  visual-inertial-{GNSS} {EKF}-based navigation,'' \emph{IEEE Robotics and
  Automation Letters}, vol.~7, no.~3, pp. 6677--6684, 2022.

\bibitem{martinelli2018nonlinear}
A.~Martinelli, ``Nonlinear unknown input observability: {E}xtension of the
  observability rank condition,'' \emph{IEEE Transactions on Automatic
  Control}, vol.~64, no.~1, pp. 222--237, 2018.

\bibitem{martinelli2022nonlinear}
------, ``Nonlinear unknown input observability and unknown input
  reconstruction: {T}he general analytical solution,'' \emph{Information
  Fusion}, vol.~85, pp. 23--51, 2022.

\bibitem{sasaoka2016multi}
T.~Sasaoka, I.~Kimoto, Y.~Kishimoto, K.~Takaba, and H.~Nakashima, ``Multi-robot
  {SLAM} via information fusion extended {K}alman filters,''
  \emph{IFAC-PapersOnLine}, vol.~49, no.~22, pp. 303--308, 2016.

\bibitem{shreedharan2023dkf}
S.~Shreedharan, ``{DKF-SLAM}: {D}istributed {K}alman filtering for multi-robot
  {SLAM},'' Ph.D. dissertation, UC San Diego, 2023.

\bibitem{mourikis2007multi}
A.~I. Mourikis and S.~I. Roumeliotis, ``A multi-state constraint {K}alman
  filter for vision-aided inertial navigation,'' in \emph{IEEE International
  Conference on Robotics and Automation (ICRA)}, 2007, pp. 3565--3572.

\bibitem{melnyk2012cooperative}
I.~V. Melnyk, J.~A. Hesch, and S.~I. Roumeliotis, ``Cooperative vision-aided
  inertial navigation using overlapping views,'' in \emph{IEEE International
  Conference on Robotics and Automation (ICRA)}, 2012, pp. 936--943.

\bibitem{zhu2021cooperative}
P.~Zhu, Y.~Yang, W.~Ren, and G.~Huang, ``Cooperative visual-inertial
  odometry,'' in \emph{IEEE International Conference on Robotics and Automation
  (ICRA)}, 2021, pp. 13\,135--13\,141.

\bibitem{chenchana2018range}
B.~Chenchana, O.~Labbani-Igbida, S.~Renault, and S.~Boria, ``Range-based
  collaborative {MSCKF} localization,'' in \emph{25th International Conference
  on Mechatronics and Machine Vision in Practice (M2VIP)}, 2018, pp. 1--6.

\bibitem{jia2022fej}
S.~Jia, Y.~Jiao, Z.~Zhang, R.~Xiong, and Y.~Wang, ``{FEJ-VIRO}: {A} consistent
  first-estimate {J}acobian visual-inertial-ranging odometry,'' in
  \emph{IEEE/RSJ International Conference on Intelligent Robots and Systems
  (IROS)}, 2022, pp. 1336--1343.

\bibitem{olfati2005distributed}
R.~Olfati-Saber, ``Distributed {K}alman filter with embedded consensus
  filters,'' in \emph{Proceedings of the 44th IEEE Conference on Decision and
  Control (CDC)}, 2005, pp. 8179--8184.

\bibitem{olfati2005consensus}
R.~Olfati-Saber and J.~S. Shamma, ``Consensus filters for sensor networks and
  distributed sensor fusion,'' in \emph{Proceedings of the 44th IEEE Conference
  on Decision and Control (CDC)}, 2005, pp. 6698--6703.

\bibitem{khan2008distributing}
U.~A. Khan and J.~M. Moura, ``Distributing the {K}alman filter for large-scale
  systems,'' \emph{IEEE Transactions on Signal Processing}, vol.~56, no.~10,
  pp. 4919--4935, 2008.

\bibitem{olfati2009kalman}
R.~Olfati-Saber, ``Kalman-consensus filter: {O}ptimality, stability, and
  performance,'' in \emph{Proceedings of the 48h IEEE Conference on Decision
  and Control (CDC)}, 2009, pp. 7036--7042.

\bibitem{cattivelli2010diffusion}
F.~S. Cattivelli and A.~H. Sayed, ``Diffusion strategies for distributed kalman
  filtering and smoothing,'' \emph{IEEE Transactions on Automatic Control},
  vol.~55, no.~9, pp. 2069--2084, 2010.

\bibitem{boyd2006randomized}
S.~Boyd, A.~Ghosh, B.~Prabhakar, and D.~Shah, ``Randomized gossip algorithms,''
  \emph{IEEE Transactions on Information Theory}, vol.~52, no.~6, pp.
  2508--2530, 2006.

\bibitem{bosse2016robust}
M.~Bosse, G.~Agamennoni, and I.~Gilitschenski, ``Robust estimation and
  applications in robotics,'' \emph{Foundations and Trends{\textregistered} in
  Robotics}, vol.~4, no.~4, pp. 225--269, 2016.

\bibitem{barron2019general}
J.~T. Barron, ``A general and adaptive robust loss function,'' in
  \emph{Proceedings of the IEEE/CVF Conference on Computer Vision and Pattern
  Recognition}, 2019, pp. 4331--4339.

\bibitem{bergstrom2014robust}
P.~Bergstr{\"o}m and O.~Edlund, ``Robust registration of point sets using
  iteratively reweighted least squares,'' \emph{Computational optimization and
  applications}, vol.~58, no.~3, pp. 543--561, 2014.

\bibitem{grisetti2010tutorial}
G.~Grisetti, R.~K{\"u}mmerle, C.~Stachniss, and W.~Burgard, ``A tutorial on
  graph-based {SLAM},'' \emph{IEEE Intelligent Transportation Systems
  Magazine}, vol.~2, no.~4, pp. 31--43, 2010.

\bibitem{dellaert2017factor}
F.~Dellaert, M.~Kaess \emph{et~al.}, ``Factor graphs for robot perception,''
  \emph{Foundations and Trends{\textregistered} in Robotics}, vol.~6, no. 1-2,
  pp. 1--139, 2017.

\bibitem{dellaert2021factor}
F.~Dellaert, ``Factor graphs: {E}xploiting structure in robotics,''
  \emph{Annual Review of Control, Robotics, and Autonomous Systems}, vol.~4,
  pp. 141--166, 2021.

\bibitem{latif2014robust}
Y.~Latif, C.~Cadena, and J.~Neira, ``Robust graph slam back-ends: A comparative
  analysis,'' in \emph{IEEE/RSJ International Conference on Intelligent Robots
  and Systems (IROS)}, 2014, pp. 2683--2690.

\bibitem{mcgann_risam_2023}
D.~McGann, J.~R. III, and M.~Kaess, ``Robust incremental smoothing and mapping
  ({riSAM}),'' in \emph{IEEE International Conference on Robotics and
  Automation (ICRA)}, London, {GB}, 2023, pp. 4157--4163.

\bibitem{sharma2011graph}
R.~Sharma, R.~W. Beard, C.~N. Taylor, and S.~Quebe, ``Graph-based observability
  analysis of bearing-only cooperative localization,'' \emph{IEEE Transactions
  on Robotics}, vol.~28, no.~2, pp. 522--529, 2011.

\bibitem{maes2019observability}
K.~Maes, M.~Chatzis, and G.~Lombaert, ``Observability of nonlinear systems with
  unmeasured inputs,'' \emph{Mechanical Systems and Signal Processing}, vol.
  130, pp. 378--394, 2019.

\bibitem{chen2015disturbance}
W.-H. Chen, J.~Yang, L.~Guo, and S.~Li, ``Disturbance-observer-based control
  and related methods—{A}n overview,'' \emph{IEEE Transactions on Industrial
  Electronics}, vol.~63, no.~2, pp. 1083--1095, 2015.

\bibitem{veluvolu2009high}
K.~C. Veluvolu and Y.~C. Soh, ``High-gain observers with sliding mode for state
  and unknown input estimations,'' \emph{IEEE Transactions on Industrial
  Electronics}, vol.~56, no.~9, pp. 3386--3393, 2009.

\bibitem{radke2006survey}
A.~Radke and Z.~Gao, ``A survey of state and disturbance observers for
  practitioners,'' in \emph{American Control Conference (ACC)}, 2006, pp.
  5183--5188.

\bibitem{chen2006micro}
V.~C. Chen, F.~Li, S.-S. Ho, and H.~Wechsler, ``{Micro-Doppler} effect in
  radar: {P}henomenon, model, and simulation study,'' \emph{IEEE Transactions
  on Aerospace and Electronic Systems}, vol.~42, no.~1, pp. 2--21, 2006.

\bibitem{chen2019micro}
V.~C. Chen, \emph{The micro-{D}oppler effect in radar}.\hskip 1em plus 0.5em
  minus 0.4em\relax Artech House, 2019.

\bibitem{ma2019moving}
Y.~Ma, J.~Anderson, S.~Crouch, and J.~Shan, ``Moving object detection and
  tracking with doppler {LiDAR},'' \emph{Remote Sensing}, vol.~11, no.~10, p.
  1154, 2019.

\bibitem{sun2021vessel}
W.~Sun, Z.~Pang, W.~Huang, Y.~Ji, and Y.~Dai, ``Vessel velocity estimation and
  tracking from {D}oppler echoes of {T/RR} composite compact {HFSWR},''
  \emph{IEEE Journal of Selected Topics in Applied Earth Observations and
  Remote Sensing}, vol.~14, pp. 4427--4440, 2021.

\bibitem{kellner2014instantaneous}
D.~Kellner, M.~Barjenbruch, J.~Klappstein, J.~Dickmann, and K.~Dietmayer,
  ``Instantaneous full-motion estimation of arbitrary objects using dual
  {D}oppler radar,'' in \emph{IEEE Intelligent Vehicles Symposium Proceedings},
  2014, pp. 324--329.

\bibitem{kellner2013instantaneous}
D.~Kellner, M.~Barjenbruch, K.~Dietmayer, J.~Klappstein, and J.~Dickmann,
  ``Instantaneous lateral velocity estimation of a vehicle using {Doppler}
  radar,'' in \emph{Proceedings of the 16th International Conference on
  Information Fusion}, 2013, pp. 877--884.

\bibitem{kellner2013egomotion}
D.~Kellner, M.~Barjenbruch, J.~Klappstein, J.~Dickmann, and K.~Dietmayer,
  ``Instantaneous ego-motion estimation using {D}oppler radar,'' in \emph{16th
  International IEEE Conference on Intelligent Transportation Systems (ITSC
  2013)}, 2013, pp. 869--874.

\bibitem{AgarwalCeresSolver2022}
\BIBentryALTinterwordspacing
S.~Agarwal, K.~Mierle, and {The Ceres Solver Team}, ``{Ceres Solver},'' 10
  2023. [Online]. Available: \url{https://github.com/ceres-solver/ceres-solver}
\BIBentrySTDinterwordspacing

\bibitem{rousseeuw2011robust}
P.~J. Rousseeuw and M.~Hubert, ``Robust statistics for outlier detection,''
  \emph{Wiley interdisciplinary reviews: {D}ata mining and knowledge
  discovery}, vol.~1, no.~1, pp. 73--79, 2011.

\bibitem{peterson2023motlee}
M.~B. Peterson, P.~C. Lusk, and J.~P. How, ``{MOTLEE}: {D}istributed mobile
  multi-object tracking with localization error elimination,'' \emph{arXiv
  preprint arXiv:2304.12175}, 2023.

\bibitem{cunningham2010ddf}
A.~Cunningham, M.~Paluri, and F.~Dellaert, ``{DDF-SAM}: {F}ully distributed
  {SLAM} using constrained factor graphs,'' in \emph{2010 IEEE/RSJ
  International Conference on Intelligent Robots and Systems (IROS)}.\hskip 1em
  plus 0.5em minus 0.4em\relax IEEE, 2010, pp. 3025--3030.

\bibitem{cunningham2013ddf}
A.~Cunningham, V.~Indelman, and F.~Dellaert, ``{DDF-SAM 2.0}: {C}onsistent
  distributed smoothing and mapping,'' in \emph{2013 IEEE international
  conference on robotics and automation}.\hskip 1em plus 0.5em minus
  0.4em\relax IEEE, 2013, pp. 5220--5227.

\bibitem{fox2000probabilistic}
D.~Fox, W.~Burgard, H.~Kruppa, and S.~Thrun, ``A probabilistic approach to
  collaborative multi-robot localization,'' \emph{Autonomous robots}, vol.~8,
  pp. 325--344, 2000.

\bibitem{roumeliotis2004propagation}
S.~I. Roumeliotis and I.~M. Rekleitis, ``Propagation of uncertainty in
  cooperative multirobot localization: {A}nalysis and experimental results,''
  \emph{Autonomous Robots}, vol.~17, no.~1, pp. 41--54, 2004.

\bibitem{carrillo2013decentralized}
L.~C. Carrillo-Arce, E.~D. Nerurkar, J.~L. Gordillo, and S.~I. Roumeliotis,
  ``Decentralized multi-robot cooperative localization using covariance
  intersection,'' in \emph{IEEE/RSJ International Conference on Intelligent
  Robots and Systems (IROS)}, 2013, pp. 1412--1417.

\bibitem{Boroson2020}
E.~R. Boroson, R.~Hewitt, N.~Ayanian, and J.-P. de~la Croix, ``Inter-robot
  range measurements in pose graph optimization,'' in \emph{IEEE/RSJ
  International Conference on Intelligent Robots and Systems (IROS)}, 2020, pp.
  4806--4813.

\bibitem{Liu2022}
R.~Liu, Z.~Deng, Z.~Cao, M.~Shalihan, B.~P.~L. Lau, K.~Chen, K.~Bhowmik,
  C.~Yuen, and U.-X. Tan, ``Distributed ranging {SLAM} for multiple robots with
  {Ultra-WideBand} and odometry measurements,'' in \emph{IEEE/RSJ International
  Conference on Intelligent Robots and Systems (IROS)}, 2022, pp.
  13\,684--13\,691.

\bibitem{smith1986representation}
R.~C. Smith and P.~Cheeseman, ``On the representation and estimation of spatial
  uncertainty,'' \emph{The international journal of Robotics Research}, vol.~5,
  no.~4, pp. 56--68, 1986.

\bibitem{smith1990}
R.~Smith, M.~Self, and P.~Cheeseman, \emph{Estimating Uncertain Spatial
  Relationships in Robotics}.\hskip 1em plus 0.5em minus 0.4em\relax Berlin,
  Heidelberg: Springer-Verlag, 1990, p. 167–193.

\bibitem{castellanos2012mobile}
J.~A. Castellanos and J.~D. Tardos, \emph{Mobile robot localization and map
  building: {A} multisensor fusion approach}.\hskip 1em plus 0.5em minus
  0.4em\relax Springer Science \& Business Media, 2012.

\bibitem{sola2017quaternion}
J.~Sola, ``Quaternion kinematics for the error-state {K}alman filter,''
  \emph{arXiv preprint arXiv:1711.02508}, 2017.

\bibitem{barrau2018invariant}
A.~Barrau and S.~Bonnabel, ``Invariant {K}alman filtering,'' \emph{Annual
  Review of Control, Robotics, and Autonomous Systems}, vol.~1, pp. 237--257,
  2018.

\bibitem{hauberg2013unscented}
S.~Hauberg, F.~Lauze, and K.~S. Pedersen, ``Unscented kalman filtering on
  riemannian manifolds,'' \emph{Journal of mathematical imaging and vision},
  vol.~46, pp. 103--120, 2013.

\bibitem{brossard2020code}
M.~Brossard, A.~Barrau, and S.~Bonnabel, ``A code for unscented {K}alman
  filtering on manifolds {(UKF-M)},'' in \emph{IEEE International Conference on
  Robotics and Automation (ICRA)}, 2020, pp. 5701--5708.

\bibitem{zhang2017feedback}
C.~Zhang, A.~Taghvaei, and P.~G. Mehta, ``Feedback particle filter on
  {R}iemannian manifolds and matrix {L}ie groups,'' \emph{IEEE Transactions on
  Automatic Control}, vol.~63, no.~8, pp. 2465--2480, 2017.

\bibitem{li2020unscented}
K.~Li, F.~Pfaff, and U.~D. Hanebeck, ``Unscented dual quaternion particle
  filter for {SE}(3) estimation,'' \emph{IEEE Control Systems Letters}, vol.~5,
  no.~2, pp. 647--652, 2020.

\bibitem{tao2023}
Y.~Tao and S.~S.-T. Yau, ``Outlier-robust iterative extended {K}alman
  filtering,'' \emph{IEEE Signal Processing Letters}, vol.~30, pp. 743--747,
  2023.

\bibitem{agamennoni2015self}
G.~Agamennoni, P.~Furgale, and R.~Siegwart, ``Self-tuning {M}-estimators,'' in
  \emph{IEEE International Conference on Robotics and Automation (ICRA)}, 2015,
  pp. 4628--4635.

\bibitem{pfeifer2019expectation}
T.~Pfeifer and P.~Protzel, ``Expectation-maximization for adaptive mixture
  models in graph optimization,'' in \emph{IEEE International Conference on
  Robotics and Automation (ICRA)}, 2019, pp. 3151--3157.

\bibitem{pfeifer2021advancing}
T.~Pfeifer, S.~Lange, and P.~Protzel, ``Advancing mixture models for least
  squares optimization,'' \emph{IEEE Robotics and Automation Letters}, vol.~6,
  no.~2, pp. 3941--3948, 2021.

\bibitem{ramezani2022aeros}
M.~Ramezani, M.~Mattamala, and M.~Fallon, ``{AEROS: AdaptivE RObust
  least-squares for graph-based SLAM},'' \emph{Frontiers in Robotics and AI},
  vol.~9, p. 789444, 2022.

\bibitem{Allan2019}
D.~A. Allan and J.~B. Rawlings, \emph{Moving Horizon Estimation}.\hskip 1em
  plus 0.5em minus 0.4em\relax Cham: Springer International Publishing, 2019,
  pp. 99--124.

\bibitem{rao2003constrained}
C.~V. Rao, J.~B. Rawlings, and D.~Q. Mayne, ``Constrained state estimation for
  nonlinear discrete-time systems: {S}tability and moving horizon
  approximations,'' \emph{IEEE transactions on automatic control}, vol.~48,
  no.~2, pp. 246--258, 2003.

\bibitem{rawlings2012optimization}
J.~B. Rawlings and L.~Ji, ``Optimization-based state estimation: {C}urrent
  status and some new results,'' \emph{Journal of Process Control}, vol.~22,
  no.~8, pp. 1439--1444, 2012.

\bibitem{civera20101}
J.~Civera, O.~G. Grasa, A.~J. Davison, and J.~M. Montiel, ``1-point {RANSAC}
  for extended {K}alman filtering: {A}pplication to real-time structure from
  motion and visual odometry,'' \emph{Journal of field robotics}, vol.~27,
  no.~5, pp. 609--631, 2010.

\bibitem{s17102164}
M.~B. Alatise and G.~P. Hancke, ``Pose estimation of a mobile robot based on
  fusion of {IMU} data and vision data using an extended {K}alman filter,''
  \emph{Sensors}, vol.~17, no.~10, 2017.

\bibitem{Jazwinski2007}
A.~H. Jazwinski, \emph{Stochastic Processes and Filtering Theory}.\hskip 1em
  plus 0.5em minus 0.4em\relax Mineola, NY: Dover Publications, Inc., 2007.

\bibitem{skoglund2015}
M.~A. Skoglund, G.~Hendeby, and D.~Axehill, ``Extended {K}alman filter
  modifications based on an optimization view point,'' in \emph{18th
  International Conference on Information Fusion (Fusion)}, 2015, pp.
  1856--1861.

\bibitem{carlone2010rao}
L.~Carlone, M.~K. Ng, J.~Du, B.~Bona, and M.~Indri, ``{Rao-Blackwellized}
  particle filters multi robot {SLAM} with unknown initial correspondences and
  limited communication,'' in \emph{IEEE International Conference on Robotics
  and Automation (ICRA)}, 2010, pp. 243--249.

\bibitem{carlone2011simultaneous}
L.~Carlone, M.~Kaouk~Ng, J.~Du, B.~Bona, and M.~Indri, ``Simultaneous
  localization and mapping using rao-blackwellized particle filters in multi
  robot systems,'' \emph{Journal of Intelligent \& Robotic Systems}, vol.~63,
  pp. 283--307, 2011.

\bibitem{saeedi2016multiple}
S.~Saeedi, M.~Trentini, M.~Seto, and H.~Li, ``Multiple-robot simultaneous
  localization and mapping: {A} review,'' \emph{Journal of Field Robotics},
  vol.~33, no.~1, pp. 3--46, 2016.

\bibitem{torres2018range}
A.~Torres-Gonz{\'a}lez, J.~R. Martinez-de Dios, and A.~Ollero, ``Range-only
  {SLAM} for robot-sensor network cooperation,'' \emph{Autonomous Robots},
  vol.~42, no.~3, pp. 649--663, 2018.

\bibitem{funabiki2020range}
N.~Funabiki, B.~Morrell, J.~Nash, and A.-a. Agha-mohammadi, ``Range-aided
  pose-graph-based {SLAM}: {A}pplications of deployable ranging beacons for
  unknown environment exploration,'' \emph{IEEE Robotics and Automation
  Letters}, vol.~6, no.~1, pp. 48--55, 2020.

\bibitem{lajoie2020door}
P.-Y. Lajoie, B.~Ramtoula, Y.~Chang, L.~Carlone, and G.~Beltrame,
  ``{DOOR-SLAM}: {D}istributed, online, and outlier resilient {SLAM} for
  robotic teams,'' \emph{IEEE Robotics and Automation Letters}, vol.~5, no.~2,
  pp. 1656--1663, 2020.

\bibitem{boroson2020inter}
E.~R. Boroson, R.~Hewitt, N.~Ayanian, and J.-P. de~la Croix, ``Inter-robot
  range measurements in pose graph optimization,'' in \emph{IEEE/RSJ
  International Conference on Intelligent Robots and Systems (IROS)}, 2020, pp.
  4806--4813.

\bibitem{rodrigues2021online}
R.~T. Rodrigues, N.~Tsiogkas, A.~Pascoal, and A.~P. Aguiar, ``Online
  range-based {SLAM} using {B}-spline surfaces,'' \emph{IEEE Robotics and
  Automation Letters}, vol.~6, no.~2, pp. 1958--1965, 2021.

\bibitem{papalia2023certifiably}
A.~Papalia, A.~Fishberg, B.~W. O'Neill, J.~P. How, D.~M. Rosen, and J.~J.
  Leonard, ``Certifiably correct range-aided {SLAM},'' \emph{arXiv preprint
  arXiv:2302.11614}, 2023.

\bibitem{papalia2023score}
A.~Papalia, J.~Morales, K.~J. Doherty, D.~M. Rosen, and J.~J. Leonard, ``Score:
  A second-order conic initialization for range-aided slam,'' in \emph{IEEE
  International Conference on Robotics and Automation (ICRA)}, 2023, pp.
  10\,637--10\,644.

\bibitem{dumbgen2023toward}
F.~D{\"u}mbgen, C.~Holmes, B.~Agro, and T.~D. Barfoot, ``Toward globally
  optimal state estimation using automatically tightened semidefinite
  relaxations,'' \emph{arXiv preprint arXiv:2308.05783}, 2023.

\bibitem{holmes2023semidefinite}
C.~Holmes, F.~D{\"u}mbgen, and T.~D. Barfoot, ``On semidefinite relaxations for
  matrix-weighted state-estimation problems in robotics,'' \emph{arXiv preprint
  arXiv:2308.07275}, 2023.

\bibitem{goudar2023optimal}
A.~Goudar, F.~D{\"u}mbgen, T.~D. Barfoot, and A.~P. Schoellig, ``Optimal
  initialization strategies for range-only trajectory estimation,'' \emph{arXiv
  preprint arXiv:2309.09011}, 2023.

\bibitem{wang2017ultra}
C.~Wang, H.~Zhang, T.-M. Nguyen, and L.~Xie, ``Ultra-wideband aided fast
  localization and mapping system,'' in \emph{IEEE/RSJ International Conference
  on Intelligent Robots and Systems (IROS)}, 2017, pp. 1602--1609.

\bibitem{song2019uwb}
Y.~Song, M.~Guan, W.~P. Tay, C.~L. Law, and C.~Wen, ``{UWB/LiDAR} fusion for
  cooperative range-only {SLAM},'' in \emph{IEEE International Conference on
  Robotics and Automation (ICRA)}, 2019, pp. 6568--6574.

\bibitem{nguyen2021viral}
T.-M. Nguyen, S.~Yuan, M.~Cao, T.~H. Nguyen, and L.~Xie, ``{VIRAL SLAM}:
  {T}ightly coupled {Camera-IMU-UWB-Lidar} {SLAM},'' \emph{arXiv preprint
  arXiv:2105.03296}, 2021.

\bibitem{liu2022distributed}
R.~Liu, Z.~Deng, Z.~Cao, M.~Shalihan, B.~P.~L. Lau, K.~Chen, K.~Bhowmik,
  C.~Yuen, and U.-X. Tan, ``Distributed ranging slam for multiple robots with
  ultra-wideband and odometry measurements,'' in \emph{IEEE/RSJ International
  Conference on Intelligent Robots and Systems (IROS)}, 2022, pp.
  13\,684--13\,691.

\bibitem{nguyen2021range}
T.~H. Nguyen, T.-M. Nguyen, and L.~Xie, ``Range-focused fusion of
  {camera-IMU-UWB} for accurate and drift-reduced localization,'' \emph{IEEE
  Robotics and Automation Letters}, vol.~6, no.~2, pp. 1678--1685, 2021.

\bibitem{nguyen2021flexible}
------, ``Flexible and resource-efficient multi-robot collaborative
  visual-inertial-range localization,'' \emph{IEEE Robotics and Automation
  Letters}, vol.~7, no.~2, pp. 928--935, 2021.

\bibitem{zhang2022lidar}
C.~Zhang, X.~Ma, and P.~Qin, ``{LiDAR-IMU-UWB}-based collaborative
  localization,'' \emph{World Electric Vehicle Journal}, vol.~13, no.~2, p.~32,
  2022.

\bibitem{nguyen2022ntu}
T.-M. Nguyen, S.~Yuan, M.~Cao, Y.~Lyu, T.~H. Nguyen, and L.~Xie, ``{NTU VIRAL}:
  {A} visual-inertial-ranging-lidar dataset, from an aerial vehicle
  viewpoint,'' \emph{The International Journal of Robotics Research}, vol.~41,
  no.~3, pp. 270--280, 2022.

\bibitem{delama2023uvio}
G.~Delama, F.~Shamsfakhr, S.~Weiss, D.~Fontanelli, and A.~Fomasier, ``{UVIO}:
  {A}n {UWB}-aided visual-inertial odometry framework with bias-compensated
  anchors initialization,'' in \emph{IEEE/RSJ International Conference on
  Intelligent Robots and Systems (IROS)}, 2023, pp. 7111--7118.

\bibitem{wang2023rvio}
J.~Wang, P.~Gu, L.~Wang, and Z.~Meng, ``{RVIO}: {A}n effective localization
  algorithm for range-aided visual-inertial odometry system,'' \emph{IEEE
  Transactions on Intelligent Transportation Systems}, 2023.

\bibitem{hu2023robust}
J.~Hu, Y.~Li, Y.~Lei, Z.~Xu, M.~Lv, and J.~Han, ``Robust and adaptive
  calibration of {UWB}-aided vision navigation system for {UAVs},'' \emph{IEEE
  Robotics and Automation Letters}, 2023.

\bibitem{hu2023tightly}
C.~Hu, P.~Huang, and W.~Wang, ``Tightly coupled visual-inertial-{UWB} indoor
  localization system with multiple position-unknown anchors,'' \emph{IEEE
  Robotics and Automation Letters}, 2023.

\bibitem{lin2023gnss}
H.-Y. Lin and J.-R. Zhan, ``{GNSS}-denied {UAV} indoor navigation with {UWB}
  incorporated visual inertial odometry,'' \emph{Measurement}, vol. 206, p.
  112256, 2023.

\bibitem{goudar2023range}
A.~Goudar, W.~Zhao, and A.~P. Schoellig, ``Range-visual-inertial sensor fusion
  for micro aerial vehicle localization and navigation,'' \emph{IEEE Robotics
  and Automation Letters}, vol.~9, no.~1, pp. 683--690, 2023.

\bibitem{verma2021multi}
J.~K. Verma and V.~Ranga, ``Multi-robot coordination analysis, taxonomy,
  challenges and future scope,'' \emph{Journal of intelligent \& robotic
  systems}, vol. 102, pp. 1--36, 2021.

\bibitem{brooks1985visual}
R.~Brooks, ``Visual map making for a mobile robot,'' in \emph{IEEE
  International Conference on Robotics and Automation (ICRA)}, vol.~2, 1985,
  pp. 824--829.

\end{thebibliography}

\end{document}